\documentclass[11pt]{article}

\usepackage[letterpaper,margin=1in]{geometry}

\usepackage{palatino}
\usepackage{authblk}
\usepackage[utf8]{inputenc} 
\usepackage[T1]{fontenc}    
\usepackage{times}  
\usepackage{helvet} 
\usepackage{courier}  
\usepackage[hyphens]{url}  
\usepackage{graphicx} 
\usepackage{subfigure}
\usepackage{thmtools, thm-restate}
\usepackage{tabularx}
\usepackage{booktabs}       

\usepackage{hhline}
\usepackage{makecell}
\usepackage{multirow}
\usepackage{multicol}
\usepackage{textcomp}

\usepackage[switch]{lineno}  %
\usepackage[hidelinks,colorlinks=true,linkcolor=blue,citecolor=blue]{hyperref}

\usepackage{cancel}
\usepackage{microtype}      

\usepackage{amsmath}
\usepackage{amssymb,amsopn,algorithm,algorithmic,theorem,float,bbm,bm,enumerate,color,multirow,gensymb}

\usepackage{natbib}
\allowdisplaybreaks[4]
\usepackage[dvipsnames]{xcolor}

\newcommand{\beq}{\begin{equation}}
\newcommand{\eeq}{\end{equation}}

{\theoremstyle{definition} }

\newtheorem{lemm}{Lemma}
\newtheorem{corr}{Corollary}

\newtheorem{asmp}{Assumption}

\newtheorem{defn}{Definition}

\newcommand{\g}{\mathbf{g}}
\newcommand{\h}{\mathbf{h}}

\renewcommand{\b}{\mathbf{b}}

\renewcommand{\u}{\mathbf{u}}
\renewcommand{\v}{\mathbf{v}}

\newcommand{\w}{\mathbf{w}}
\newcommand{\x}{\mathbf{x}}

\newcommand{\cI}{{\cal I}}
\newcommand{\cL}{{\cal L}}

\newcommand{\cP}{{\cal P}}

\newcommand{\cZ}{{\cal Z}}
\newcommand{\cA}{{\cal A}}

\newcommand{\vertiii}[1]{{\left\vert\kern-0.25ex\left\vert\kern-0.25ex\left\vert #1
    \right\vert\kern-0.25ex\right\vert\kern-0.25ex\right\vert}}

\DeclareMathOperator{\tr}{Tr}

\newcommand{\proof}{\noindent{\itshape Proof:}\hspace*{1em}}
 \newcommand{\qed}{\nolinebreak[1]~~~\hspace*{\fill} \rule{5pt}{5pt}\vspace*{\parskip}\vspace*{1ex}}

\newcommand {\commentout}[1] {}

\def\ints{{{\rm Z} \kern -.35em {\rm Z} }}  
\def\smallints{{{\rm Z} \kern -.3em {\rm Z} }}  
\def\pints{{{\rm I} \kern -.15em {\rm N} }}      
\newcommand{\reals}{\mathbb R}

\def\cplx{{{\rm I} \kern -.45em {\rm C} }}       
\def\l2{\rm {\mathcal L}^{2}(\reals)}            

\newcommand{\nr}{\nonumber}
\newcommand{\be}{\begin{eqnarray}}
\newcommand{\ee}{\end{eqnarray}}
\newcommand{\bea}{\begin{eqnarray}}
\newcommand{\eea}{\end{eqnarray}}
\newcommand{\beaa}{\begin{eqnarray*}}
\newcommand{\eeaa}{\end{eqnarray*}}
\newcommand{\bnad}{\begin{nad}}
\newcommand{\enad}{\end{nad}}

\title{Noisy Truncated SGD: Optimization and Generalization}
\date{}
\author[1]{Yingxue~Zhou \thanks{Equal Contribution}}
\author[1]{Xinyan~Li \textsuperscript{*}}
\author[2]{Arindam~Banerjee}

\affil[1]{Department of Computer Science \& Engineering, 
University of Minnesota, Twin Cities}
\affil[2]{Department of Computer Science,
University of Illinois Urbana-Champaign}
\affil[ ]{Emails: \texttt{\{zhou0877@umn.edu, lixx1166@umn.edu, arindamb@illinois.edu\}}}

\begin{document}

\maketitle

\begin{abstract}

Recent empirical work on stochastic gradient descent (SGD) applied to over-parameterized deep learning has shown that most gradient components over epochs are quite small.  Inspired by such observations, we rigorously study properties of Truncated SGD (T-SGD), that truncates the majority of small gradient components to zeros. 
Considering non-convex optimization problems, we show that the convergence rate of T-SGD matches the order of vanilla SGD. We also establish the generalization error bound for T-SGD.
Further, we propose Noisy Truncated SGD (NT-SGD), which adds Gaussian noise to the truncated gradients. We prove that NT-SGD has the same convergence rate as T-SGD for non-convex optimization problems. We demonstrate that with the help of noise, NT-SGD can provably escape from saddle points and requires less noise compared to previous related work. We also prove that NT-SGD achieves better generalization error bound compared to T-SGD because of the noise. Our generalization analysis is based on uniform stability and we show that additional noise in the gradient update can boost the stability. 
Our experiments on {a variety of benchmark} datasets (MNIST, Fashion-MNIST, CIFAR-10, and CIFAR-100) with various networks (VGG and ResNet) validate the theoretical properties of  NT-SGD, i.e.,  NT-SGD matches the speed and accuracy of vanilla SGD while effectively working with sparse gradients, and can 
successfully escape poor local minima.
\end{abstract}

\section{Introduction}

\begin{figure}[t]
\hspace*{-3mm}
    \centering
    \subfigure[VGG-5, MNIST]
    {\includegraphics[width=0.32\columnwidth]{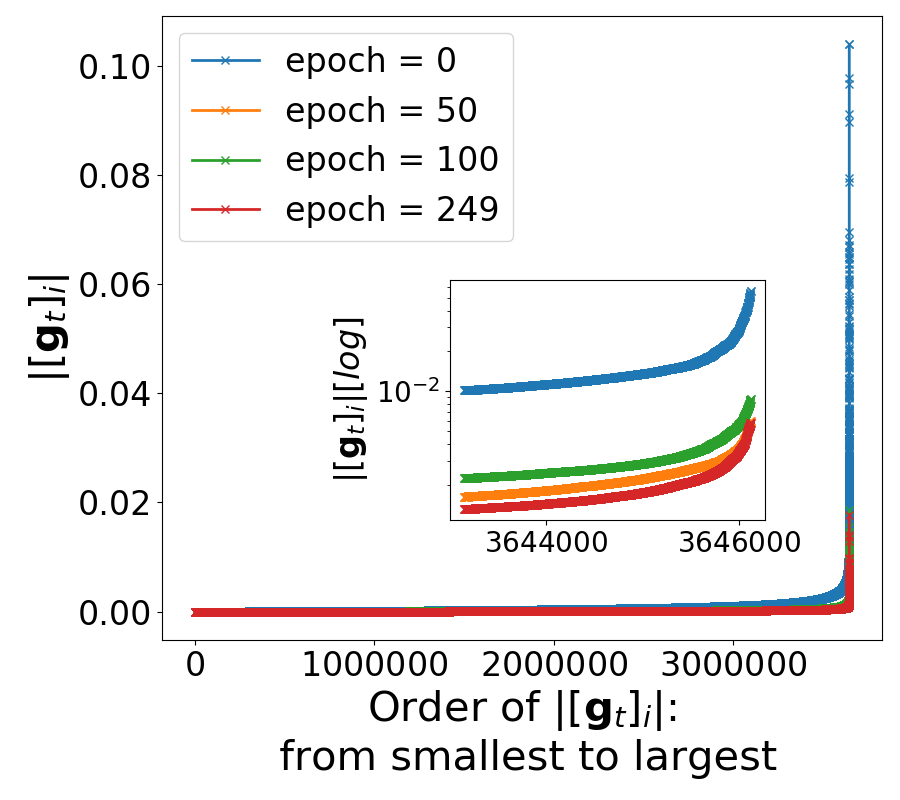}}
    \subfigure[ResNet-18, CIFAR-10]{\includegraphics[width=0.32\columnwidth]{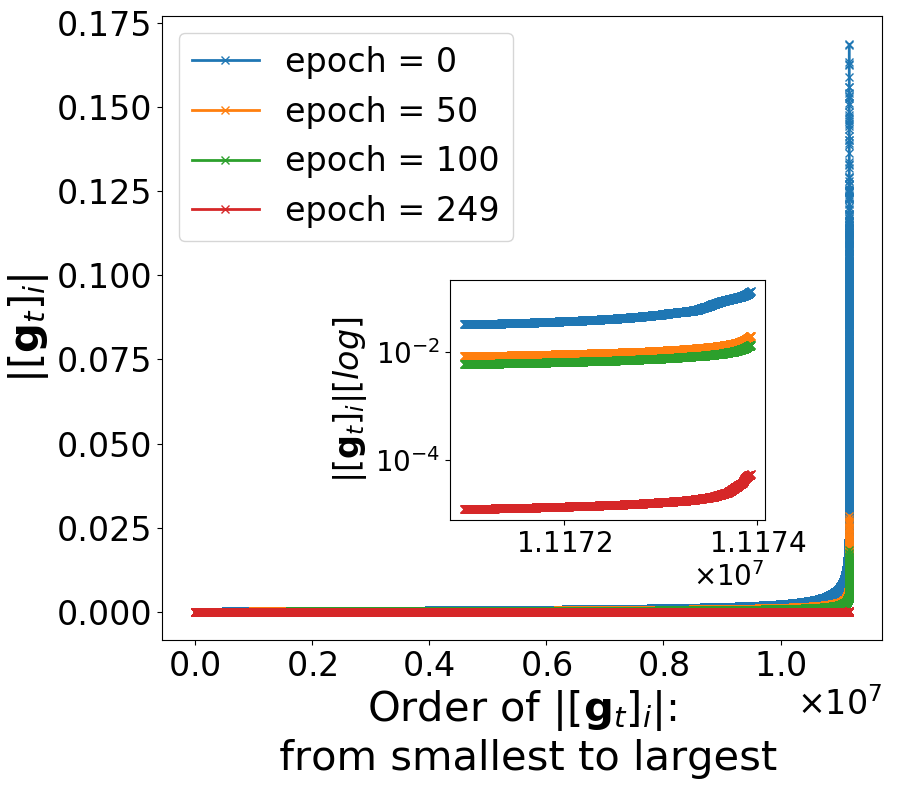}}
    \subfigure[Gradient Sparsity vs. Test Acc]{\includegraphics[width=0.32\textwidth]{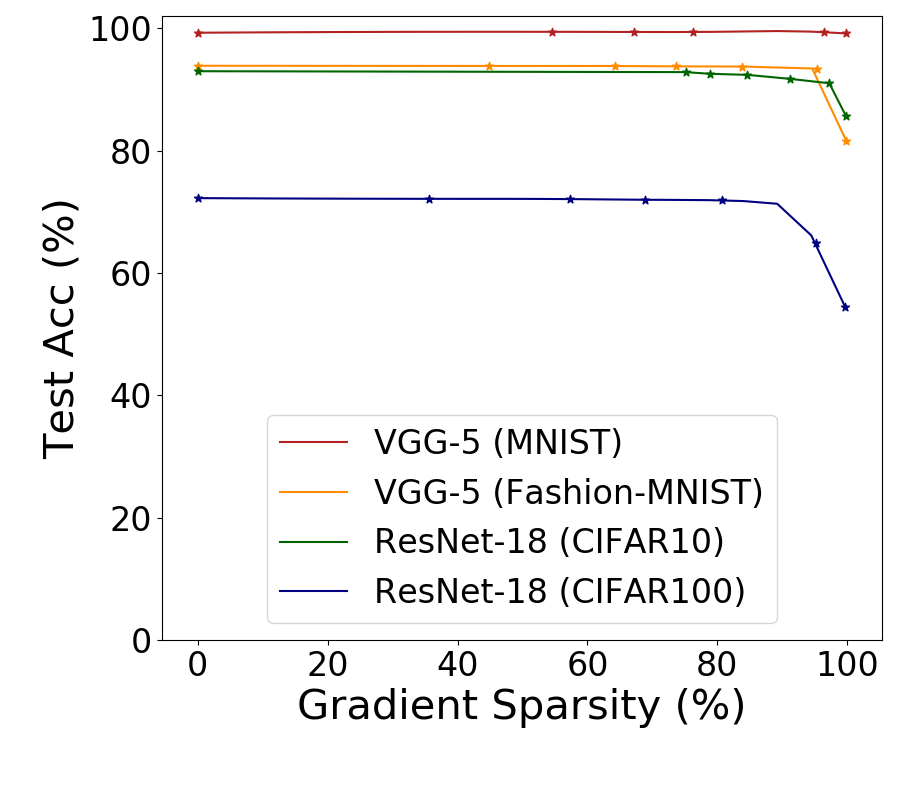}}
    \caption{(a)-(b): Sorted absolute values of stochastic gradient components $|[\g_t]_i|$ for VGG-5 ($p=3,646,154$) trained on MNIST and ResNet-18 ($p=11,173,962$) trained on CIFAR-10 respectively. (c): Test accuracy versus the gradient sparsity for networks trained by Truncated-SGD (T-SGD) which truncates (hard thresholds) the majority of small gradient components to zeros. $\g_t$ is the $p$-dimensional stochastic gradient computed from mini-batch and $[\g_t]_i$ denotes the $i$-th component of $\g_t$ for $i \in [p]$. During training, gradient components decay very fast and only a small portion is significant. We can truncate more than 90\% of the small gradient components to zero without adversely effecting the test accuracy. 
    }
     \label{fig:ut_decay}
\end{figure}

While deep networks have made amazing breakthroughs in processing images, video, natural language, and speech \citep{lecun2015,mina18,dutt18}, training deep networks remains a challenging problem. The large number of parameters in such models leads to substantial computation and communication costs. Fortunately, recent empirical observation suggests that during training most components of the the stochastic gradients of over-parameterized deep networks are near zero \citep{Aji2017,renggli2018,zhou2020bypassing,zhang2021wide}  (Figures~\ref{fig:ut_decay}(a)-(b)). Thus, hard thresholding or truncating gradient components corresponding to small values to zeros is a potentially promising direction. As shown in Figure~\ref{fig:ut_decay}(c), one can truncate 80-90\% or more of the small gradient components to zero in every iteration with virtually no change in generalization performance. Related observations have been explored by gradient sparsification methods \citep{Aji2017,Alistarh2018,stich2018,zhang2021wide} which aim to reduce the communication bottleneck in  distributed  SGD.

Figures~\ref{fig:ut_decay}(a)-(c) represent the motivation for the current work. Given the nature of gradients as shown in Figures~\ref{fig:ut_decay}(a)-(b), in this paper, we focus on rigorously understanding why truncated SGD methods should work well both in terms of optimization and generalization. Note that hard thresholding or truncating the stochastic gradients in every iteration leads to a potentially different trajectory compared to vanilla SGD, and potentially substantially different trajectory compared to adaptive gradient methods \citep{kingma2014adam, duchi2011adaptive, reddi2018adaptive}, which effectively use large step sizes for gradient components with small values. Hence, the empirical results in Figure~\ref{fig:ut_decay}(c), especially the robustness of the test set accuracy to 80-90\% gradient sparsity, do not necessarily automatically follow from the skew structure of the un-truncated stochastic gradients in Figure~\ref{fig:ut_decay}(a)-(b).

Despite several efforts~\citep{Alistarh2018,stich2018,signsgd,quantized-sgd} on deriving theoretical convergence results for gradient sparsification methods for non-convex problems, most existing results have proved convergence to stationary points, which include saddle points. However, the existence of saddle points can slow down and even prevent the algorithm from converging. Moreover, understanding of the generalization behavior of such sparse or truncated gradient methods is limited to empirical observations and has not been well studied in theory. 

To bridge the gap, we first analyze the optimization and generalization properties of Truncated-SGD (T-SGD) that truncates (hard thresholds) the majority of small gradient components to zeros in each iteration. Considering non-convex smooth problems, we analyze the convergence of T-SGD to stationary point
i.e., bound on the norm of the empirical gradient. We show that convergence rate of NT-SGD matches that of vanilla SGD. Then, we provide an $O(T/n)$ generalization bound for T-SGD,  given $n$ samples and $T$ iteration steps, based on uniform stability \citep{hardt2016train,bousquet2002stability}. To the best of our knowledge, we are the first to perform such analysis for truncated or sparse gradient methods.

Next, we propose Noisy Truncated SGD (NT-SGD), a variant of T-SGD that adds Gaussian noise to all components of the truncated sparse gradient vector, and analyze the optimization and generalization behavior of NT-SGD. For optimization, we consider non-convex smooth problems and show that NT-SGD matches the convergence rate of SGD. We prove that NT-SGD can escape from saddle points with less noise compared to previous related work \citep{jin2019nonconvex}.  In particular, we show that the noise variance needed to escape can be improved from a constant to $O({1}/{\Lambda})$, where $\Lambda \in [1, p]$ is the stable rank, measuring the curvature at a saddle point. If a saddle point has significant negative curvatures in many directions, $\Lambda$ becomes large, resulting in a small noise. 

We also analyze the generalization error of NT-SGD using uniform stability \citep{bousquet2002stability,hardt2016train} from the perspective of discretized generalized Langevin dynamics  \citep{mou2018generalization, raginsky2017non}. We show that the generalization bound of NT-SGD is $O(\sqrt{T}/n\sigma)$, which improves the dependence on iteration number $T$ over the $O(T/n)$ generalization bound of T-SGD with the help of noise.

Further, we show higher gradient sparsity leads to smaller constants in the $O(\cdot)$ notation implying better stability and hence generalization. In fact, extremely high sparsity adversely impacts the optimization but helps the generalization, illustrating a trade-off. The bound also has an inverse dependency on $\sigma$, showing that increasing the noise variance $\sigma^2$ actually improves stability. Such inverse dependence has been found in recent work on smoothed analysis applied to other learning problems~\citep{sivakumar2020structured,haghtalab2020smoothed}. Our generalization analysis applies to a general version of NT-SGD which covers a wide range of noise scaling and applies to non-smooth problems.

Finally, we evaluate both T-SGD and NT-SGD on various benchmark classification tasks, viz., MNIST, Fashion-MNIST, CIFAR-10 and CIFAR-100. Empirical evidence illustrates that in spite of significant gradient truncation, e.g., more than 90\% of gradient components truncated to zeros, both T-SGD and NT-SGD converges as fast as vanilla SGD and generalizes as well as SGD. In addition, we also empirically demonstrate that NT-SGD escapes sharp minima with the help of the noise.

The rest of the paper is organized as follows. We start with a brief review of related work in Section \ref{sec:related_work}. In Section \ref{sec:prelim}, we introduce notation and preliminaries for the paper. In Section \ref{sec:tsgd}, we formally introduce our T-SGD algorithm, characterize its optimization, and provide a generalization bound based on uniform stability. We study the optimization and generalization behavior of NT-SGD in Section~\ref{sec:ntsgd}, show it provably converges in expectation, investigate its behavior of escaping from saddle points, and demonstrate that NT-SGD achieves better generalization error bound compared to T-SGD as the stability improves considerably due to noise. Section \ref{sec:experiments} presents the empirical evidence to support our theories. We conclude the paper in Section \ref{sec:conclusion}. All proofs and additional experimental results are deferred to Appendix. 
\label{sec:intro}

\section{Related work}

{\bf Gradient sparsification.} Gradient sparsification has been well-studied in the last decade to save the communication cost in synchronous SGD.  Methods such as gradient quantization \citep{quantized-sgd,wen2017,jiang2018,stich2018,wangni2018,haddadpour2019} that quantizes the gradient
to a small number of bits, gradient sparsification \citep{Aji2017} that selects top $k$ components of the gradient, signSGD \citep{signsgd, karimireddy2019error} that only considers signs of gradient components, low-precision SGD \citep{desa2017,desa2018,yang2019swalp} that limits the number of bits to transmit, have been proposed to reduce the communication cost. Most of the existing works on gradient sparsification focused on deriving the convergence rate of empirical risk minimization problems,
i.e. their performance on training data. For non-convex problems,  \cite{Alistarh2018,quantized-sgd,signsgd,karimireddy2019error} showed compressed gradients achieve the same rate of convergence as SGD, i.e., ${1}/{\sqrt{T}}$, to stationary point.
Although \cite{karimireddy2019error} proposed a 
``max-margin'' explanation for the discrepancy in generalization behaviors among different methods, 
to the best of our knowledge, generalization bounds 
have not been studied yet.

{\bf Noisy gradient methods.} Introducing additional noise in the stochastic gradient has been popular in training deep nets. Noisy gradient methods have proven to be useful in escaping from saddle points \cite{jin2017escape,jin2019nonconvex}. Non-convex optimization is ubiquitous in machine learning applications, especially for deep neural networks.  Since finding the global minimum in non-convex problems generally is NP-hard (in general intractable) \citep{wang2019differentially}, the utility of an algorithm is typically measured by the convergence to a first-order stationary point which can be a local minimum, a local maximum, or a saddle point.
\cite{du2017gradient} showed gradient descent (GD) can be significantly slowed down by saddle points, taking exponential time to escape. Later,
\cite{jin2017escape,jin2019nonconvex} showed adding isotropic noise is enough for both GD and SGD to escape saddle points without additional assumptions \citep{daneshmand2018escaping}. Besides, the study on 
Stochastic gradient Langevin dynamics (SGLD) \citep{mou2018generalization,raginsky2017non,li2019generalization} 
has demonstrated the benefit of adding noise to guarantee good generalization properties. 
Other works on noisy gradient such as DP-SGD (differentially private SGD) \citep{bassily2019private, bassily2014private, wang2019differentially} have 
utilized the randomness in noise to protect the privacy of training data.


{\bf Generalization via uniform stability.} Stability is a classical approach to derive generalization bounds pioneered by \cite{rogers1978finite,devroye1979distribution}. Stability measures the sensitivity of the learning algorithm to changes in the dataset such as leaving one of the samples out or replacing it with a different one. There has been a variety of stability notions depending on the choice of how to measure the effect of the change in datasets \citep{shalev2010learnability,bousquet2002stability, Kearns}. 
The notion of uniform stability was introduced by \cite{bousquet2002stability} to derive general high-probability bounds on the generalization error.  \cite{hardt2016train} provided the first generalization bound of SGD using uniform stability, which is most related to our work.  Uniform stability has also been used to derive generalization bound for noisy gradient methods such as SGLD \citep{mou2018generalization,raginsky2017non,li2019generalization} and differentially private SGD \citep{bassily2019private}. 
It is worth mentioning that the prior approach to derive generalization bounds on the gradient relies on \emph{uniform convergence} of empirical gradient to population gradient 
\citep{mei2016landscape,foster2018uniform,wang2019differentially}. 


\label{sec:related_work}

\section{Background and Preliminaries}
Many fundamental machine learning tasks involve solving empirical risk minimization (ERM): given a loss function $\ell$ and a dataset $S=\left\{z_{1}, \ldots, z_{n}\right\}$ drawn i.i.d. from the underlying distribution $\mathcal{P}$, 
find a model $\mathbf{w} \in \mathbb{R}^{p}$ that minimizes the empirical risk, i.e., 
\begin{equation}
 \cL_S(\w) =  \frac{1}{n} \sum_{i=1}^n \ell(\w, z_i)~.
\end{equation}

The ultimate goal is to find a minimizer of the population risk \cite{shalev2014understanding, bubeck2015convex}, i.e.,   
\begin{equation}
\cL_{\cP}(\w) = \mathbb{E}_{z \sim \cP} [\ell(\w, z)]~.
\end{equation}

Thus, given a minimizer $\w_S^\star$ obtained by solving the empirical risk minimization, one needs to characterize the generalization error of $\w_S^\star$, i.e., $\cL_{\cP}(\w_S^\star) - \cL_{S}(\w_S^\star)$.
  
In this paper, we consider non-convex smooth loss functions which is popular in practice.

We also make the following assumptions about the non-convex loss function $\ell$ throughout the paper \citep{bubeck2015convex, shalev2010learnability}:

\begin{asmp}\label{asmp: bounded_gradient}
For any $\w \in \mathbb{R}^p$ and example $z$, the loss $\ell$ has bounded gradient, i.e., $\| \nabla \ell (\w, z) \| \leq G$.
\end{asmp}

\begin{asmp} \label{asmp: zi_grad}
For any sample $z$, the loss function $\ell$ is bounded below by $\ell^\star$  and has $L$-Lipschitz gradient, i.e.,
\begin{equation} 
 \| \nabla \ell(\w, z) - \nabla \ell(\w^\prime, z) \| \leq  L \|\w - \w^\prime\|,  ~ \forall   \w,\w^\prime \in \mathbb{R}^p.
 \end{equation}
\end{asmp}

Assumption~\ref{asmp: bounded_gradient} implies that the norms of both population gradient and empirical gradient are bounded by $G$. The above assumptions are widely used in non-convex optimization problems \citep{shalev2010learnability, wang2019differentially,hardt2016train}.

\textbf{Notations.} For a vector $\mathbf{v} \in \mathbb{R}^{p}$,  $[\mathbf{v}]_{i}$ denotes the $i$-th coordinate of $\mathbf{v},$ where $i \in[p]$. We use $\| \mathbf{v} \|$ to represent the $\ell_{2}$-norm of $\mathbf{v}$. For a matrix $A$, $\|A\|_2$ is the spectral norm of $A$, and $\lambda_{\min}(A)$ denotes the minimum eigenvalue of $A$. We use $H(\w)$ to denote the Hessian of $\cL_S(\w)$, i.e, $H(\w) = \nabla^2 \cL_S(\w)$.

\textbf{Generalization via Stability.} Now we introduce the stability. Consider the general setting that $\cA$ is a randomized algorithm (such as SGD) and $S = \{z_1, ..., z_n\}$ be $n$ samples drawn i.i.d. from $\cP$. Denote $\cA(S)$ as the output of $\cA$ with the input $S$ and $\ell(\cA(S); z)$ denote the loss function of $\cA(S)$ applying to a sample $z$. Note that $\cA(S)$ is random due to the randomness of $\cA$. Thus, $\ell(\cA(S); z)$ is also a random variable due to $\cA(S)$.

With $\mathcal{L}_{\mathcal{P}}(\mathcal{A}(S)) = \mathbb{E}_{z\sim \cP}[ \ell(\mathcal{A}(S);z)]$ to denote the population risk and $\mathcal{L}_S(\mathcal{A}(S)) = \frac{1}{n}\sum_{z\in S} \ell(\mathcal{A}(S);z)$ to be the empirical risk, the expected generalization error is  \citep{hardt2016train}  

\begin{equation}
\text{err}_{\text {gen}}(\cA(S)) \triangleq \mathbb{E}_{S, \mathcal{A}}  \left[\mathcal{L}_{S}(\mathcal{A}(S))- \mathcal{L}_{\mathcal{P}}(\mathcal{A}(S)) \right] ~.
\end{equation}

To bound the generalization error of $\cA$, we employ the notion of uniform stability from \cite{hardt2016train,bousquet2002stability}.  
\begin{defn}
[Uniform stability] \label{defn: us} A randomized algorithm $\cA$ is $\alpha$-uniformly stable if for all dataset $S, S^\prime \in \cZ^{n}$ such that $S$ and $S^\prime$  differ in at most one example, we have 

\begin{equation}
\sup_{z} \mathbb{E}_{\cA}\left[ \ell(\cA(S); z)- \ell\left(\cA(S^\prime);z\right)\right] \leq \alpha~.
\end{equation}
\end{defn}

Here, the expectation is taken only over the randomness of $\cA$. 
In the context of optimization, $\cA$ can be a gradient-based iterative algorithm that minimize the empirical risk.

In this paper, $\cA$ is referred to as  T-SGD or NT-SGD algorithm. We recall the important argument that uniform stability implies generalization in expectation \citep{hardt2016train}.

\begin{restatable}{theo}{theogen}
\textbf{(Generalization via uniform stability)} \label{theo:gen_stab}
Let $\cA$ be $\alpha$-uniformly stable with respect to the gradient. Then we have 

\begin{equation}
     | \mathbb{E}_{S, \cA}  [\cL_{S}(\cA(S)) - \cL_{\cP}(\cA(S))] | \leq \alpha~.
\end{equation}
\end{restatable}

With $\cA$ being T-SGD or NT-SGD and $\w_S = \cA(S)$ being the output of $\cA$ applied to training data $S$, we can quantify the expected generalization error, i.e., $\text{err}_{\text {gen}}(\w_S)$. 
\label{sec:prelim}

\section{Truncated SGD (T-SGD)}
\label{sec:tsgd}

The T-SGD algorithm is motivated by the observation that during training, only a small fraction of gradient components are significant (Figure \ref{fig:ut_decay}(a)-(b)) \citep{Aji2017,renggli2018,zhou2020bypassing}. In fact, truncating the small gradient components 
to zeros in every iteration seem to have minimal effect on the performance (Figure~\ref{fig:ut_decay}(c)). Note, T-SGD and gradient truncation is opposite to what adaptive gradient methods such as Adam \citep{reddi2018adaptive} do. While Adam applies small step sizes to large gradient coordinates to ensure the even update in each direction, T-SGD only updates the directions with large gradient values. The pseudo-code of T-SGD is given in Algorithm \ref{algo:soft_cut}. Given $n$ training samples, at each iteration $t$, T-SGD first samples a mini-batch $B_t$ uniformly with replacement from $S$ and computes the mini-batch gradient $ g\left(\mathbf{w}_{t}, B_{t}\right)=\frac{1}{\left|B_{t}\right|} \sum_{z_{i} \in B_{t}} \nabla \ell\left(\mathbf{w}_{t}, z_{i}\right)$. Then, for $\varepsilon \in [0, 1]$, T-SGD calls gradient truncation $\textbf{GT}(g\left(\mathbf{w}_{t}, B_{t}\right), \varepsilon^2)$ in Algorithm \ref{algo: gt} to calculate a hard-threshold $\kappa_{\varepsilon,t}$ and  truncates components of  $g(\w_t, B_t)$ based on  $\kappa_{\varepsilon,t}$ to get truncated gradient $\tilde \g_t$ so that: 
\begin{equation}
[\tilde \g_t]_i = \begin{cases} \left[g(\w_t, B_t)\right]_i ~, & ~\text{if} ~|\left[g(\w_t, B_t)\right]_i| \geq \kappa_{\varepsilon, t}~,\\ 
0~, & \text{otherwise}~. 
\end{cases}
\end{equation}

The cut threshold $\kappa_{\varepsilon, t}$ is dynamically changed over iterate $t$. The value of $\kappa_{\varepsilon, t}$ is determined such that after truncating  $g(\w_t, B_t)$ to $\tilde \g_t$, we have $\| \tilde \g_t\|^2 \geq (1-\varepsilon^2)\| g(\w_t, B_t)\|^2$. Note that larger $\varepsilon$ leads to larger $\kappa_{\varepsilon,t}$ and sparser truncated gradients $\tilde \g_t$.

\begin{restatable}{theo}{theotsgdopt}\label{theo:tsgd_opt}
Under Assumptions \ref{asmp: bounded_gradient} and \ref{asmp: zi_grad}, for any $T >0$, T-SGD with $\eta_t = \eta = O(\frac{1}{\sqrt{T}})$ for any $0 \leq \varepsilon \leq \min(1, \frac{L}{2\sqrt{T}})$, we have 
\begin{equation}
    \mathbb{E}\|  \nabla \cL_S(\w_J)\|^2 \leq O\left({LG^2}/{\sqrt{T}}\right)
\end{equation}
where $\mathbf{w}_{J}$ is uniformly sampled from $\left\{\mathbf{w}_{1}, \mathbf{w}_{2}, \ldots, \mathbf{w}_{T}\right\}$ produced by T-SGD
and the expectation is over the randomness of T-SGD  and the random draw of $\w_J$.
\end{restatable}

\begin{algorithm}[t] 
\caption{Truncated SGD}
	\begin{algorithmic}[1] \label{algo:soft_cut}
		\STATE \textbf{Input}: Training set $S$, certain loss $\ell(\cdot)$, initial point $\w_0$
		\STATE \textbf{Set}:  Noise parameter $\sigma$, iteration time $T$,  learning rate $\eta_t$, cut rate $\varepsilon^2$.
		\FOR{$t = 0,...,T$}
		\STATE $g(\w_t, B_t) = \frac{1}{|B_t|}\sum_{z_i \in B_t}\nabla \ell(\w_t,z_i)$, with $B_t$ uniformly sampled from $S$ with replacement.
		\STATE  Call $\textbf{GT}(g(\w_t, B_t), \varepsilon^2)$ to calculate the cut threshold $\kappa_{\varepsilon, t}$.
		\FOR{$i = 1,...,p$}
		    \STATE If $|\left[g(\w_t, B_t)\right]_i|< \kappa_{\varepsilon, t}$, then $[\tilde \g_t]_i = 0 $, else $[\tilde \g_t]_i = \left[g(\w_t, B_t)\right]_i $.
		\ENDFOR
		\STATE Update parameter using sparse gradient $ \tilde \g_t$: $\w_{t+1}=\w_{t}-\eta_t \tilde \g_t$.
		\ENDFOR 
	\end{algorithmic}
\end{algorithm}

\begin{algorithm}[h]
\caption{Gradient Truncation (GT)}
\begin{algorithmic}[1] \label{algo: gt}
\STATE  \textbf{Input}: Gradient $\g = \left[[\g]_1,,...,[\g]_p \right] \in \mathbb{R}^{d}$, cut rate $\varepsilon^2$
\STATE Sort the squares of gradient coordinates $[\g]_1^2,...,[\g]_p^2$ by descending order: $[\g]_{(1)}^2 \geq, ..., \geq [\g]_{(p)}^2$.
\STATE $g_{CumSum} = 0$
\FOR{$i =1,...,p$}
\IF{$g_{CumSum} \geq (1-\varepsilon^2)\|\g\|^2$ }
\STATE \textbf{Return} $| [\g]_i |$ and \textbf{Halt}.
\ENDIF
\STATE $g_{CumSum} = g_{CumSum} + [\g]_i^2$
\ENDFOR
\end{algorithmic}
\end{algorithm}

Theorem \ref{theo:tsgd_opt} shows that the convergence rate of T-SGD is $O(1/\sqrt{T})$ which matches the rate of SGD, meaning that truncating a portion of the small gradient components will not affect the convergence rate.

Next we bound the generalization error of T-SGD based on uniform stability \citep{bousquet2002stability,hardt2016train}. To establish the uniform stability of T-SGD, we consider two datasets $S$ and $S^\prime$ that differ in at most one example. The goal is to bound $\sup_{z} \mathbb{E} [\ell(\w_T; z)- \ell\left(\w_T^\prime;z\right)]$ where $\w_T = \cA(S)$ and $\w_T^\prime = \cA(S')$ are respectively the outputs of T-SGD with inputs $S$ and $S^\prime$. 
Here the expectation is over the randomness of T-SGD and we omit the subscript T-SGD in the expectation $\mathbb{E}$. Since the function is $G$-Lipschitz  by Assumption \ref{asmp: bounded_gradient}, we have $\sup_{z} \mathbb{E}|\nabla \ell(\w_T); z)-\nabla \ell\left(\w_T^\prime);z\right)| \leq \sup_z G~ \mathbb{E} \|\w_T -\w_T^\prime \|$. Thus, the problem is reduced to bounding
$\mathbb{E}\|\w_T -\w_T^\prime \|$. We follow the gradient expansivity approach, first proposed by \cite{hardt2016train}, to show that $ \mathbb{E}\|\w_{t+1} -\w_{t+1}^\prime \|$ can be bounded by $ \mathbb{E}\|\w_{t} -\w_{t}^\prime \|$ with additional terms for any $t \leq T$, and applying the inequality recursively.

\begin{restatable}{theo}{tsgdstab}
Under Assumptions \ref{asmp: bounded_gradient} and \ref{asmp: zi_grad}, suppose that we run T-SGD for $T$ iterations with batch size
$|B_t| = 1$, $\forall t \in [T]$, $\varepsilon \leq O(\frac{1}{n})$. Then,

\hspace{2mm}  \indent 1. T-SGD with step size $\eta_t = O(\frac{1}{\sqrt{T}})$,  is $O\left(\frac{G^2T}{n}\right)$-uniformly stable.

\hspace{2mm} \indent 2. T-SGD with step size $\eta_t = \frac{c}{t}$ for any $c > 0$,  is $O\left(\frac{G^2T^{1 - \frac{1}{Lc +1}}}{n}\right)$-uniformly stable.
\label{theo:general_stab_tsgd}
\end{restatable}

Theorem \ref{theo:general_stab_tsgd} shows the stability of T-SGD scales linearly with the number of iterations if one chooses step size $\eta_t = O(1/\sqrt{T})$ 
in order to match the step size for optimization as
in Theorem \ref{theo:tsgd_opt}. If one chooses a decaying step size as $\eta_t =c/t$ for T-SGD for $c >0$, the stability bound can be improved to $O\left({G^2 T^{1 - {1}/{Lc +1}}}/{n}\right)$~. However, a decaying step size $\eta_t =c/t$ will make T-SGD converges much slower, i.e., $\mathbb{E}\|  \nabla \cL_S(\w_J)\|^2 \leq O(LG^2/\log T)$ for $\eta_t = O(1/t)$, illustrating a fundamental trade-off between optimization and stability \citep{chen2018stability}.

With Theorem \ref{theo:gen_stab}, we obtain the following generalization error bound for T-SGD.

\begin{restatable}{corr}{corrgentsgd} \label{corr:gen_tsgd}
Under Assumptions \ref{asmp: bounded_gradient} and \ref{asmp: zi_grad}, suppose that we run T-SGD for $T$ iterations with  batch size
$|B_t| = 1$, $\forall t \in [T]$, $\varepsilon \leq \min(1, \frac{c^\prime}{n})$ for any $c^\prime > 0$. Then, 

\hspace{2mm}  \indent 1. T-SGD with step size $\eta_t = O(\frac{1}{\sqrt{T}})$ has generalization error bound as
\begin{equation}
\operatorname{err}_{\mathrm{gen}}(\w_T) \leq O\left({G^2T}/{n}\right)~. 
\end{equation}
\hspace{2mm} \indent 2. T-SGD with step size $\eta_t = \frac{c}{t}$ for any $c > 0$ has generalization error bound as
\begin{equation}
\operatorname{err}_{\mathrm{gen}}(\w_T) \leq O\left(\frac{G^2T^{1 - \frac{1}{Lc +1}}}{n}\right)~.
\end{equation}
\end{restatable}
The stability and generalization analysis works for general batch size and corresponding results can be found in the proofs of Theorem \ref{theo:general_stab_tsgd}. The above bound holds for $\w_J$ that is uniformly sampled from $\left\{\mathbf{w}_{1}, \mathbf{w}_{2}, \ldots, \mathbf{w}_{T}\right\}$ as well.

\section{Noisy Truncated SGD (NT-SGD)}
\label{sec:ntsgd}
In this section, we introduce a variant of T-SGD called Noisy truncated SGD (NT-SGD) by adding Gaussian noise in the truncated iterative update. Given the truncated gradient $\tilde \g_t$ as in T-SGD,  NT-SGD updates the iterates as follows, for $\beta \in [0, \frac{1}{2}]$,
\begin{align}
\mathbf{w}_{t+1}=\mathbf{w}_{t}-\eta_{t} \tilde \g_t + \eta_{t}^{\frac{1}{2} + \beta} \b_t,~ \text{where}~ \b_t \sim \mathcal{N}\left(0,\sigma^2\mathbb{I}\right)~. 
\label{eq:generalized_nt_sgd}
\end{align}

Note that NT-SGD with $\beta = \frac{1}{2}$ captures differential private SGD type of algorithm \citep{bassily2019private} and $\beta = 0$ captures the SGLD type algorithm \citep{mou2018generalization,welling2011bayesian} without truncating the gradient. The pseudo code of NT-SGD is given in Algorithm \ref{algo:ntsgd} which is obtained by replacing line 8 in Algorithm \ref{algo:soft_cut} with $\mathbf{w}_{t+1}=\mathbf{w}_{t}-\eta_{t} \tilde \g_t + \eta_{t}^{\frac{1}{2} + \beta} \b_t,~ \text{where}~ \b_t \sim \mathcal{N}\left(0,\sigma^2\mathbb{I}\right)~$.

\begin{algorithm}[t] 
\caption{Noisy Truncated SGD}
	\begin{algorithmic}[1] \label{algo:ntsgd}
		\STATE \textbf{Input}: Training set $S$, certain loss $\ell(\cdot)$, initial point $\w_0$.
		\STATE \textbf{Set}:  Noise parameter $\sigma$, iteration time $T$,  learning rate $\eta_t$, cut rate $\varepsilon^2$,  $\beta \in [0, \frac{1}{2}]$.
		\FOR{$t = 0,...,T$}
		\STATE $g(\w_t, B_t) = \frac{1}{|B_t|}\sum_{z_i \in B_t}\nabla \ell(\w_t,z_i)$, with $B_t$ uniformly sampled from $S$ with replacement.
		\FOR{$i = 1,...,p$}
		    \STATE If $|\left[g(\w_t, B_t)\right]_i|< \kappa_{\varepsilon, t}$, then $[\tilde \g_t]_i = 0 $, else $[\tilde \g_t]_i = \left[g(\w_t, B_t)\right]_i $.
		\ENDFOR
		\STATE Update parameter using sparse gradient $ \tilde \g_t$: $\mathbf{w}_{t+1}=\mathbf{w}_{t}-\eta_{t} \tilde \g_t + \eta_{t}^{\frac{1}{2} + \beta} \b_t,~ \text{where}~ \b_t \sim \mathcal{N}\left(0,\sigma^2\mathbb{I}\right)~. $
		\ENDFOR 
	\end{algorithmic}
\end{algorithm}

\subsection{Optimization}
\label{sec:ntsgd_opt}
We first present rate of convergence and then show that NT-SGD can escape from saddle points.

\begin{restatable}{theo}{theoopt}\label{theo:opt}
Under Assumptions \ref{asmp: bounded_gradient} and \ref{asmp: zi_grad}, for any $T >0$,  NT-SGD with $\eta_t = \eta = \frac{1}{\sqrt{T}}$, $\beta = \frac{1}{2}$ $\sigma^2 = \frac{R}{p}$ for any $R \geq 0$, $0\leq \varepsilon \leq \min(1, \frac{\eta L}{2})$, we have
\begin{equation}\label{eq:ntsgd_optimization}
    \mathbb{E}\|  \nabla \cL_S(\w_J)\|^2 \leq O\left({L\left(G^2 + R^2)\right)}/{\sqrt{T}}\right)
\end{equation}
where $\mathbf{w}_{J}$ is uniformly sampled from $\left\{\mathbf{w}_{1}, \mathbf{w}_{2}, \ldots, \mathbf{w}_{T}\right\}$ produced by NT-SGD
and the expectation is over the randomness of NT-SGD and the random draw of $\w_J$.
\end{restatable}

Theorem \ref{theo:opt} encapsulates the converges rate of T-SGD (when $\varepsilon^2 = 0$) and SGD (when  $R =0$).

Now we analyze NT-SGD in terms of its ability to escape from saddle points. We first review some basic definitions and assumptions. A first-order stationary point can be a local minimum, a local maximum, or even a saddle point \citep{jin2019nonconvex}:

\begin{defn}
For a differentiable function $f: \mathbb{R}^p \rightarrow \mathbb{R}$, a stationary point $\x \in \mathbb{R}^p$ is a
\begin{itemize}
 \item \textbf{local minimum}, if there exists $\delta>0$ such that $f(\mathbf{x}) \leq f(\mathbf{y})$ for any $\mathbf{y}$ with $\|\mathbf{y}-\mathbf{x}\| \leq \delta$.
\item \textbf{local maximum}, if there exists $\delta>0$ such that $f(\mathbf{x}) \geq f(\mathbf{y})$ for any $\mathbf{y}$ with $\|\mathbf{y}-\mathbf{x}\| \leq \delta$.
\item \textbf{saddle point}, otherwise.
\end{itemize}
\end{defn}

Since distinguishing saddle points from local minima for smooth functions is  NP-hard in general \citep{anandkumar2016efficient}, following \cite{jin2019nonconvex}, we focus on escaping from ``strict saddle points''. Let $\lambda_{\min}(A)$ represent the smallest eigenvalue of a matrix $A$. A ``strict saddle point'' is defined as follows:

\begin{defn}

For a twice-differentiable function $f$, $\x$ is a strict saddle point if $\x$ is a stationary point and $\lambda_{\min }\left(\nabla^{2} f(\mathbf{x})\right)<0$.

\end{defn}

A strict saddle point has zero gradient so that gradient descent will be stuck, but there are directions along which the function can decrease. 
Let $\nabla^2 \ell (\w, z)$ be the Hessian of the loss $\ell(\w, z)$. We make the following assumption about the loss. 

\begin{asmp} \label{asmp: hessian-lip}
The twice-differentiable function $\ell$ is $\rho$-Hessian Lipschitz, i.e., for any $z \in \cZ$, $\forall \w_1, \w_2 \in \mathbb{R}^p$, $\left\|\nabla^{2} \ell\left(\w_1, z\right)-\nabla^{2} \ell\left(\w_2, z\right)\right\| \leq \rho\left\|\w_1-\w_2\right\|$.
\end{asmp}

We consider $\w_{t_0}$ for $t_0 \geq 0$ to be a strict saddle point with sharp negative curvature, i.e., $\lambda_{\min}(H(\w_{t_0})) \leq-\sqrt{\rho \gamma}$ for a certain $\gamma >0$, where $H(\w_{t_0}) = \nabla^{2} \cL_{S} (\mathbf{w}_{t_0})$. 
Inspired by the analysis in \cite{daneshmand2018escaping,jin2019nonconvex}, in Theorem \ref{theo:escape} we show that NT-SGD can provably escape saddle points.

\begin{restatable}{theo}{theoescape}
\label{theo:escape}
\textbf{(Escaping Saddle Points)} Under Assumptions \ref{asmp: bounded_gradient}, \ref{asmp: zi_grad}, and \ref{asmp: hessian-lip}, let $\w_{t_0}$ be the strict saddle point such that $\lambda_{\min }\left(H (\mathbf{w}_{t_0})\right) \leq-\sqrt{\rho \gamma}$.
Running NT-SGD starting from $\w_{t_0}$ for 
\begin{equation}
\tau \geq \left(24 + 4\log\left(\frac{\sqrt{\gamma}/(2\sqrt{\rho}\eta + G)}{ G \max\{1, 10G/\gamma\}}+ 4p\right)  \right)/\left(\eta^2\rho\gamma\right) 
\end{equation}
iterations, with
\begin{equation}
\eta = \min\{\frac{1}{L},~ \frac{\sqrt{\gamma} \Lambda_{\tau}}{144\sqrt{\rho} p G},~ \frac{\gamma \Lambda_{\tau}}{576p GL}\}~, \qquad \sigma^2 \geq \frac{576 G^2}{\Lambda_{\tau}} \cdot \max \{1, 10G/\gamma \}~,  
\end{equation}
where $\Lambda_{\tau} = \frac{\tr\left((\mathbb{I}-\eta H(\w_{t_0}))^{2\tau}\right)}{\|(\mathbb{I}-\eta H(\w_{t_0}))^{2\tau}\|_2}$ is the stable rank\footnote{Here, the stable rank 
of a  positive-semi-definite matrix $B$ means the ratio of the trace of $B$ to the spectral norm of $B$, i.e., $\frac{\tr(B)}{\|B\|}$. So  $\Lambda_{\tau} = \frac{\tr\left((\mathbb{I}-\eta H(\w_{t_0}))^{2\tau}\right)}{\|(\mathbb{I}-\eta H(\w_{t_0}))^{2\tau}\|_2}$.
} of $(\mathbb{I}-\eta H(\w_{t_0}))^{2\tau}$, yields: 
\begin{equation}
    \mathbb{E}[\cL_{S}(\w_{t_0+\tau})] - \cL_{S}(\w_{t_0}) \leq - \frac{3 \sqrt{\gamma} G}{2\sqrt{\rho}}~,
\end{equation}
where $\w_{t_0+\tau}$ is the $\tau$-th iterate starting from $\w_{t_0}$, and the expectation is over the randomness of NT-SGD which includes the draw of the mini-batch and the noise.
\end{restatable}

Theorem \ref{theo:escape} shows that running NT-SGD for $\tau$ iterations from the saddle point $\w_{t_0}$ with suitable choice of step size $\eta$ and noise variance $\sigma^2$  guarantees the loss to decrease in expectation. 
The dependence of $\tau$ on $\rho\gamma$ and $\eta$, i.e., $\tau \geq O(1/(\eta^2\rho\gamma))$ ignoring other factors, shows that the sharper the negative curvature and the larger the step size $\eta$ is, the fewer the iterations needed to escape from a saddle point.  Theorem \ref{theo:escape} shows that $\eta$ needs to be small and has a potential dimensional dependency,   i.e., $\eta \leq O({\Lambda_{\tau}}/{p})$.  It is worth mentioning that our result improves $\eta$ from $O({1}/{p})$ \citep{jin2017escape} to $O({\Lambda_{\tau}}/{p})$ with the stable rank $\Lambda_\tau \in [1,~p]$. If the saddle point $\w_{t_0}$ has significant negative curvatures in many directions, $\Lambda_\tau$ is large 
implying a larger step size $\eta$, which results in a smaller $\tau$ implying fewer iterations needed to escape. Also, the required noise variance $\sigma^2$ can be improved from a constant to $O({1}/{\Lambda_\tau})$. With large $\Lambda_\tau$, less noise is needed in order to escape.
Note that with the result on escaping saddle points, one can derive a bound on convergence to a second-order stationary point following approaches in \cite{daneshmand2018escaping,jin2019nonconvex}. 

\subsection{Improved Generalization Bound}
\label{sec:ntsgd_gen}

To establish the bound on stability, we leverage discretized Langevin dynamics and Fokker-Planck equations, which have been used in the existing analysis of SGLD \citep{mou2018generalization, raginsky2017non}. Considering $T$ steps, we denote $\w_1, \ldots, \w_T$ as the sequence of parameters after applying  NT-SGD to $S$ and denote $p_1, \ldots, p_T$ as the corresponding distributions over parameters. Let $S'$ be a neighboring dataset which differs from $S$ at a singe sample, and $\w_1^\prime, \ldots, \w_T^\prime$ and $p_1^\prime, \ldots, p_T^\prime$ be the sequence of parameters and the corresponding distributions. Given two probability distributions with probability density functions of $p$ and $q$ respectively, 
let $D_{H}(p \| q)$ denote the squared Hellinger distance between
the density function $p$ and $q$: $D_{H}(p \| q) \triangleq \frac{1}{2} \int_{\mathbb{R}^{d}}(\sqrt{p}-\sqrt{q})^{2} d \w~$. Our analysis uses the following bound on uniform stability based on the Hellinger distance \citep{mou2018generalization}: 

\begin{restatable}{prop}{propstabhellinger}
\label{prop: stab_hellinger}
Let $S, S^\prime$ be two datasets of size $n$ differing in one sample point. Let $\cA$ be a randomized algorithm applied on $S$ and $S'$ to obtain distributions $p$ and $p'$ over the parameters. If $\cA$ is $\alpha$-uniformly stable as in Definition~\ref{defn: us}, assuming the loss function $\ell(\w, z)$ is uniformly bounded by $C$, then we have $\alpha \leq 2 C\sqrt{D_H(p \| p')}$.
\end{restatable}

Thus, our analysis focuses on bounding $D_{H}\left(p_{T} \| p_{T}^{\prime}\right)$. We first focus on bounding $D_{H}\left(p_{t} \| p_{t}^{\prime}\right)$ for $t \in [T]$. Note that $D_{H}\left(p_{t} \| p_{t}^{\prime}\right) = 0$ for $t =0$. For $t >0$, analyzing $D_{H}\left(p_{t} \| p_{t}^{\prime}\right)$ requires establishing discretized Fokker-Planck equations that measure the dynamics of $p_t$ and $p_t^\prime$ within one step. Thus, we can show

\begin{restatable}{lemm}{lemmstephell}\label{lemm: step_hell}
For $\eta_t^{1-2\beta} \leq \frac{\sigma^2 \ln 4}{12 (1-\epsilon^2) G^2}$ in  NT-SGD, with Assumption \ref{asmp: bounded_gradient}, we have, for all $t \in [T]$, 
\begin{equation}
D_{H}\left(p_{t+1} \| p_{t+1}^{\prime}\right) \leq D_{H}\left(p_{t} \| p_{t}^{\prime}\right)+\frac{16\left(1-\epsilon^{2}\right) G^{2} \eta_{t}^{1-2 \beta}}{\sigma^{2} n^{2}}~.
\end{equation}
\end{restatable}

Since $D_{H}\left(p_{t} \| p_{t}^{\prime}\right) = 0$ for $t =0$, by induction, we have
\begin{equation}
D_{H}\left(p_{T} \| p_{T}^{\prime}\right) \leq \frac{16\left(1-\epsilon^{2}\right) G^{2} \sum_{t=1}^{T} \eta_{t}^{1-2 \beta}}{\sigma^{2} n^{2}}~.
\end{equation}

With Proposition \ref{prop: stab_hellinger}, Theorem \ref{theo:gen_stab}, and Lemma \ref{lemm: step_hell},  we have the expected generalization error bound as follows:

\begin{restatable}{theo}{theogenstab} \label{theo: gen}
\vspace{-2mm}
Consider $T$ iterations of generalized NT-SGD with 
$\eta_t^{1-2\beta} \leq \frac{\sigma^2 \ln 4}{12 (1-\epsilon^2) G^2}$\footnote{This condition suggests that step size $\eta_t$ needs to be bounded by a constant for $t \in [T]$.}, $\beta \in [0,\frac{1}{2}]$, $\sigma^2$ and $\epsilon \in [0, 1]$, $|B_t| =1, \forall t \in [T]$. Suppose the loss function is uniformly bounded by $C$. Under Assumption \ref{asmp: bounded_gradient}, we have 
\begin{equation}\label{eq:ntsgd_generalization}
\operatorname{err}_{\text {gen }}\left(\mathbf{w}_{T}\right) \leq O\left({C G \frac{1}{{n\sigma}}\sqrt{(1-\epsilon^2)\sum_{t=1}^T \eta_t^{1-2\beta}}} \right)
\end{equation}
\end{restatable}


The stability-based bound exhibits a $O(\frac{1}{n})$ rate of convergence, with the complexity factor mainly depends on the square root of the aggregated step sizes $\sum_{t=1}^T \eta_t^{1-2\beta}$, implying that a small step size leads to a small generalization error. The bound also suggests that the larger cut rate $\varepsilon^2$ is, the more stable the algorithm becomes in terms of a constant factor. Such dependence is due to the fact that with a large $\varepsilon^2$, the majority of the gradient updates reduce to zeros, thus the difference between running NT-SGD on $S$ and $S^\prime$ becomes small. In an extreme case, when $\varepsilon^2=1$, such difference reduces to zero as all gradient components are zeros. Besides, the generalization improves as noise variance $\sigma^2$ increases since the noise helps to smooth the divergence between running a randomized algorithm on neighboring datasets.
We also note that the step size plays an important role in the generalization bound. Below, we give some examples with respect to the step size along with the choice of $\beta$. 

\begin{corr}
Choose $\beta = 0$ for NT-SGD, under the conditions in Theorem \ref{theo: gen}, 

1. If $\eta_t = O(1)$, then we have ~$\operatorname{err}_{\text {gen }}\left(\mathbf{w}_{T}\right) \leq O(\frac{\sqrt{(1-\epsilon^2)T}}{\sigma n})$;

2. If $\eta_t = O({1}/{\sqrt{T}})$, then we have ~ $\operatorname{err}_{\text {gen }}\left(\mathbf{w}_{T}\right) \leq O(\frac{\sqrt{(1-\epsilon^2)}T^{1/4}}{\sigma n})$;

3. If $\eta_t = O({1}/{t})$, then we have ~ $\operatorname{err}_{\text {gen }}\left(\mathbf{w}_{T}\right) \leq O(\frac{\sqrt{(1-\epsilon^2) \log T}}{\sigma n})$. 
\end{corr}

The above results show that as the step size decreases, the stability and generalization
error bound improves. As discussed in \cite{chen2018stability}, there is a trade-off between convergence and stability controlled by the step size; one needs to balance the optimization error and stability by choosing a suitable step size. The bound applies to NT-SGD when $\beta = 1/2$, with a generalization bound of $O(\frac{\sqrt{T}}{n})$.




Compared to T-SGD that has $O(T/n)$
generalization error with $\eta_t =O({1}/{\sqrt{T}})$, which only exploits randomness from choosing mini-batch to derive the stability analysis of T-SGD without considering injecting Gaussian noise. NT-SGD improves generalization bound to $O(T^{1/4}/n)$ with $\beta = 0$.  

We discuss a few additional properties of the above results. First, Theorem \ref{theo: gen} only requires the assumptions on bounded gradient and loss function, i.e., $\ell$ is Lipschitz and bounded. Thus, our bound  in Theorem \ref{theo: gen} applies to both convex and non-convex functions. Second, the above results hold for non-smooth loss functions as well. In contrast, the stability analysis of T-SGD requires the loss function to be smooth, i.e., Assumption \ref{asmp: zi_grad}.


\section{Experimental Results}

\begin{figure}[t]
    \centering
    \subfigure[MNIST, $\sigma = 10^{-3}$]%
    {
    \includegraphics[width=.4\textwidth]{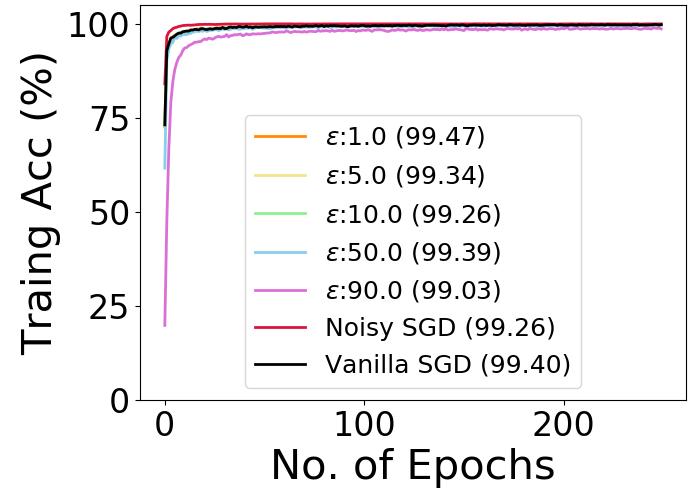}}
    \subfigure[Fashion, $\sigma = 10^{-3}$]%
    {
    \includegraphics[width=.4\textwidth]{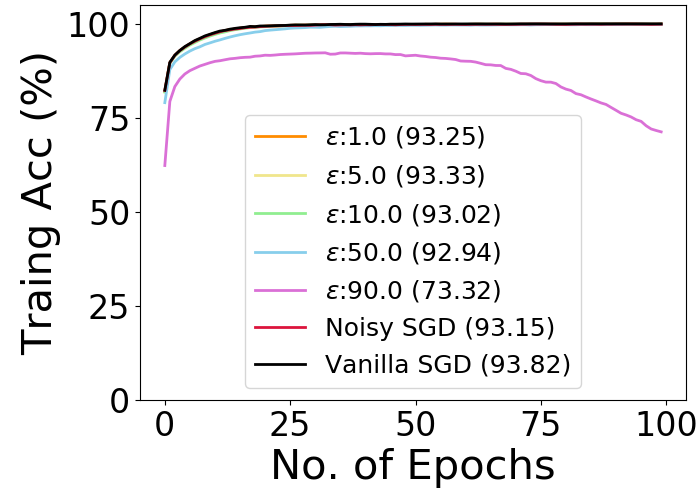}}
    \subfigure[CIFAR10, $\sigma = 10^{-4}$]
    {
    \includegraphics[width=.4\textwidth]{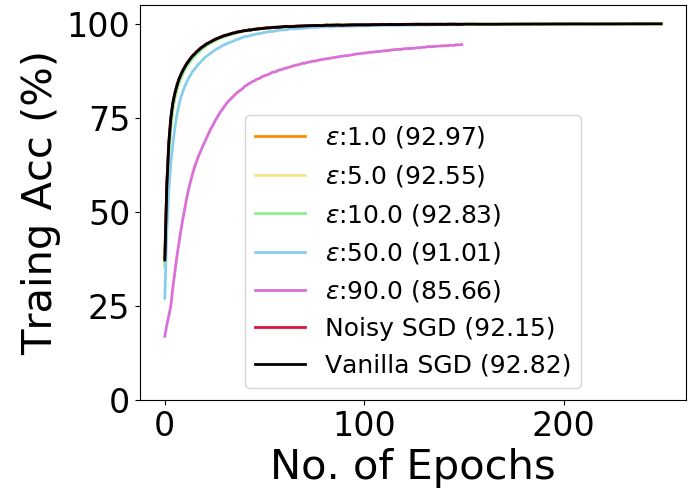}}
    \subfigure[CIFAR100, $\sigma = 10^{-3}$]
    {
    \includegraphics[width=.4\textwidth]{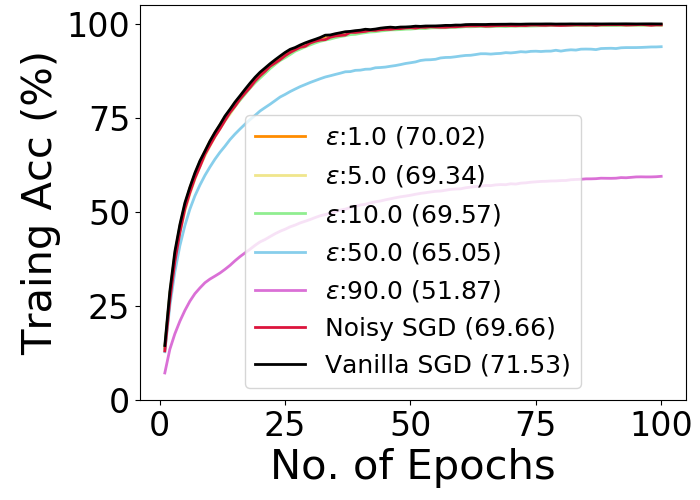}}
    \caption{ Training dynamics of NT-SGD at a fixed noise level $\sigma$. 
    Legends indicate the choice of $\varepsilon$ and the numbers in brackets are the test accuracy at convergence. NT-SGD with moderate injected noise matches the performance of vanilla SGD. However, NT-SGD can suffer if we truncate too many gradient coordinates.
    } 
     \label{fig:lazy_sgd_train_sigma}
\end{figure}

In this section, we conduct a series of experiments to investigate the performance of T-SGD and NT-SGD, and address the following question: under various cut rates $\varepsilon^2$ and noise levels $\sigma$, how does NT-SGD perform in terms of both optimization and generalization? We also investigates how does the injected noise help NT-SGD escape from poor local minima
\footnote{Since 
a saddle point is hard to construct for complicated deep learning models, we follow the experimental design in \cite{zhu2019} to show that NT-SGD can escape poor local minima.}.

\textbf{Experimental setup.} We consider several deep learning scenarios: VGG-5 ($p=3,646,154$) from the family of Visual Geometry Group network \cite{vgg} trained on MNIST \citep{mnist} and Fashion-MNIST \citep{fashion-mnist}, and ResNet-18 ($p=11,173,962$) from the family of residual neural network \citep{resnet} trained on CIFAR-10 \citep{cifar10} and CIFAR-100 \citep{cifar10}. Vanilla SGD and noisy SGD\footnote{For noisy SGD, we use the following gradient update: $\mathbf{w}_{t+1}=\mathbf{w}_{t}-\eta_{t} (\g_t + \b_t),~ \text{where}~ \b_t \sim \mathcal{N}\left(0,\sigma^2\mathbb{I}\right)$.} are considered as baseline methods. We use constant learning rate $\eta = 0.1$ and batch size 100 for all datasets, and minimize cross-entropy loss for a fixed number of epochs. We have not included dropout, batch normalization and other common regularization used in deep learning since the focus of experiments is to compare the performance of NT-SGD with baselines. All experiments have been run on NVIDIA Tesla K40m GPUs, and we repeat each experiment 5 times and report the mean of training and test accuracy.

\begin{figure}[t]
    \centering
    \subfigure[MNIST, $\varepsilon^2 = 10\%$]%
    {
    \includegraphics[width=.4\textwidth]{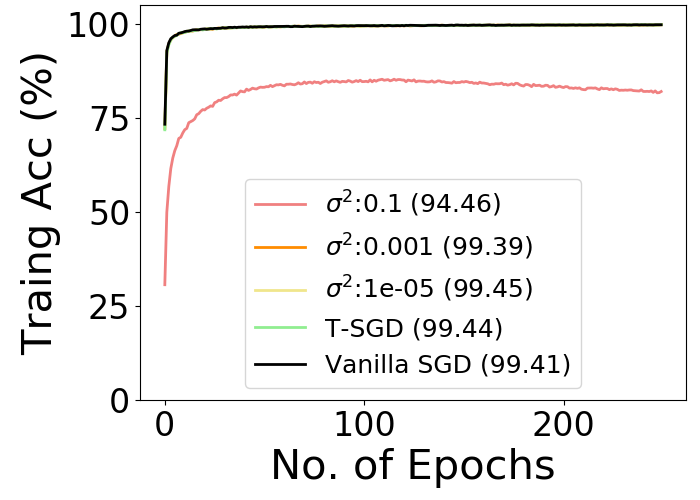}}
    \subfigure[Fashion, $\varepsilon^2 = 10\%$]%
    {
    \includegraphics[width=.4\textwidth]{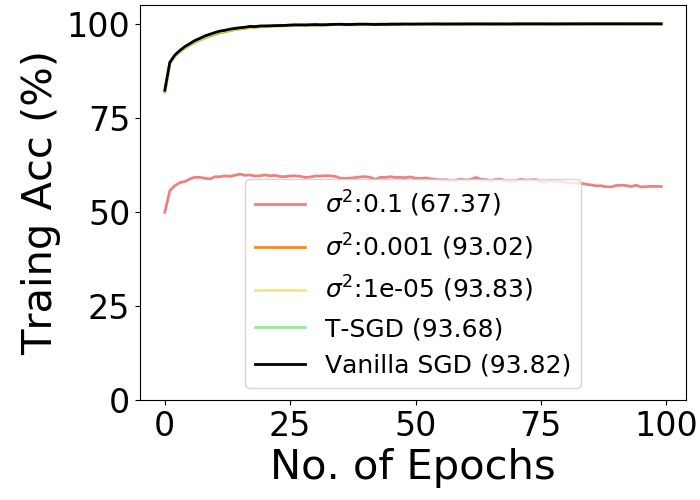}}
    \subfigure[CIFAR10, $\varepsilon^2 = 10\%$]
    {
    \includegraphics[width=.4\textwidth]{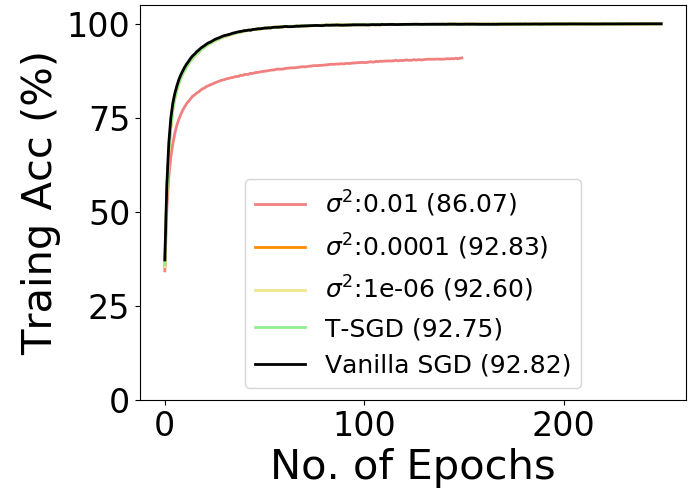}}
    \subfigure[CIFAR100, $\varepsilon^2 = 10\%$]
    {
    \includegraphics[width=.4\textwidth]{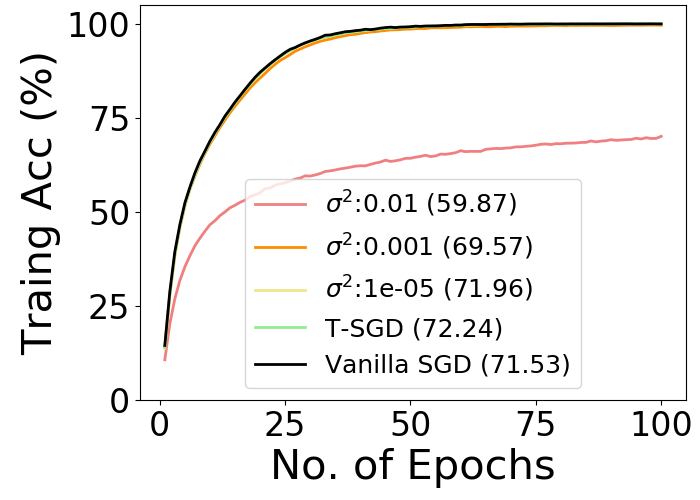}}
    \caption{ Training dynamics of NT-SGD at a fixed cut rate $\varepsilon^2$. Legends indicate the choice of $\sigma^2$, and the numbers in brackets are the test accuracy at convergence. NT-SGD with moderate injected noise matches the performance of vanilla SGD. However, NT-SGD can suffer if we add too much noise. 
    } 
     \label{fig:lazy_sgd_train_eps}
\end{figure}

\begin{figure}[t]
    \centering
    \subfigure[VGG-5, MNIST]{
    \includegraphics[width=0.4\textwidth]{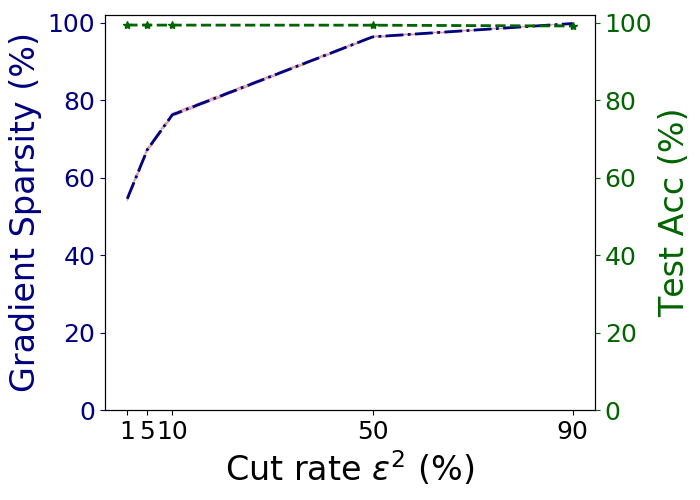}
    }
    \subfigure[VGG-5, Fashion]{
    \includegraphics[width=0.4\textwidth]{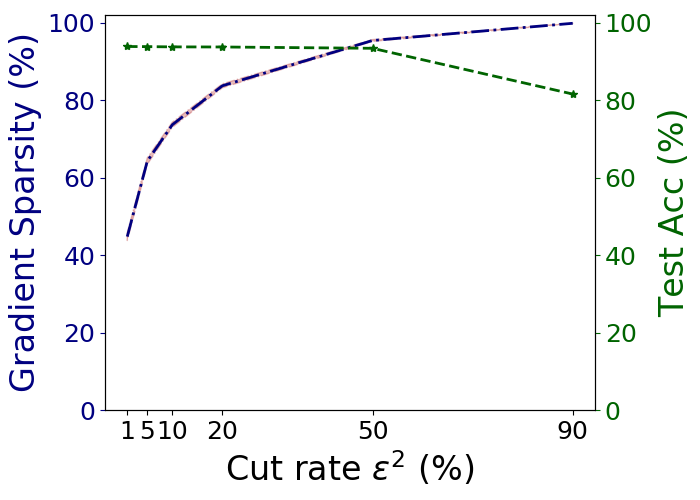}
    }\hspace{2mm}
    \subfigure[ResNet-18, CIFAR10]{
    \includegraphics[width=0.4\textwidth]{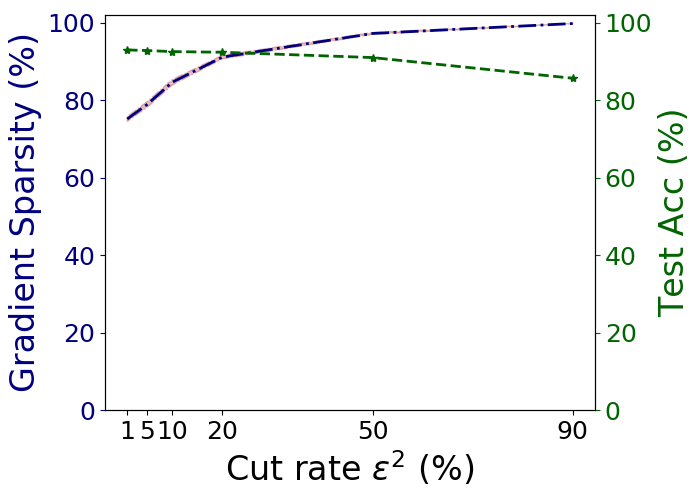}
    }
    \subfigure[ResNet-18, CIFAR100]{
    \includegraphics[width=0.4\textwidth]{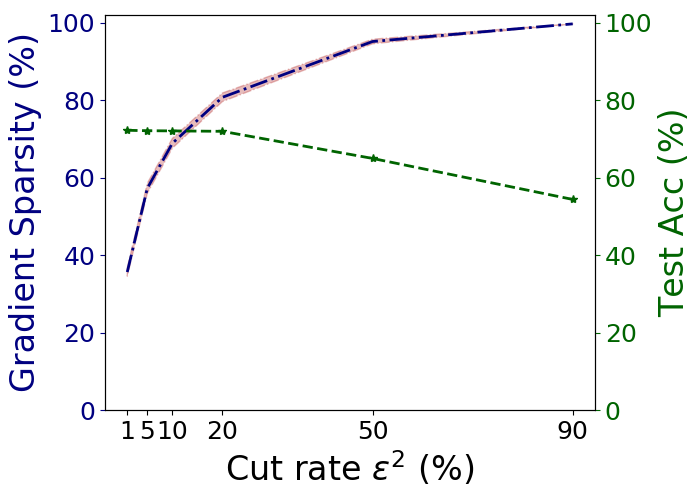}
    }   
    \caption{Gradient sparsity (orange) and the corresponding test accuracy in percentage (blue) versus cut rate $\varepsilon^2$ at a fixed noise level $\sigma$ for VGG-5 ($p=3,646,154$) ((a)-(b)) and ResNet-18 ($p=11,173,962$) ((c)-(d)). Solid lines represent the mean values of gradient sparsity averaged over training epochs of 5 independent runs and shaded bands indicate the $95\%$ confidence intervals. For a moderate cut rate, i.e., $\varepsilon^2 \leq 50\%$, we can drop more than 90\% of coordinates without significantly degrading the test accuracy.
    }
    \label{fig:lazy_sgd_sparse}
\end{figure}

\begin{table*}[t]
\caption{Gradient sparsity (\%) and test accuracy (\%) of NT-SGD for different cut rates $\varepsilon^2$. With over 70\% of gradient components been dropped to zeros, T-SGD and NT-SGD achieve similar or even slightly better test accuracy compared with vanilla SGD.}
\label{table:sparse_test_acc}
\centering
\resizebox{0.99\textwidth}{!}{
\begin{tabular}{c|c|c|c|c|c|c|c|c|c}
\toprule
\multirow{4}{*}{Data} &{Vanilla SGD} & \multicolumn{2}{c|}{\makecell{T-SGD\\$\varepsilon^2=10\%$}} & \multicolumn{2}{c|}{\makecell{NT-SGD\\$\varepsilon^2=10\%$}}&  \multicolumn{2}{c|}{\makecell{T-SGD\\$\varepsilon^2=20\%$}} & \multicolumn{2}{c}{\makecell{NT-SGD\\$\varepsilon^2=20\%$}}\\
\hhline{~---------}
 & Test Acc. & \makecell{Gradient\\Sparsity} & Test Acc.&\makecell{Gradient\\Sparsity} & Test Acc.&\makecell{Gradient\\Sparsity} & Test Acc.&\makecell{Gradient\\Sparsity} & Test Acc.\\
\hline
MNIST & 99.41 & 76.60 & 99.44& 76.66 & 99.39& {96.77} & 99.34&{96.72} & 99.37\\
\hline
Fashion-MNIST & 93.82& 73.67 & 93.68& 75.52 & 93.02&83.39& 93.80& 85.18 & 93.29\\
\hline
CIFAR-10 & 92.82 & 86.10& 92.75& 85.68 & {92.83}& {91.96} & 92.34& {91.52} & 92.73\\
\hline
CIFAR-100 &  72.11& 71.11 & {72.24}& 71.27 & 71.96 &83.37 & 71.72 & 83.22 & 72.07\\
\bottomrule
\end{tabular}}
\end{table*}

\begin{figure}[t]
    \centering
    \includegraphics[width=.85\textwidth]{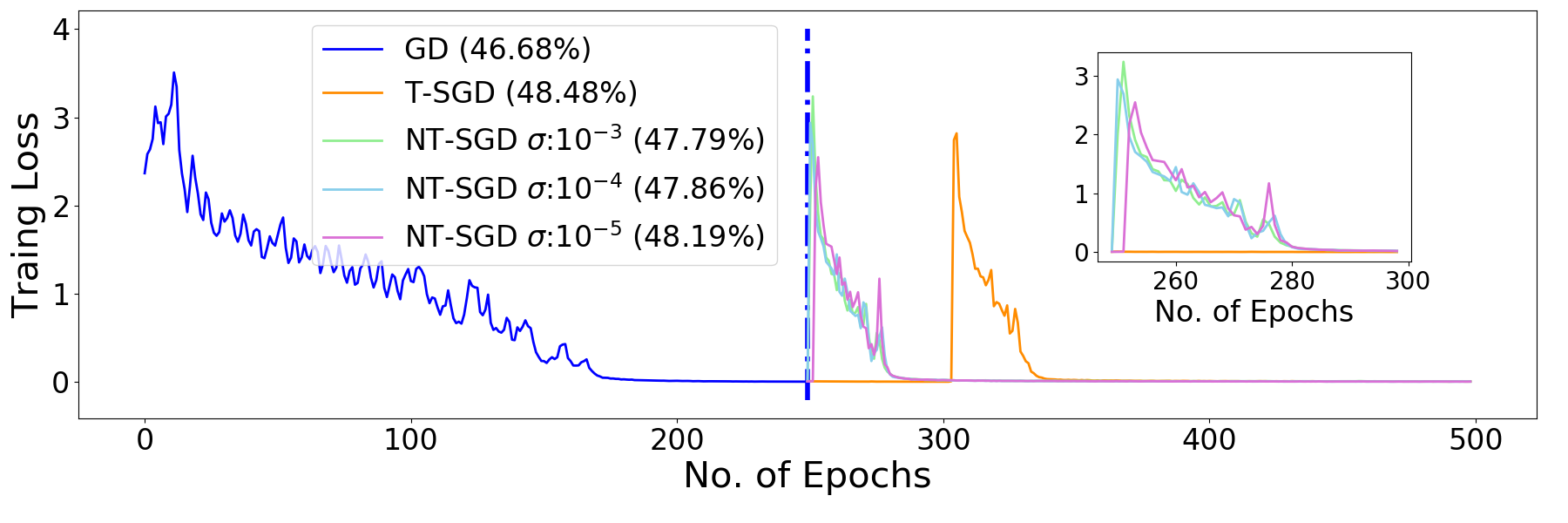}
    \caption{Dynamics of NT-SGD initialized at $\theta_{GD}^*$ found by GD, for ResNet-18 on a small subset of CIFAR-10 dataset ($n=1,000$). Injecting a small amount of noise can help NT-SGD effectively escape from the sharp minimum $\theta_{GD}^*$, and the more noise we add, the faster NT-SGD escapes. T-SGD with no injected noise is the last to escape (the orange peak around epoch 300). The final test accuracy of new minima found by NT-SGD improves up to $2\%$.} 
     \label{fig:escape_train}
\end{figure}

{\bf Convergence and generalization.} Figures~\ref{fig:lazy_sgd_train_sigma} and \ref{fig:lazy_sgd_train_eps} show the training dynamic of NT-SGD under a fixed noise level $\sigma$ and under a fixed cut rate $\varepsilon^2$ respectively. The joint selection of $\varepsilon^2$ and $\sigma$ controls the optimization and generalization behavior of NT-SGD. With suitable selected  $\varepsilon^2$ and $\sigma$, i.e., $\varepsilon^2 \leq 50\%$, NT-SGD matches the training dynamics exhibited by vanilla SGD and finds minima that generalize as well as vanilla SGD. Such observations verify our claim in Theorem~\ref{theo:opt} that NT-SGD can achieve the same rate of convergence as vanilla SGD. However, an extremely large $\varepsilon^2$ can damage the performance as too many gradient components have been truncated, e.g., in Figure~\ref{fig:lazy_sgd_train_eps}, NT-SGD with $\varepsilon^2 = 90\%$ (the corresponding gradient sparsity\footnote{Gradient sparsity (\%) is the ratio of the number of coordinates been truncated to the full gradient dimension.} $\geq 95\%$) has low test accuracy. Additional results on other choices of noise level $\sigma$ and cut rate $\varepsilon^2$ can be found in Appendix~\ref{sec:app_exp}.

The influence of the cut rate $\varepsilon^2$ on the behaviors of NT-SGD (Figure \ref{fig:lazy_sgd_train_eps}) is two-fold: on one hand, large $\varepsilon^2$ makes NT-SGD more stable thus improves the generalization. This is due to the fact that, during training, $\varepsilon^2$ directly determines the actual cutting threshold $\kappa_{\varepsilon, t} $, thus also directly determines the gradient sparsity. As shown in Figure~\ref{fig:lazy_sgd_sparse}, the sparsity in gradients increases as we increase $\varepsilon^2$. NT-SGD with $\varepsilon^2 \leq 50\%$ can drop more than $90\%$ of gradient components to zeros and only suffers a minor performance degradation (Table~\ref{table:sparse_test_acc}). On the other hand, gradient truncation slows down the optimization (Figure \ref{fig:lazy_sgd_train_sigma}), since it introduces bias in the gradient.

Noise variance $\sigma^2$ also plays an important role when studying the behavior of NT-SGD. In general, a large $\sigma$ increases the optimization error (See \eqref{eq:ntsgd_optimization}), helps NT-SGD escape from saddle points (Theorem~\ref{theo:escape}), and also decreases the generalization error (Eq.~\eqref{eq:ntsgd_generalization}). A large $\sigma$ may negatively affect the generalization, hence the performance of NT-SGD with $\sigma = 0.1$ (the red line in Figure~\ref{fig:lazy_sgd_train_eps} (a)) suffers from a $5\%$ performance drop in the final test accuracy.

{\bf Escaping from sharp minima.} We follow the experimental design in \cite{zhu2019}, providing an empirical example to demonstrate that NT-SGD can escape from sharp minima effectively. We initialize NT-SGD from the sharp minimum $\theta^*_{GD}$ found by Gradient Descent (GD), use a constant cutting threshold $\kappa = 10^{-3}$, and compare the escaping behavior of NT-SGD at 4 different levels of noise $\sigma \in \{0.0,~10^{-3},~10^{-4},~10^{-5}\}$. Figure \ref{fig:escape_train} shows NT-SGD successfully escapes from $\theta^*_{GD}$, and the higher $\sigma$ is, the fewer epochs NT-SGD needs to escape from poor local minima. For example, T-SGD with no additional noise $\sigma = 0$ (orange line) takes the longest epochs ($\approx 300$) to escape from $\theta^*_{GD}$, and increasing $\sigma$ from $10^{-5}$ to $10^{-3}$ reduces the number of epochs for NT-SGD to escape from $\theta_{GD}^*$ (see zoomed-in view in Figure~\ref{fig:escape_train}). NT-SGD find minima that generalize better than the one found by GD, with approximately $2\%$ higher test accuracy (see the number provided within the parentheses in the legend). Such observations demonstrate that injecting a small amount of noise can help NT-SGD effectively escape from saddle points, and the more noise we add, the faster it escapes.

\label{sec:experiments}

\section{Conclusions}

In this paper, we analyze Truncated SGD (T-SGD), a sparse gradient algorithm that reduces small gradient components to zeros, and propose Noisy Truncated SGD (NT-SGD), a perturbed version of T-SGD that adds Gaussian noise to all components after gradient truncation. We establish the optimization rate of convergence for both T-SGD and NT-SGD, and prove that NT-SGD is capable of escaping from saddle points with the help of a small amount of injected Gaussian noise. We also derive generalization error bounds of T-SGD and NT-SGD based on uniform stability. We demonstrate that NT-SGD achieves better generalization error bound compared to T-SGD with considerably improved stability due to the noise. Empirical evidence demonstrates that both T-SGD and NT-SGD matches the speed and accuracy of vanilla SGD, and NT-SGD can successfully escape sharp minima, which support our theoretical analysis. 


\label{sec:conclusion}

\section*{Acknowledgment}
The research was supported by NSF grants IIS-1908104, OAC-1934634, IIS-1563950, IIS-1447566, IIS-1447574, IIS-1422557, CCF-1451986. The authors would like to thank Minnesota Supercomputing Institute (MSI) at the University of Minnesota for providing the computing support.

\bibliography{references}

\newpage
\appendix
\section{Additional Experimental Results}
\label{sec:app_exp}
In this section, we first describe the datasets. Then, we present additional experimental results discussed in Section~\ref{sec:experiments}.

\noindent {\bf MNIST dataset:} 60,000 black and white training images, including handwritten digits 0 to 9. We use a subset of MNIST with $n = 10,000$ data points where 1,000 samples from each class are randomly selected. Each image of size $28 \times 28$ is first re-scaled into [0,1] by dividing each pixel value by 255, then z-scored by subtracting the mean and dividing the standard deviation of the training set.

\noindent {\bf Fashion-MNIST dataset:} 60,000 gray-scale training images and 10,000 test images, including 10 clothing categories such as shirts, dresses, sandals, etc. Each image of size $28 \times 28$ is first re-scaled into [0,1] by dividing each pixel value by 255, then z-scored by subtracting the mean and dividing the standard deviation of the training set.

\noindent {\bf CIFAR-10 dataset:} 60,000 color images consisting of 10 categories, e.g., airplane, cat, dog etc. The training set includes 50,000 images while the test set contains the rest 10,000 images. Each image of size $32 \times 32$ has 3 color channels. We first re-scale each image into [0, 1] by dividing each pixel value by 255, then each image is normalized by subtracting the mean and dividing the standard deviation of the training set for each color channel. We also use \textit{RandomCrop} and \textit{RandomHorizontalFlip} for data augmentation.

\noindent {\bf CIFAR-100 dataset:} 60,000 color images consisting of 100 categories, e.g., airplane, cat, dog etc. The training set includes 50,000 images while the test set contains 10,000 images. Each image of size $32 \times 32$ has 3 color channels. We first re-scale each image into [0, 1] by dividing each pixel value by 255, then each image is normalized by subtracting the mean and dividing the standard deviation of the training set for each color channel.  We also use \textit{RandomCrop} and \textit{RandomHorizontalFlip} for data augmentation.

\begin{figure*}[t]
    \centering
    \subfigure[VGG-5, MNIST, $\sigma = 0.1$]{\includegraphics[width=0.48\textwidth]{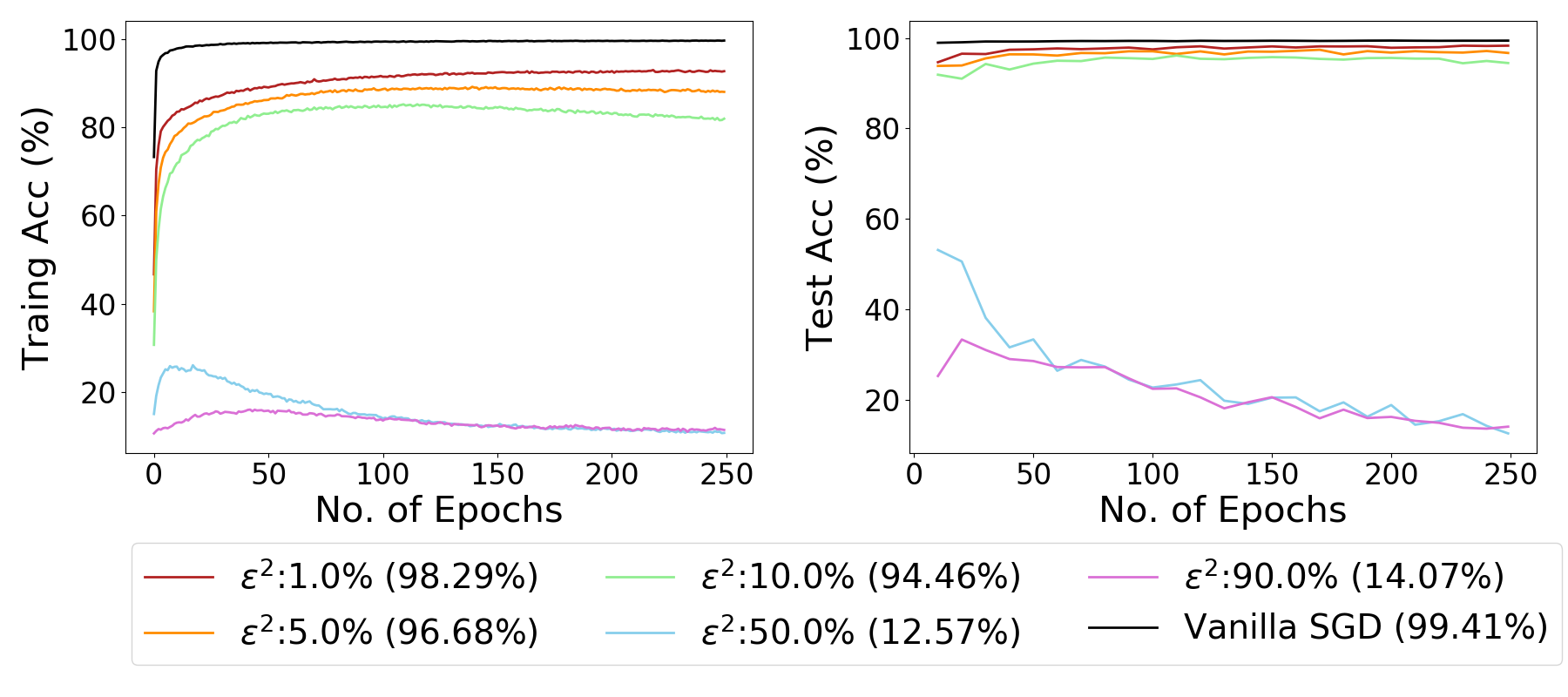}}
    \subfigure[VGG-5, MNIST, $\sigma = 10^{-5}$]{\includegraphics[width=0.48\textwidth]{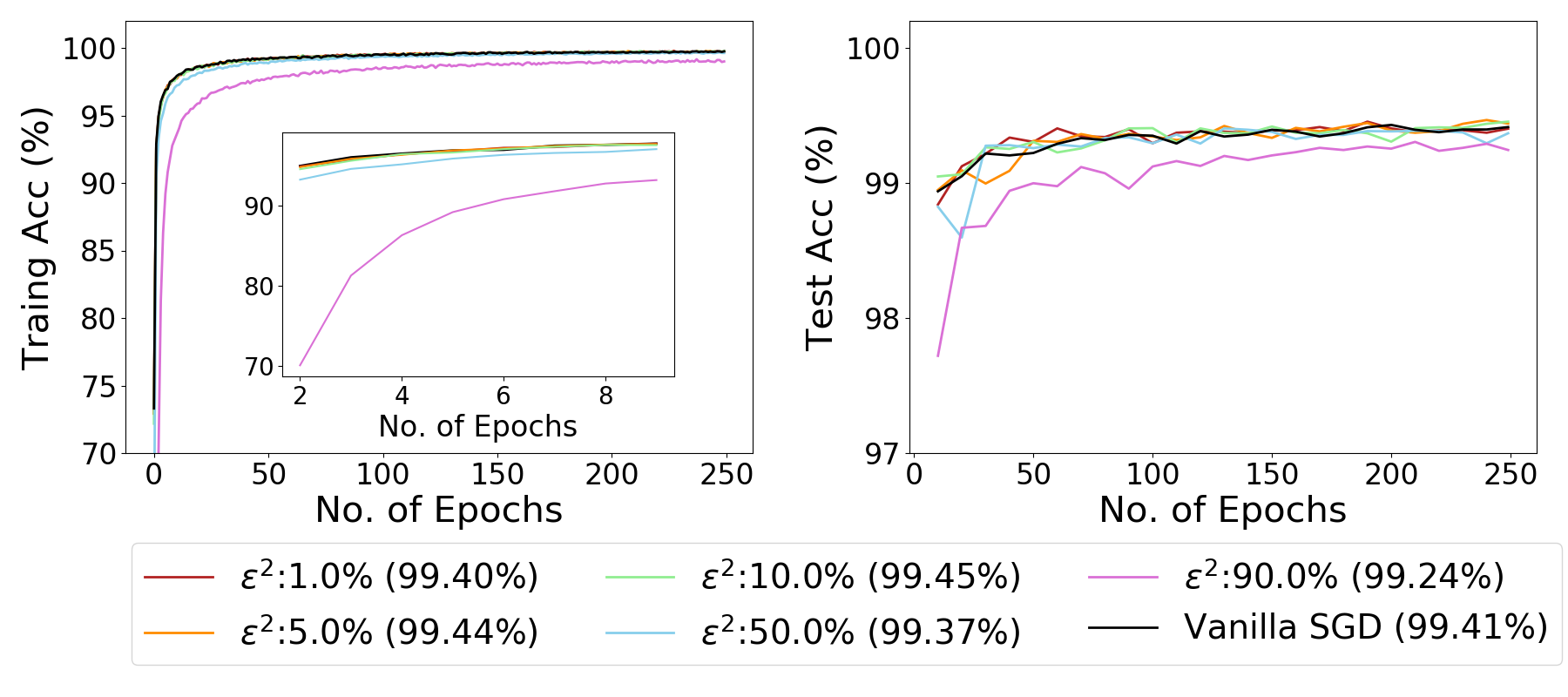}}
    \subfigure[VGG-5, MNIST, $\sigma = 0.1$]{\includegraphics[width=0.48\textwidth]{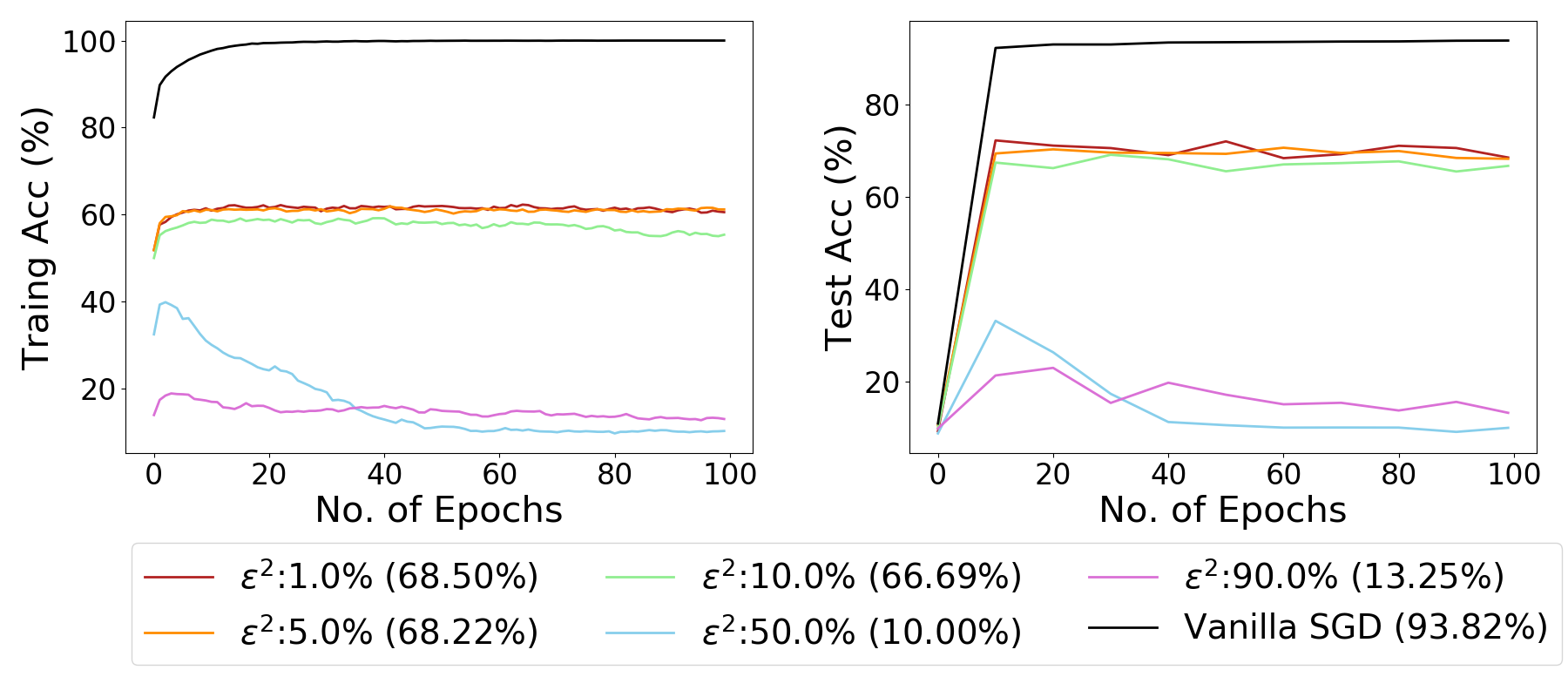}}
    \subfigure[VGG-5, MNIST, $\sigma = 10^{-5}$]{\includegraphics[width=0.48\textwidth]{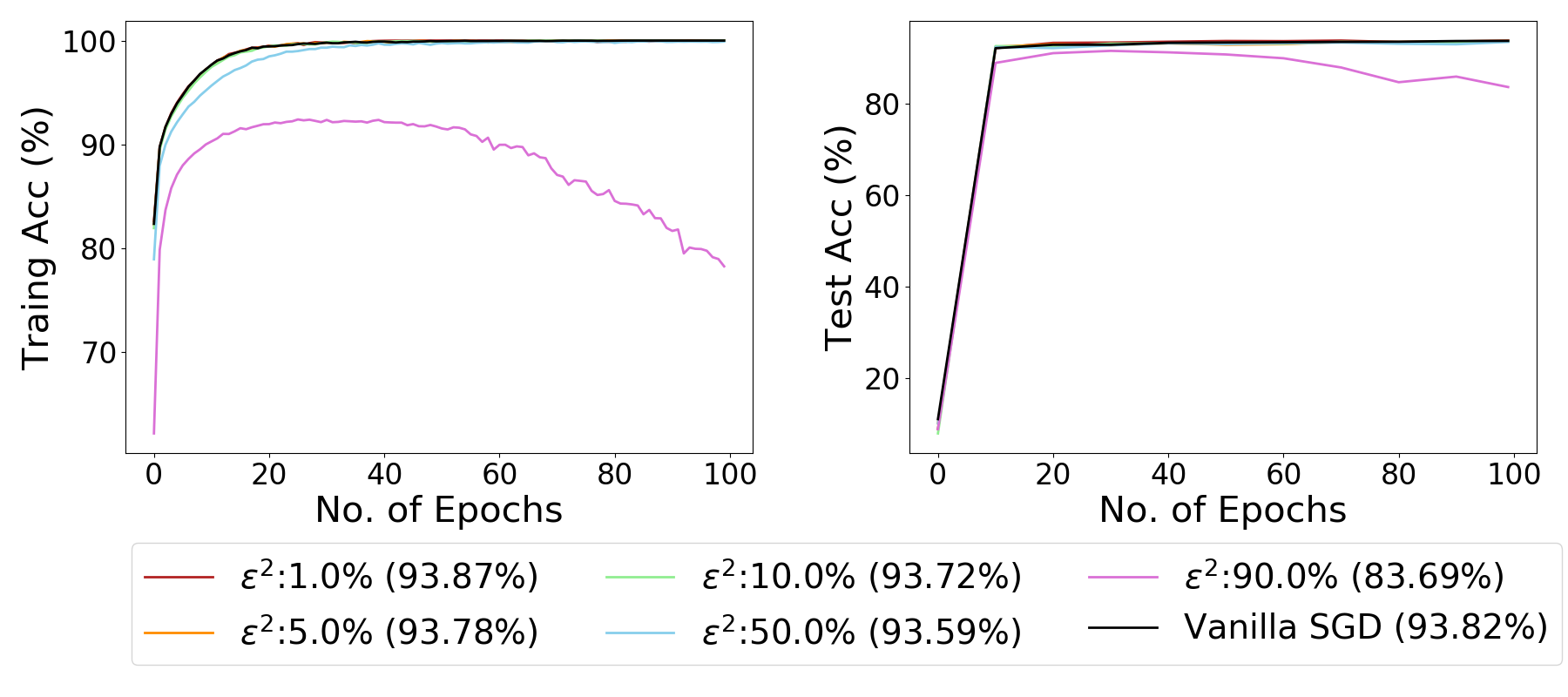}}
    \subfigure[ResNet-18, CIFAR-10, $\sigma = 10^{-2}$]{\includegraphics[width=0.48\textwidth]{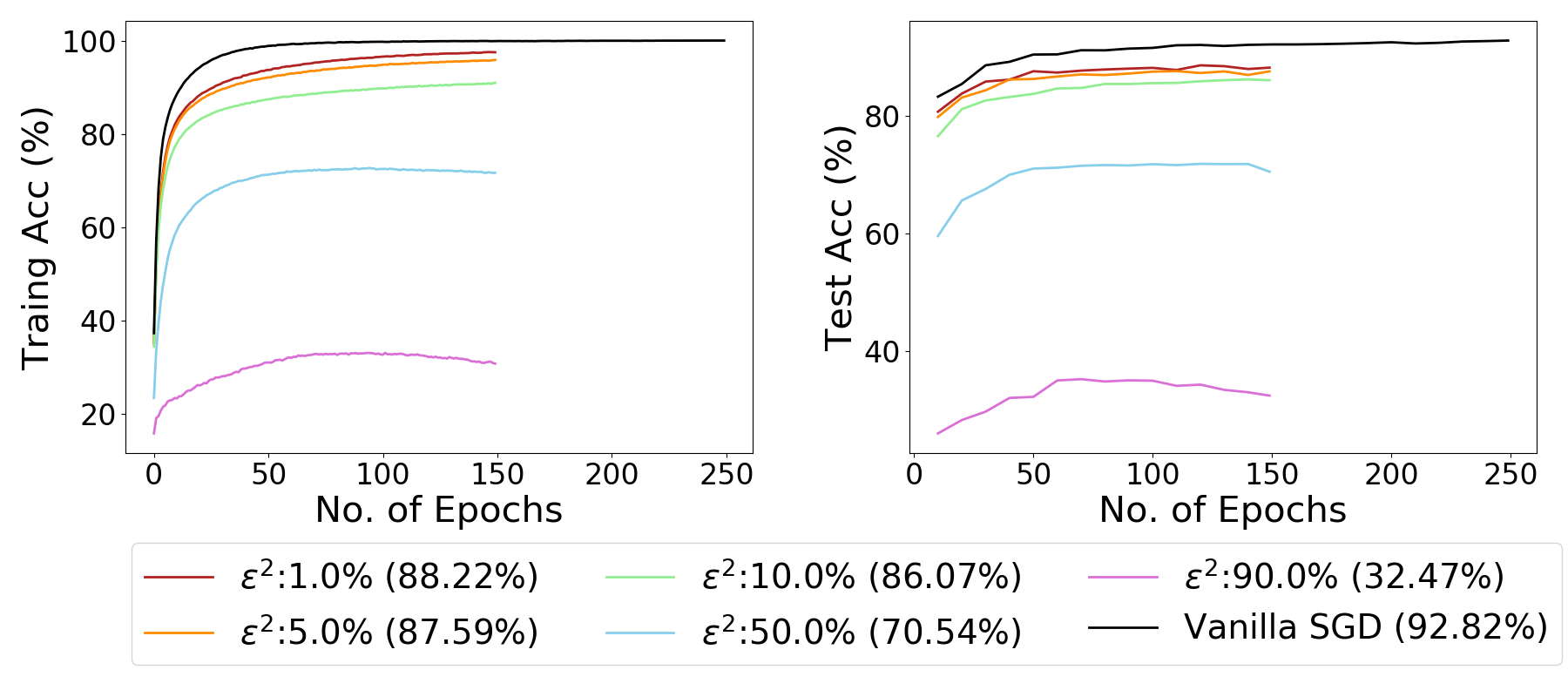}}
    \subfigure[ResNet-18, CIFAR-10, $\sigma = 10^{-4}$]{\includegraphics[width=0.48\textwidth]{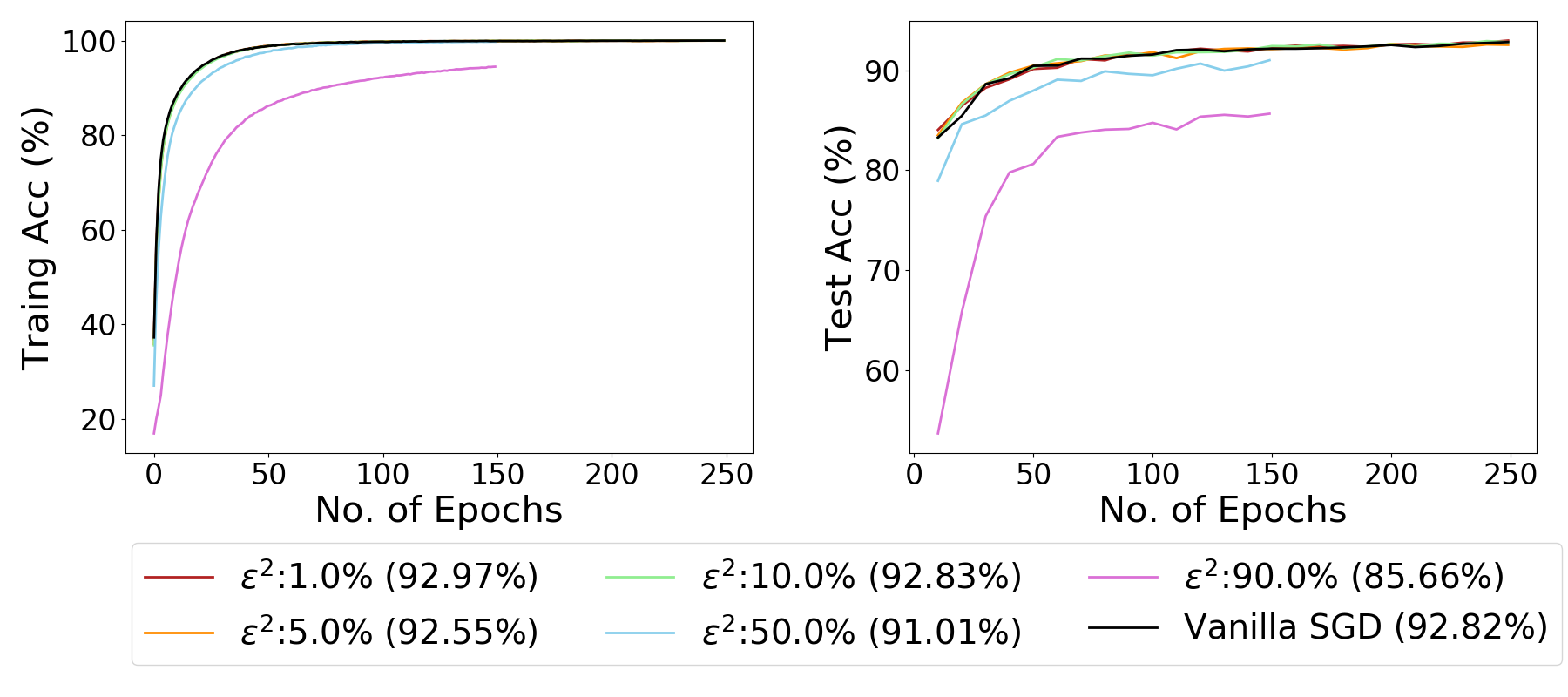}}
    \subfigure[ResNet-18, CIFAR-100, $\sigma = 10^{-2}$]{\includegraphics[width=0.48\textwidth]{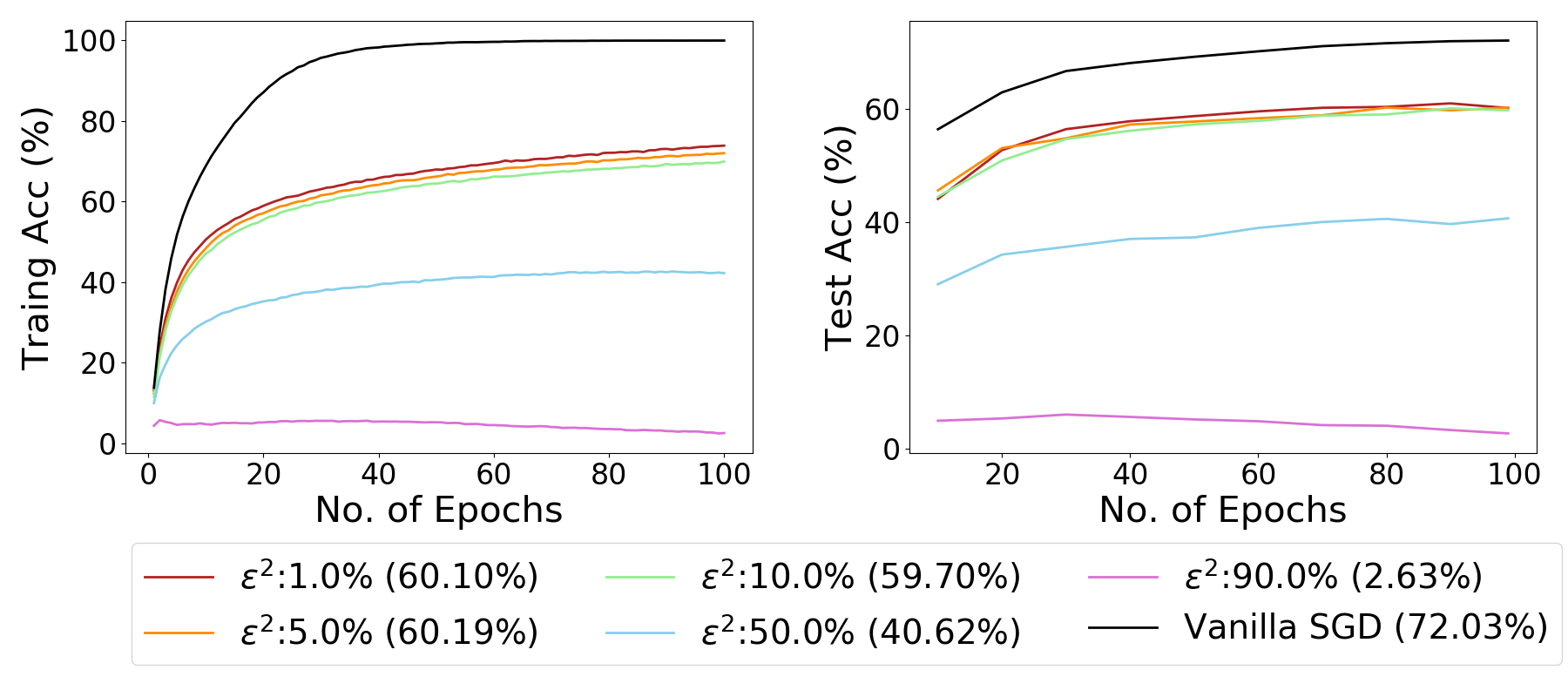}}
    \subfigure[ResNet-18, CIFAR-100, $\sigma = 10^{-5}$]{\includegraphics[width=0.48\textwidth]{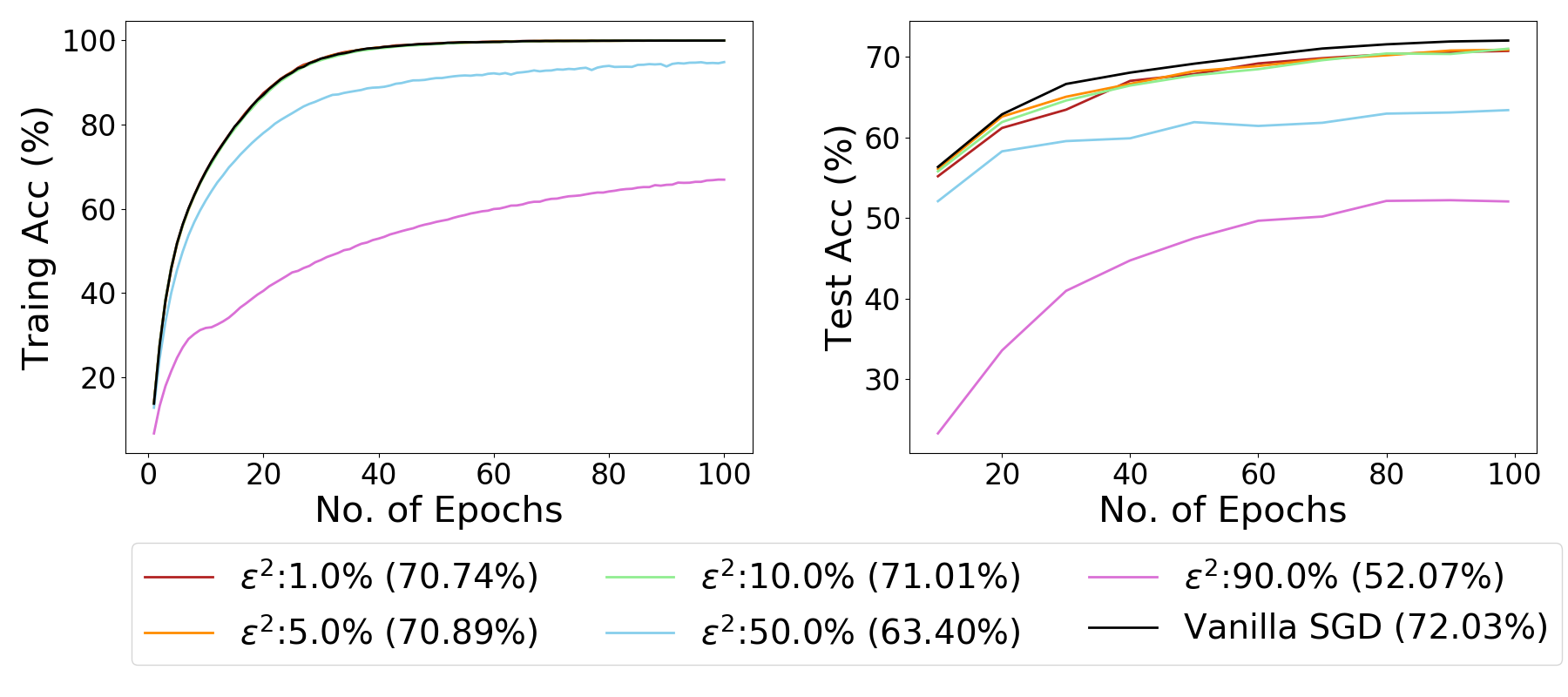}}
    \caption{Training and test dynamics of NT-SGD and Vanilla SGD (black line) as we vary cut rates $\varepsilon^2$ at fixed noise level for MNIST ((a)-(b)), Fashion-MNIST ((c)-(d)), CIFAR-10 ((e)-(f)), and CIFAR-100 ((g)-(h)). The X-axis is the number of epochs, and the Y-axis is the train/test accuracy. Legends indicate the choice of $\varepsilon^2$ and the number within the parentheses represents the corresponding test accuracy at last epoch. The inset plot provides a zoom-in view of what happened at ``elbow''.  With suitable selected  $\varepsilon^2$ and $\sigma$, i.e., $\varepsilon^2 \leq 50\%$, NT-SGD shows comparable performance to vanilla SGD and finds minima that generalize as well as vanilla SGD. However, in the extreme case, the large injected noise ($\sigma=0.1$) causes NT-SGD to settle down at different local minima, depending on the selection of cut rate $\varepsilon^2$. In general, appropriate cut rate, i.e., $\varepsilon^2\leq50\%$, generalizes well while the extreme high value of $\varepsilon^2 = 90\%$ harms the generalization.
    } 
    \label{fig:lazy_sgd_eps_app}
\end{figure*}

\noindent  \textbf{Additional results. } Figure~\ref{fig:lazy_sgd_eps_app} presents the training and test dynamics of NT-SGD and Vanilla SGD under different cut rates $\varepsilon^2$ with a fixed noise variance $\sigma$ for MNIST ((a)-(b)), Fashion-MNST ((c)-(d)), CIFAR-10 ((e)-(f)), and CIFAR-100 ((g)-(h)). 
The range of different $\varepsilon^2$ we considering is $\varepsilon^2 \in \{1.0\%, 5.0\%, 10.0\%, 50\%, 90\%\}$. When $\sigma$ is small, i.e., $\sigma = 10^{-4}$ or $10^{-5}$ ( Figure~\ref{fig:lazy_sgd_eps_app} (b), (d), (f), and (h)), NT-SGD with an appropriate cut rate, i.e., $\varepsilon^2\leq 50\%$ performs similarly to Vanilla SGD. However, for a very large $\sigma$, i.e., $\sigma = 0.1$ (Figure~\ref{fig:lazy_sgd_eps_app} (a) and (c)) or $\sigma = 0.01$ (Figure~\ref{fig:lazy_sgd_eps_app} (e) and (g)), there is a performance degradation for various $\varepsilon^2$. This observation is consistent with the theoretical bound in Theorem \ref{theo:opt} that the error rate increases as the total noise variance $R = p\sigma^2$ increases.  

\begin{figure*}[h!]
    \centering    
    \subfigure[ResNet-18, CIFAR-10, $\varepsilon^2 = 1\%$]{\includegraphics[width=0.48\textwidth]{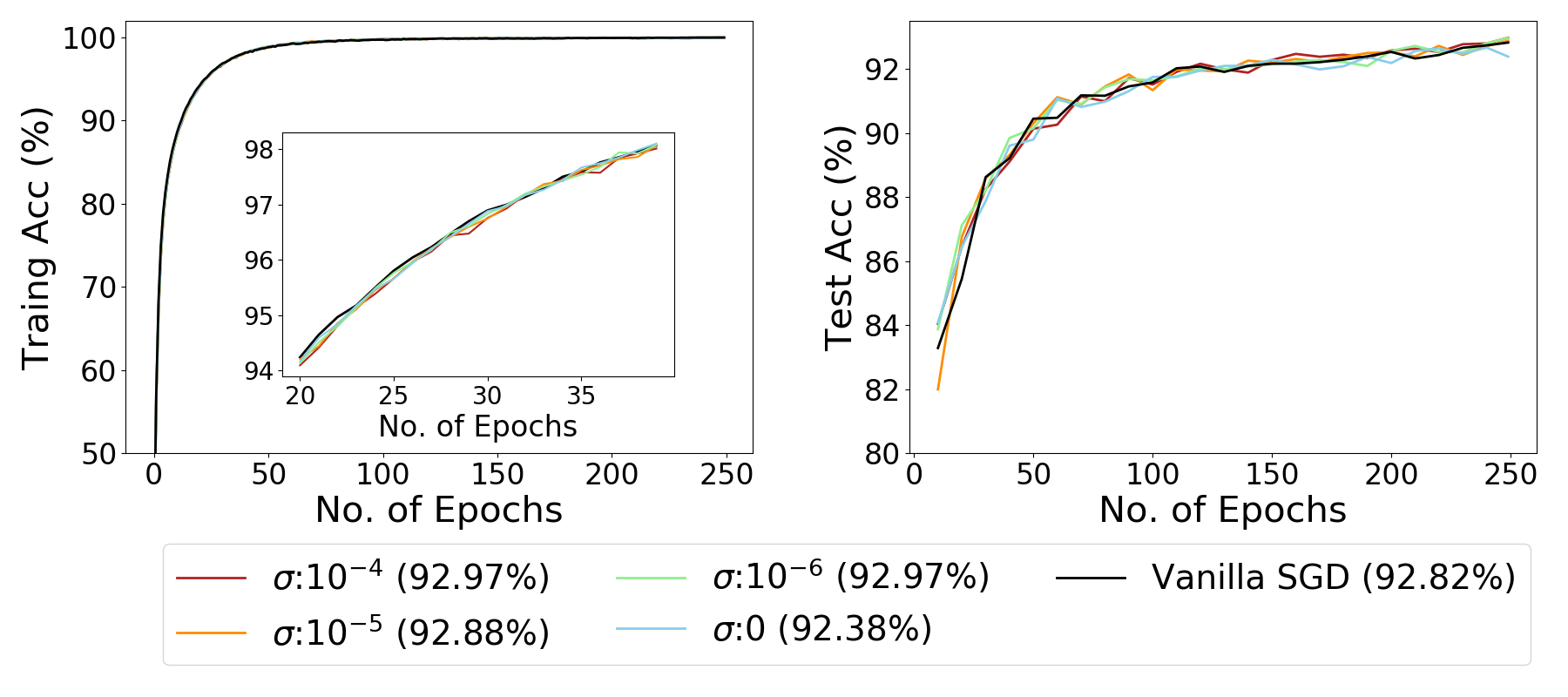}}
    \subfigure[ResNet-18, CIFAR-10, $\varepsilon^2 = 5\%$]{\includegraphics[width=0.48\textwidth]{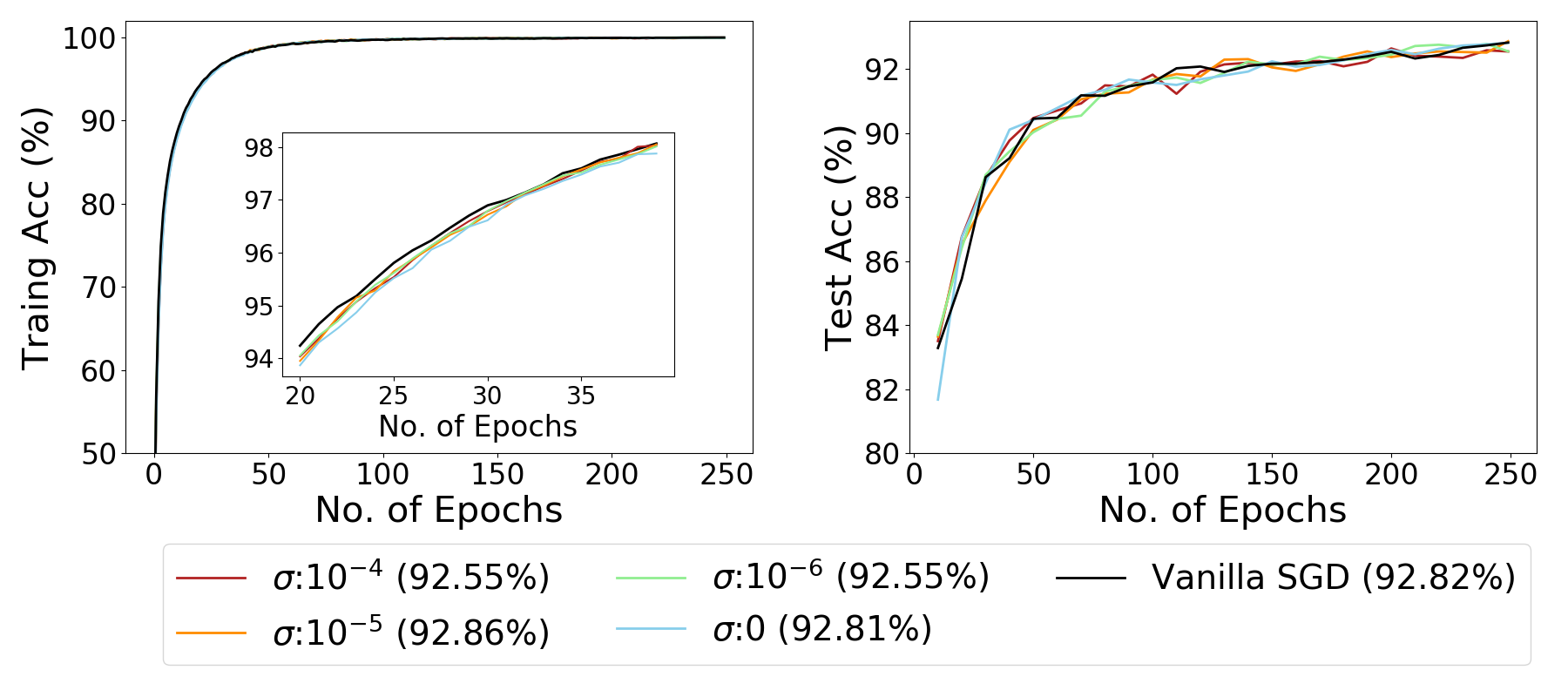}}
    \subfigure[ResNet-18, CIFAR-10, $\varepsilon^2 = 20\%$]{\includegraphics[width=0.48\textwidth]{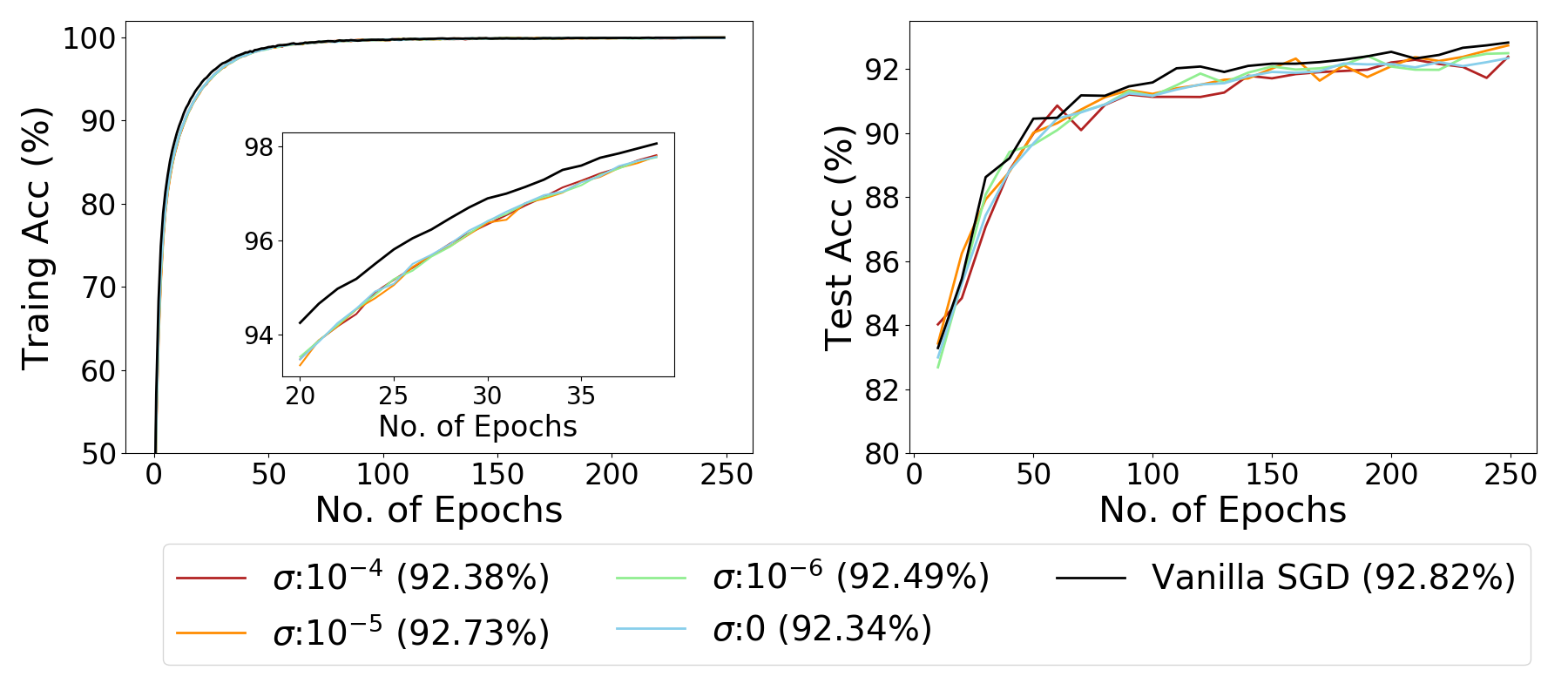}}
    \subfigure[ResNet-18, CIFAR-10, $\varepsilon^2 = 90\%$]{\includegraphics[width=0.48\textwidth]{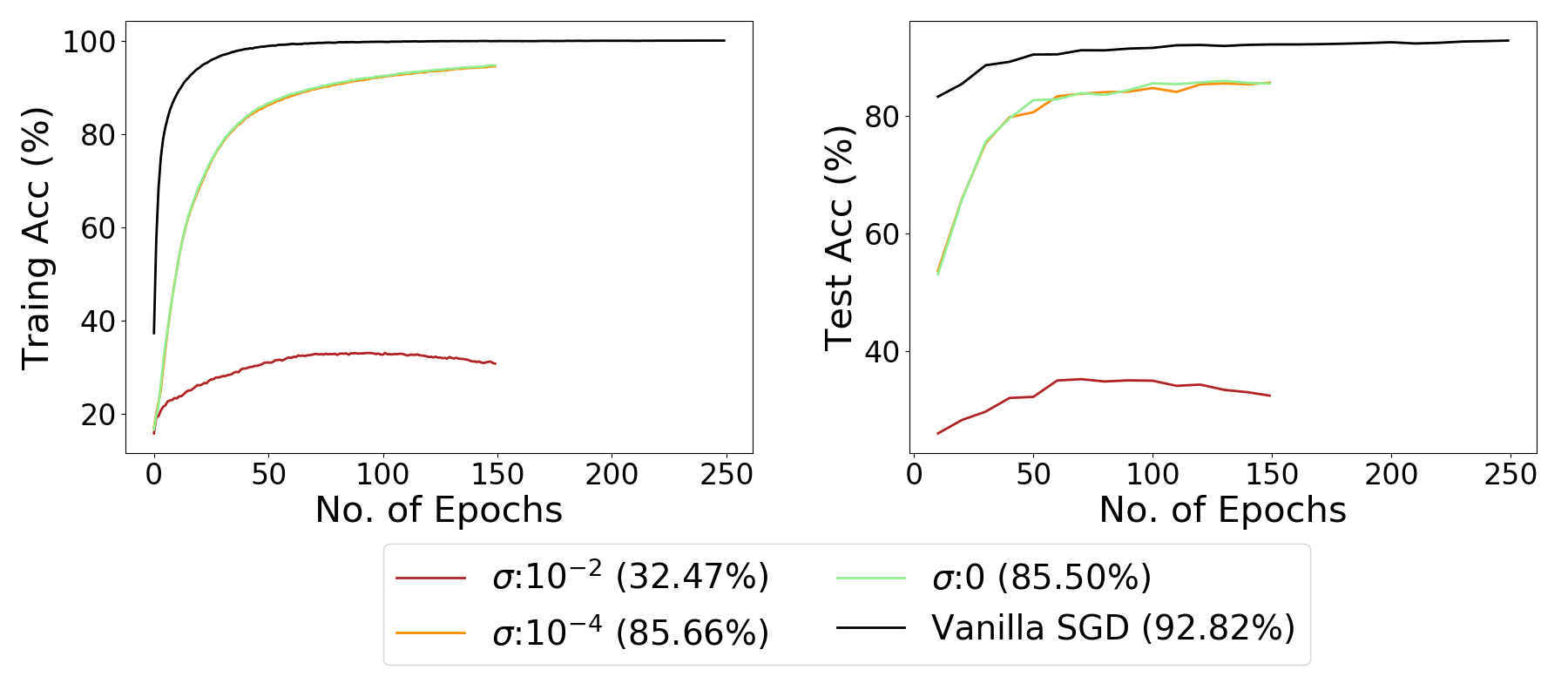}}
    \subfigure[ResNet-18, CIFAR-100, $\varepsilon^2 = 1\%$]{\includegraphics[width=0.48\textwidth]{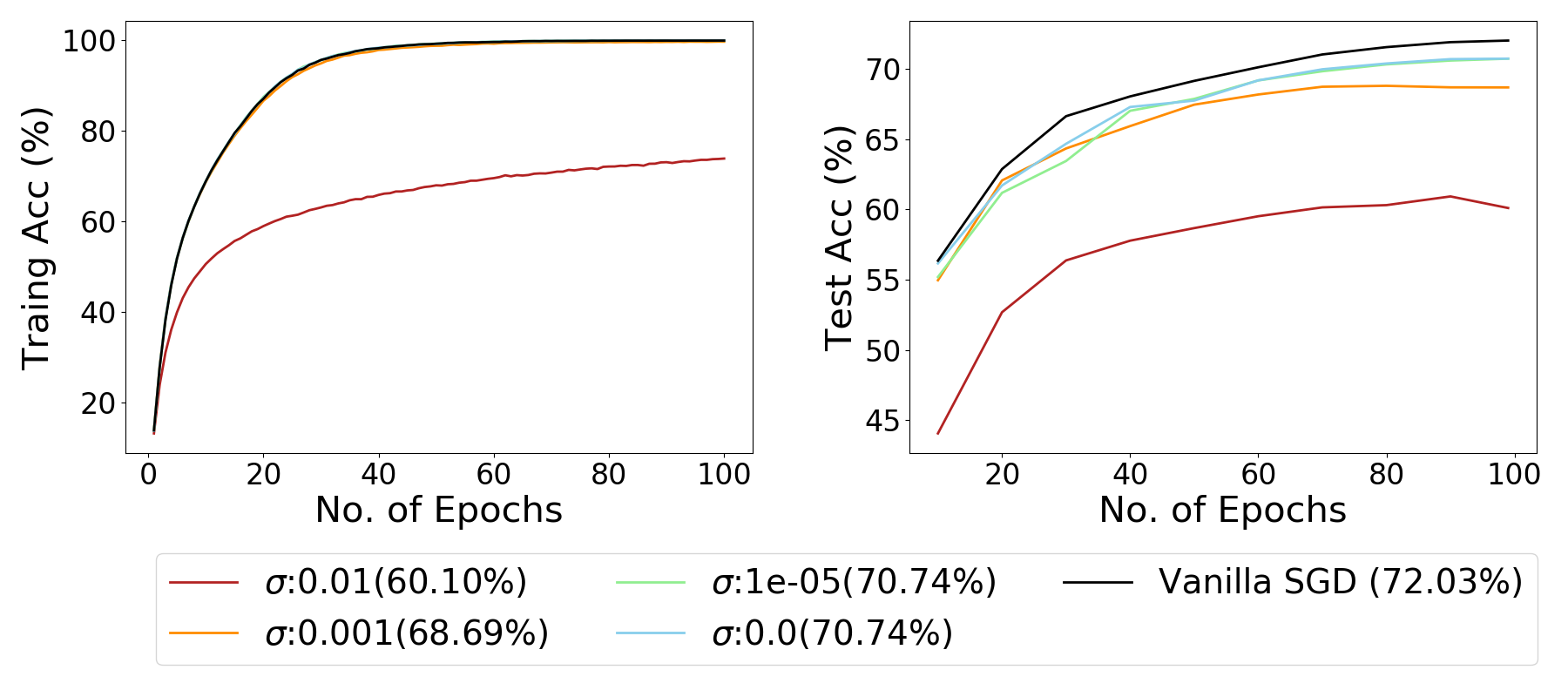}}    
    \subfigure[ResNet-18, CIFAR-100, $\varepsilon^2 = 5\%$]{\includegraphics[width=0.48\textwidth]{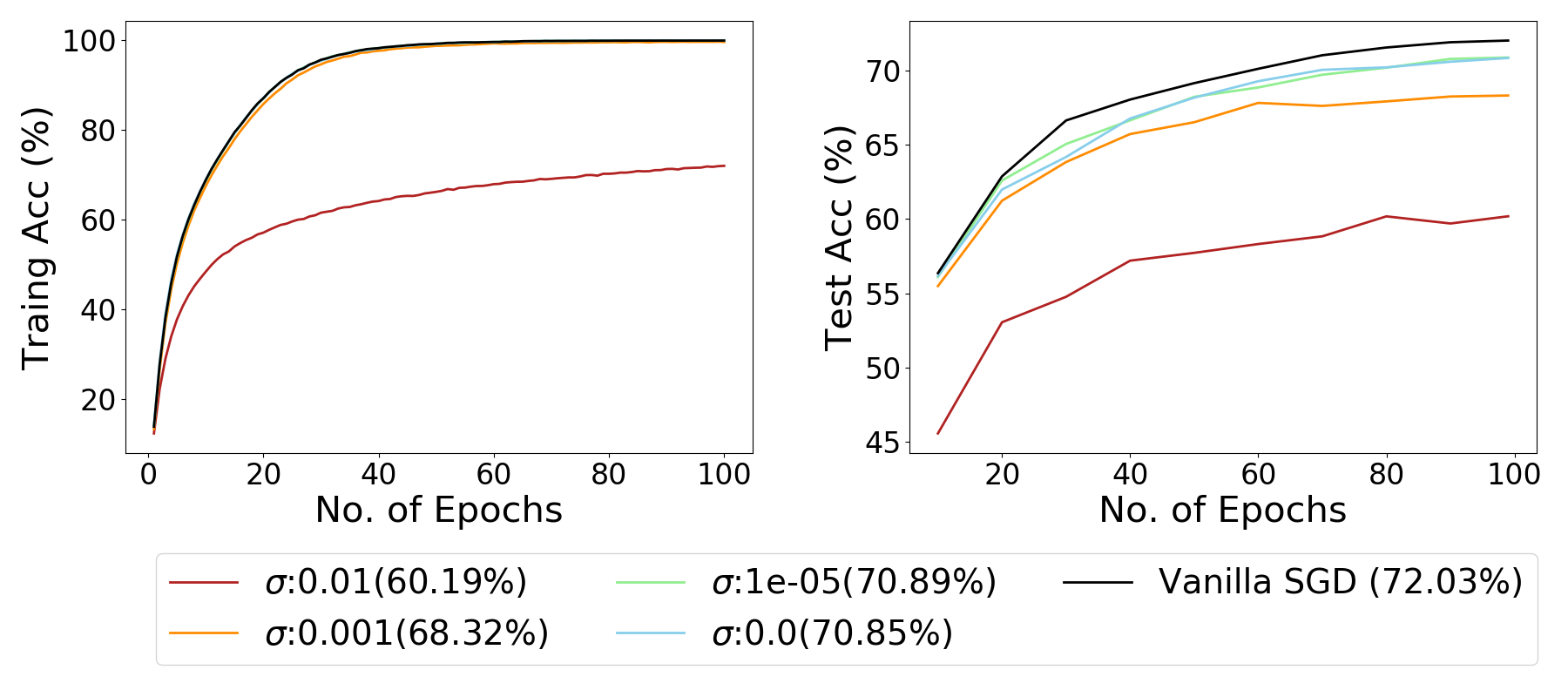}} 
    \subfigure[ResNet-18, CIFAR-100, $\varepsilon^2 = 50\%$]{\includegraphics[width=0.48\textwidth]{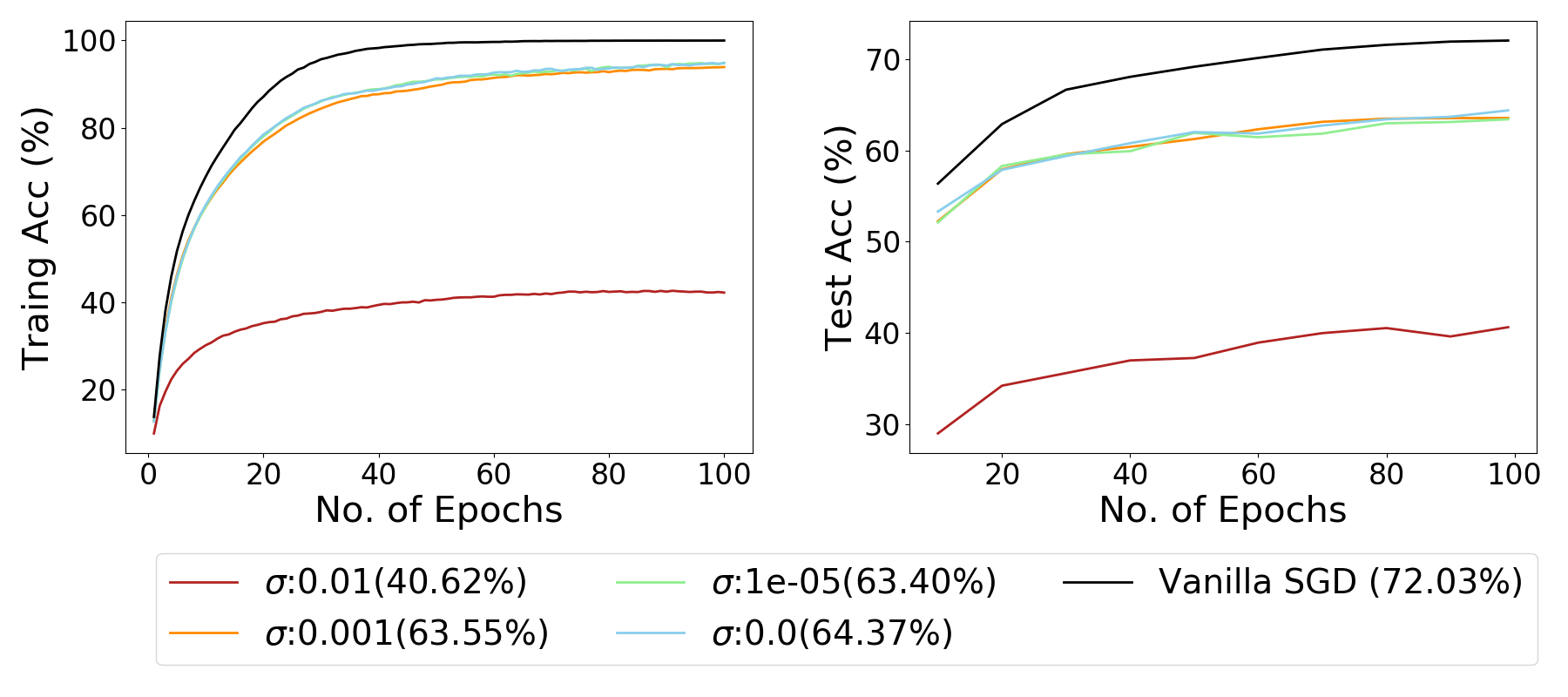}} 
    \subfigure[ResNet-18, CIFAR-100, $\varepsilon^2 = 90\%$]{\includegraphics[width=0.48\textwidth]{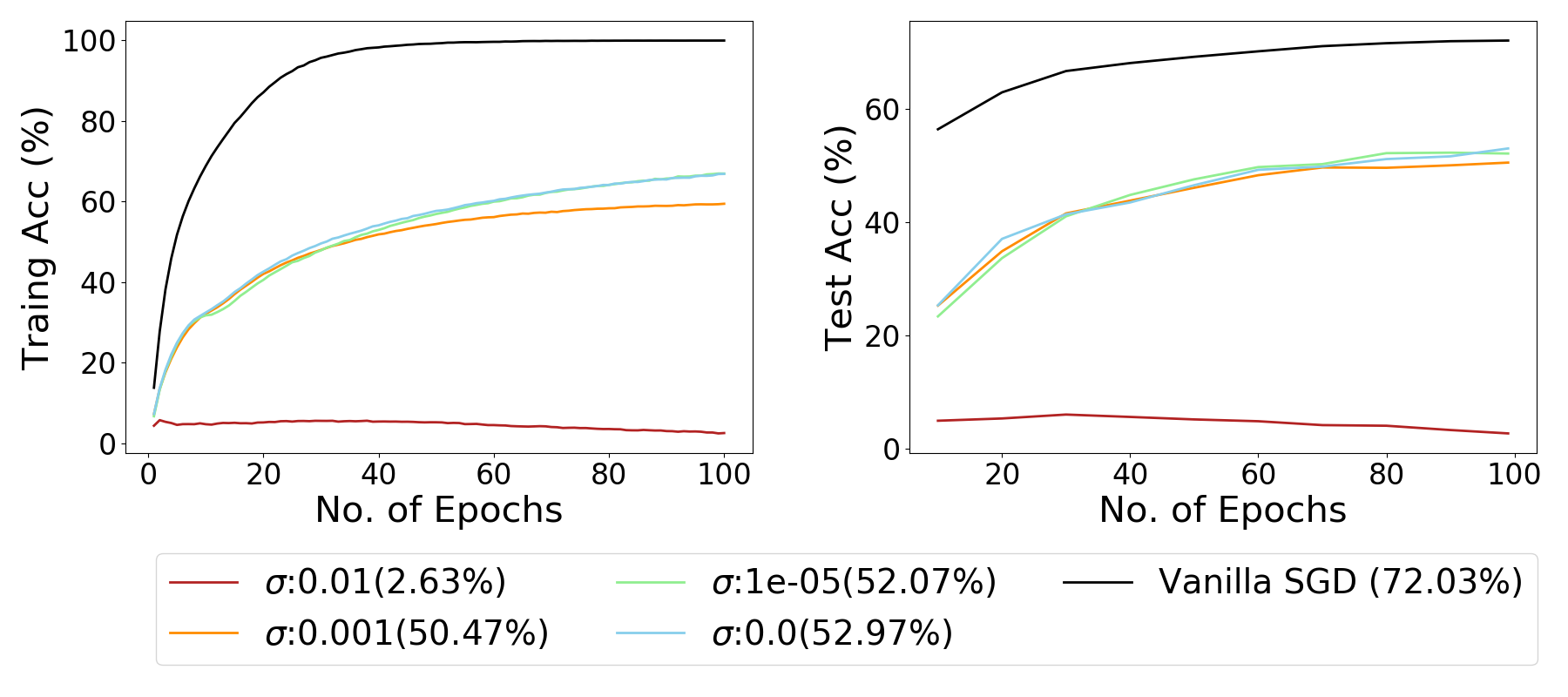}} 
    \caption{ Training and test dynamics of T-SGD ($\sigma=0$), NT-SGD, and Vanilla SGD (black line) with different noise levels $\sigma$ for MNIST ((a)-(d)), Fashion-MNIST ((e)-(h)), CIFAR-10 ((i)-(l)), and CIFAR-100 ((m)-(p)) with $\varepsilon^2 \in \{1\%,~5\%,~50\%,~90\%\}$. The X-axis is the number of epochs, and the Y-axis is the train(test) accuracy. 
    Legends indicate the choice of $\sigma$ and the number within the parentheses represents the corresponding test accuracy at last epoch. The inset plot provides a zoom-in view of what happened at ``elbow''. Both T-SGD and NT-SGD with small amount of noise can match the performance of SGD. However, T-SGD and NT-SGD may generalize poorly with large injected noise, e.g., $\sigma=0.1$.}
    \label{fig:lazy_sgd_sigma_app}
\end{figure*}

Figure~\ref{fig:lazy_sgd_sigma_app} presents the training and test dynamics of NT-SGD and Vanilla SGD under different noise variance $\sigma$ with a fixed cut rate $\varepsilon^2$. In particular, we show results for $\varepsilon^2 \in \{1\%,~5\%,~50\%,~90\%\}$. The range of $\sigma$ we considering is $\sigma \in \{10^{-1}, 10^{-3}, 10^{-5}, 0\}$ for MNIST and Fashion-MNIST, $\sigma \in \{10^{-4}, 10^{-5}, 10^{-6}, 0\}$ for CIFAR-10, and $\sigma \in \{10^{-2}, 10^{-3}, 10^{-5}, 0\}$ for CIFAR-100. The results shows that NT-SGD with small amount of noise can match the performance of SGD. However, NT-SGD may generalize poorly with large injected noise, e.g., $\sigma=0.1$ in Figure~\ref{fig:lazy_sgd_sigma_app} (a)-(d).


\begin{table*}[t]
\caption{Gradient sparsity (\%) and test accuracy (\%) of NT-SGD for selected $\varepsilon^2$. With over 70\% (sometimes even more than 90\%, see numbers in bold) of gradient components been dropped to zeros, T-SGD and NT-SGD achieve similar or even slightly better test accuracy compared with vanilla SGD.}
\label{table:sparse_test_acc_full}
\centering
\resizebox{0.99\textwidth}{!}{
\begin{tabular}{c|c|c|c|c|c|c|c|c|c}
\toprule
\multirow{4}{*}{Data} &{Vanilla SGD} & \multicolumn{2}{c|}{\makecell{T-SGD\\$\varepsilon^2=10\%$}} & \multicolumn{2}{c|}{\makecell{NT-SGD\\$\varepsilon^2=10\%$}}&  \multicolumn{2}{c|}{\makecell{T-SGD\\$\varepsilon^2=20\%$}} & \multicolumn{2}{c}{\makecell{NT-SGD\\$\varepsilon^2=20\%$}}\\
\cmidrule(r){2-10}
 & Test Acc. & \makecell{Gradient\\Sparsity} & Test Acc.&\makecell{Gradient\\Sparsity} & Test Acc.&\makecell{Gradient\\Sparsity} & Test Acc.&\makecell{Gradient\\Sparsity} & Test Acc.\\
\midrule
MNIST & 99.41(0.04) & 76.60(0.29) & {99.44(0.05)}& 76.66(0.43) & 99.39(0.09)& \textbf{96.77(0.05)} & 99.34(0.13)& \textbf{96.72(0.10)} & 99.37(0.08)\\
\midrule
Fashion-MNIST & 93.82(0.15)& 73.67(0.83) & 93.68(0.10)& 75.52(0.20) & 93.02(0.27)&83.39(0.15) & 93.80(0.09)& 85.18(0.25) & 93.29(0.12)\\
\midrule
CIFAR-10 & 92.82(0.14) & 86.10(0.70) & 92.75(0.33)& 85.68(0.69) & {92.83(0.28)}&\textbf{91.96(0.33)} & 92.34(0.46)& \textbf{91.52(0.37)} & 92.73(0.21)\\
\midrule
CIFAR-100 &  72.11(0.26)& 71.11(0.20) & 72.24(0.24)& 
71.27(0.16) & 71.96(0.40) &83.37(0.15) & 71.72(0.22) & 83.22(0.06) & 72.07(0.14)\\
\bottomrule
\toprule
\multirow{4}{*}{Data} &{Vanilla SGD} & \multicolumn{2}{c|}{\makecell{T-SGD\\$\varepsilon^2=50\%$}} & \multicolumn{2}{c|}{\makecell{NT-SGD\\$\varepsilon^2=50\%$}}&  \multicolumn{2}{c|}{\makecell{T-SGD\\$\varepsilon^2=90\%$}} & \multicolumn{2}{c}{\makecell{NT-SGD\\$\varepsilon^2=90\%$}}\\
\cmidrule(r){2-10}
& Test Acc. & \makecell{Gradient\\Sparsity} & Test Acc.&\makecell{Gradient\\Sparsity} & Test Acc.&\makecell{Gradient\\Sparsity} & Test Acc.&\makecell{Gradient\\Sparsity} & Test Acc.\\
\midrule
MNIST &99.41(0.04) & \textbf{96.77(0.05)} & 99.34(0.13)& \textbf{96.87(0.08)} & 99.37(0.08) & \textbf{99.88(0.01)} & 99.23(0.04) & \textbf{99.88(0.01)} & 99.24(0.08)\\
\midrule
Fashion-MNIST& 93.82(0.15) & \textbf{95.62(0.08)} & 93.32(0.27)& \textbf{95.72(0.04)} & 93.40(0.26)& 99.99(0.00) & 62.34(7.81) & 99.99(0.00) & 81.58(3.44)\\
\midrule
CIFAR-10 &92.82(0.14)& 97.90(0.12) & 90.58(0.69)& \textbf{97.86(0.13)} & 91.01(0.27)& 97.90(0.12) & 90.58(0.69) & \textbf{97.86(0.13)} & 91.01(0.27)\\
\midrule
CIFAR-100 & 72.11(0.26)& 97.16(0.03) & 65.23(0.31)&97.14(0.05) & 64.96(0.38)& 99.76(0.03) & 53.96(1.52) & 99.75(0.00) & 54.37(1.18)\\
\bottomrule
\end{tabular}}
\end{table*}

\begin{figure*}[h!]
    \centering
    \subfigure[VGG-5, MNIST, $\varepsilon^2 = 1\%$]{\includegraphics[width=0.48\textwidth]{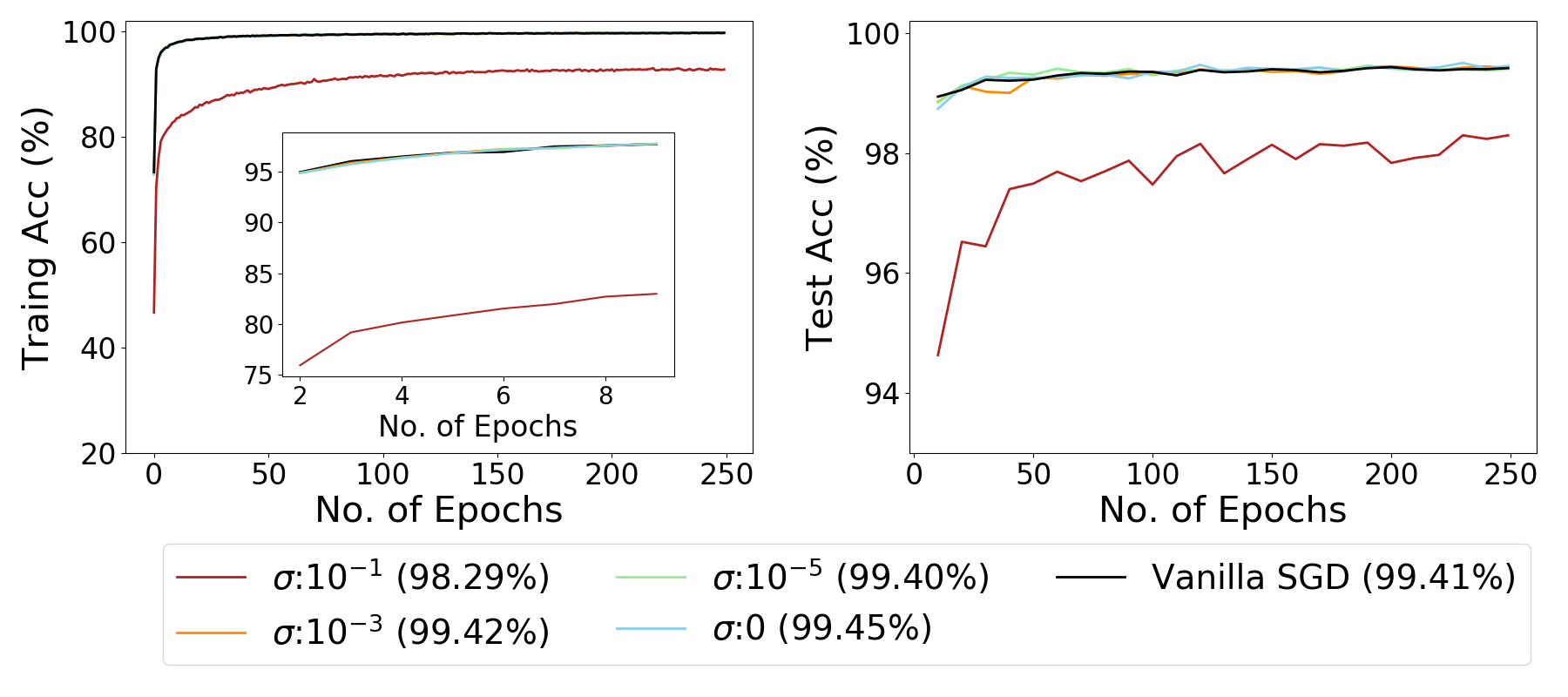}}
    \subfigure[VGG-5, MNIST, $\varepsilon^2 = 5\%$]{\includegraphics[width=0.48\textwidth]{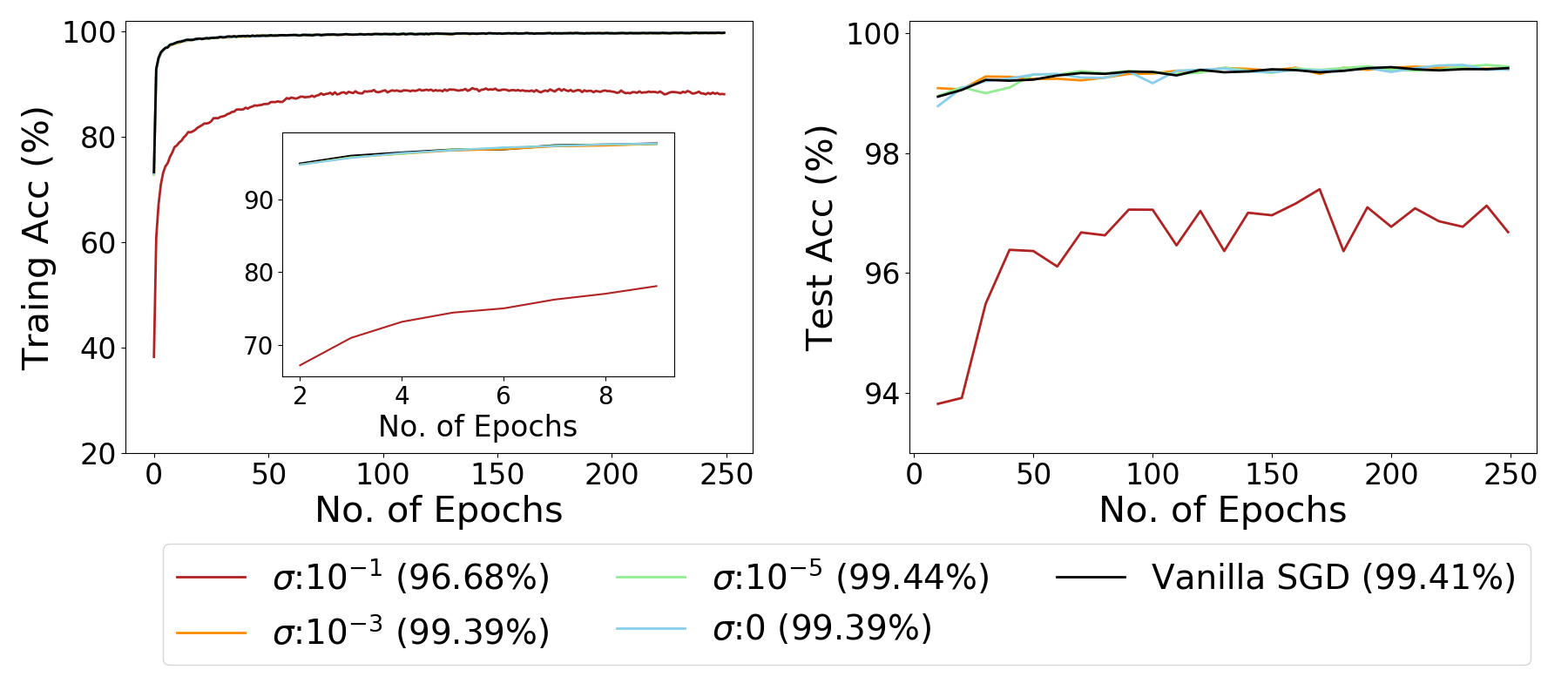}}
    \subfigure[VGG-5, MNIST, $\varepsilon^2 = 50\%$]{\includegraphics[width=0.48\textwidth]{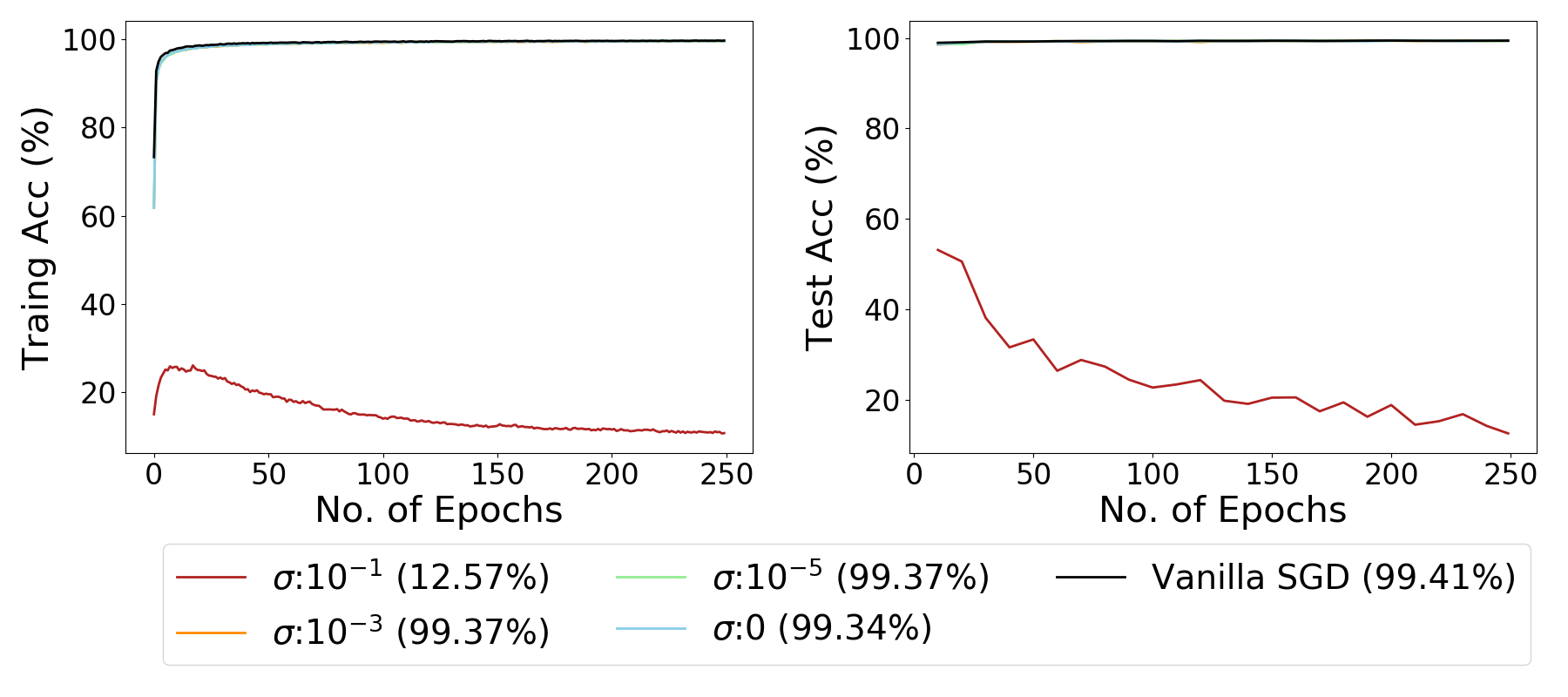}}
    \subfigure[VGG-5, MNIST, $\varepsilon^2 = 90\%$]{\includegraphics[width=0.48\textwidth]{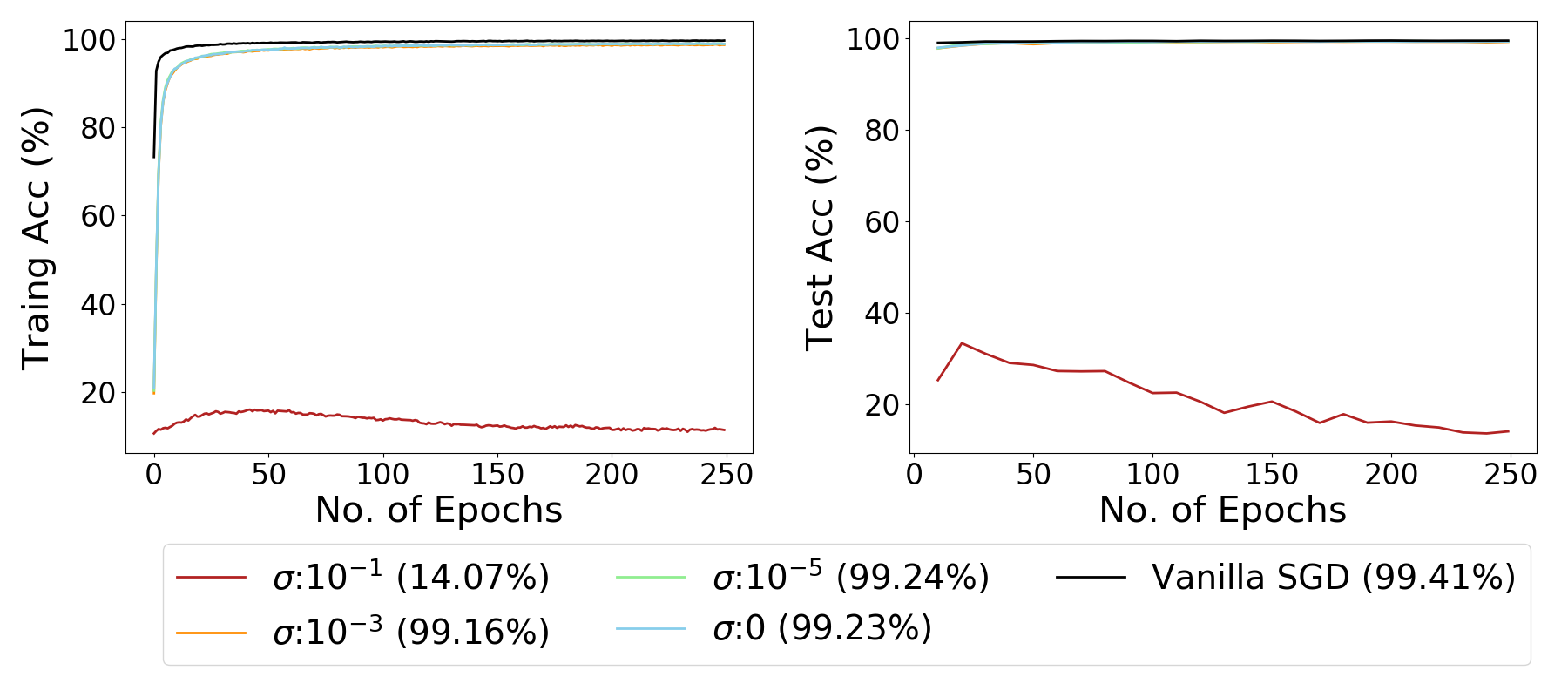}}
    \subfigure[VGG-5, Fashion-MNIST, $\varepsilon^2 = 1\%$]{\includegraphics[width=0.48\textwidth]{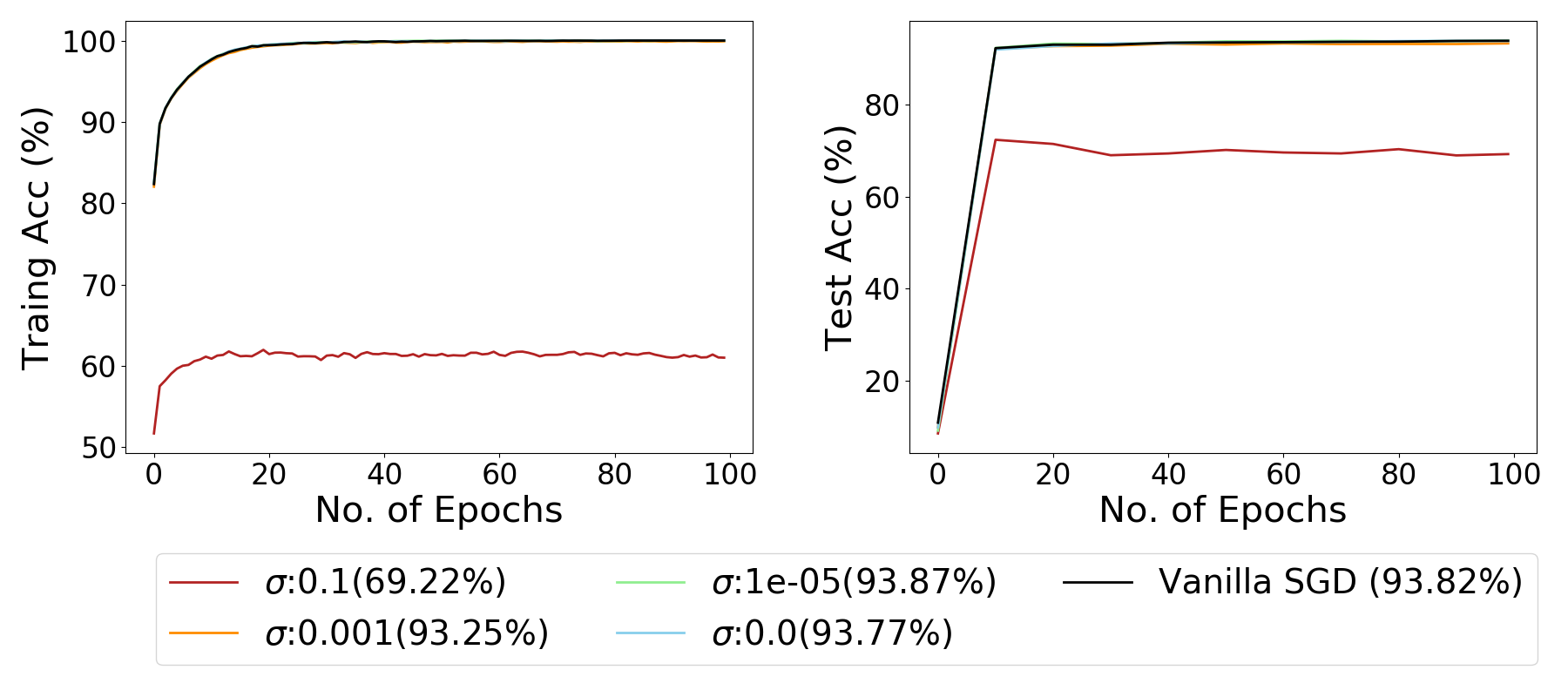}}
    \subfigure[VGG-5, Fashion-MNIST, $\varepsilon^2 = 5\%$]{\includegraphics[width=0.48\textwidth]{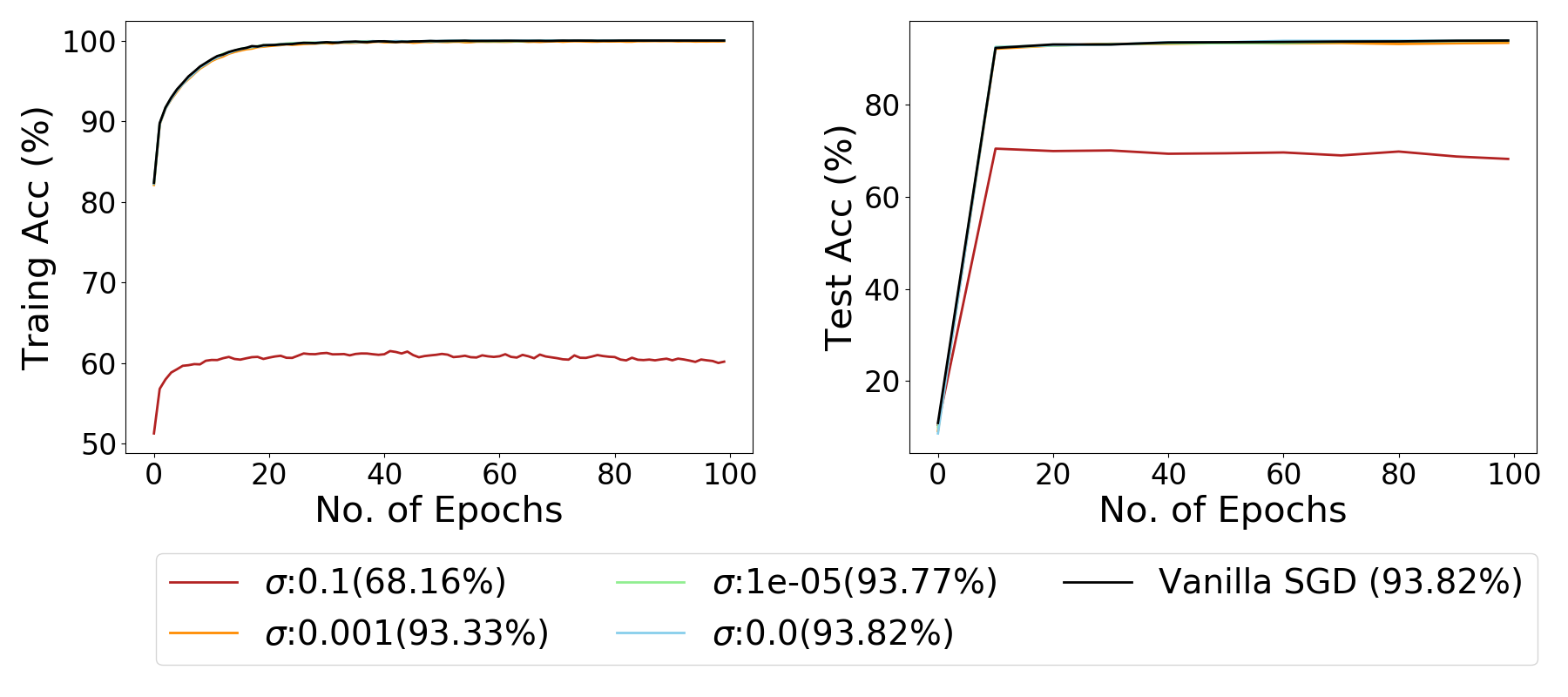}}
    \subfigure[VGG-5, Fashion-MNIST, $\varepsilon^2 = 50\%$]{\includegraphics[width=0.48\textwidth]{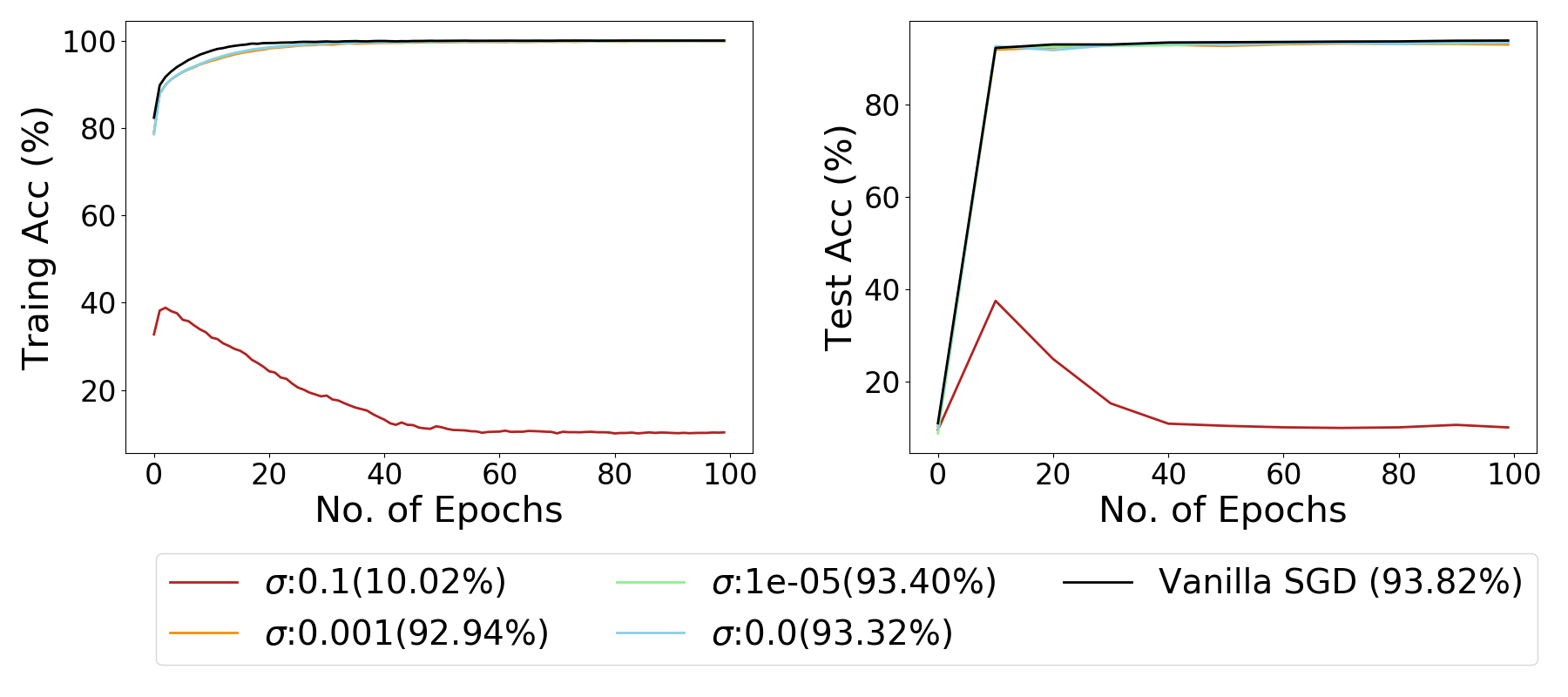}}
    \subfigure[VGG-5, Fashion-MNIST, $\varepsilon^2 = 90\%$]{\includegraphics[width=0.48\textwidth]{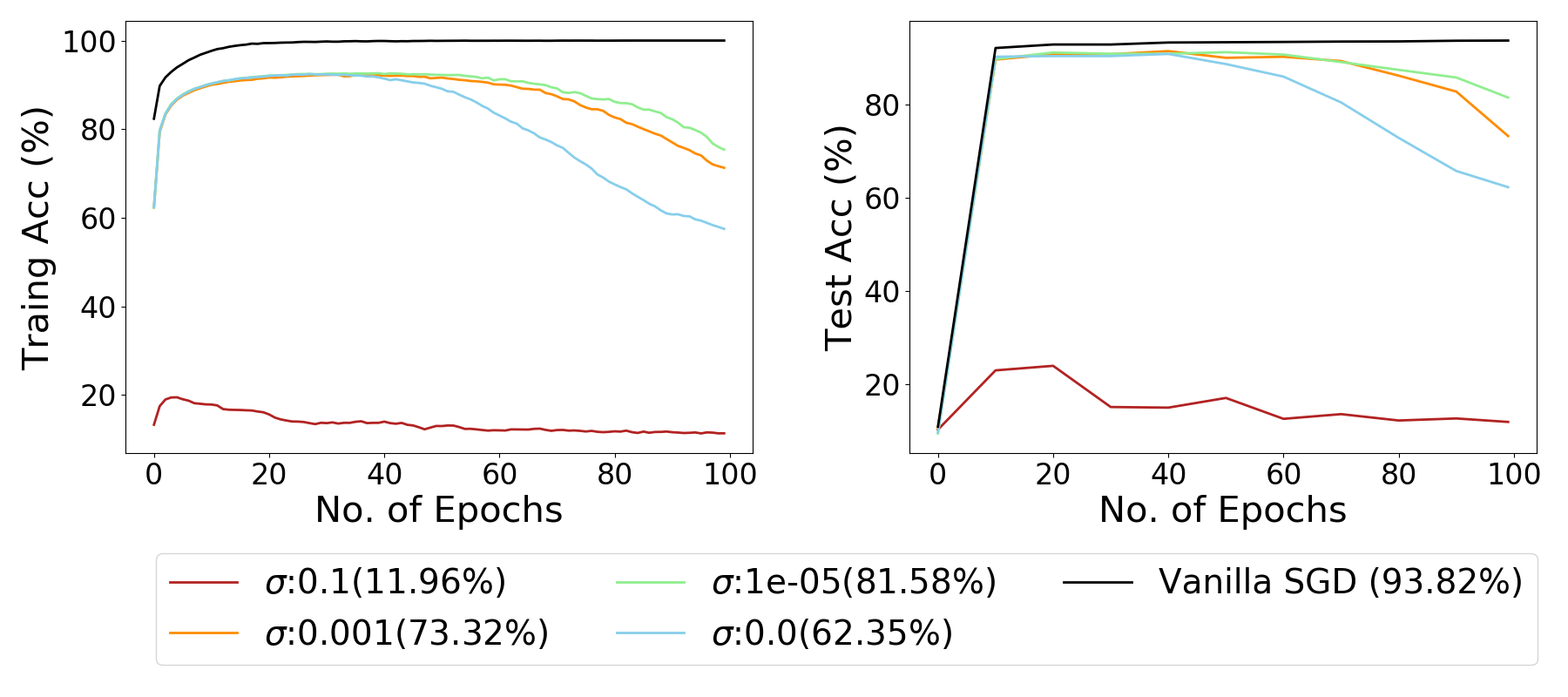}}
\end{figure*}    
    
\clearpage

\section{Proofs for Section \ref{sec:tsgd}}
\label{sec:app_tsgd}
\theotsgdopt*

\proof At every iterate $t$, since $B_t$ is uniformly sampled from $S$, the mini-batch gradient is an unbiased estimate of the empirical gradient, i..e, $\mathbb{E}[g(\w_t, B_t)] = \nabla \cL_S(\w_t)$. Note that $\kappa_{\varepsilon,t}$ is dynamically changed over $t$ and the value of which is decided by $\varepsilon^2$ to satisfy $\|  \tilde \g_t\|^2 = (1-\varepsilon^2)\| g(\w_t, B_t)\|^2$. We have for $i = 1,..., p$,

$$  [\tilde \g_t]_i =\left\{
\begin{aligned}
\left[g(\w_t, B_t)\right]_i, \ \ \text{if} \ \  |\left[g(\w_t, B_t)\right]_i| \geq \kappa_{\varepsilon,t} \\
0, \ \ \text{if} \ \  |\left|\left[g\left(\mathbf{w}_{t}, B_{t}\right)\right]_{i}\right|| < \kappa_{\varepsilon,t} , \\
\end{aligned}
\right. 
$$

To present the analysis, we decompose $g(\w_t, B_t)$ as
\begin{align*}
 g(\w_t, B_t) =  \tilde \g_t + \v_t~,  
\end{align*}
where

$$  [\v_t]_i =\left\{
\begin{aligned}
0, \ \ \text{if} \ \  |\left[g(\w_t, B_t)\right]_i| \geq \kappa_{\varepsilon,t} \\
\left[g(\w_t, B_t)\right]_i, \ \ \text{if} \ \  |\left|\left[g\left(\mathbf{w}_{t}, B_{t}\right)\right]_{i}\right|| < \kappa_{\varepsilon,t} , \\
\end{aligned}
\right. 
$$
Then we have $\|\v_t \|^2 = \varepsilon^2 \|g(\w_t, B_t) \|^2$.
Recall the update at iterate $t$ is
\begin{align*}
    \w_{t+1}=\w_{t}-\eta_t \tilde \g_t.
\end{align*}

By the smoothness \footnote{Assumption \ref{asmp: zi_grad} suggests that $\cL_S$ is $L$-smoothness.} of $\cL_S(\w)$, conditioned on iterate $t$, we have

\begin{align*}
\mathbb{E} [\cL_S(\w_{t+1})] &\leq \cL_S(\w_t) + \mathbb{E} [\left\langle \nabla \cL_S(\w_t), \w_{t+1} - \w_{t} \right\rangle ]  + \frac{L}{2} \mathbb{E} [ \|  \w_{t+1} - \w_{t}\|^2] \\
& = \cL_S(\w_t) - \eta_t \mathbb{E}[\left\langle \nabla \cL_S(\w_t),~ \tilde \g_t \right\rangle ]  + \frac{L \eta_t^2}{2} \mathbb{E}[ \|  \tilde \g_t \|^2 ] \\
& \overset{(a)}{\leq}  \cL_S(\w_t) - \eta_t\mathbb{E}
[\left\langle \nabla \cL_S(\w_t),~ g(\w_t, B_t) - \v_t \right\rangle ]
+ \frac{L \eta_t^2}{2}(1-\varepsilon^2)G^2  \\
& \overset{(b)}{\leq } \cL_S(\w_t) - \eta_t \|\nabla \cL_S(\w_t)\|^2 + \eta_t\| \nabla \cL_S(\w_t)\| \mathbb{E}[\| \v_t\|] + \frac{L \eta_t^2}{2}(1-\varepsilon^2)G^2  \\
& \overset{(c)}{\leq } \cL_S(\w_t) - \eta_t \| \nabla \cL_S(\w_t)\|^2 + \eta_t G \varepsilon \mathbb{E}[\|g(\w_t, B_t) \|] + \frac{L \eta_t^2}{2}(1-\varepsilon^2)G^2  \\
& \overset{(d )}{\leq } \cL_S(\w_t) - \eta_t \| \nabla \cL_S(\w_t)\|^2 + \frac{L \eta_t^2}{2} (2-\varepsilon^2)G^2 ,
\end{align*}
where $(a)$ is true because $\b_t$ is independent of $\nabla \cL_S(\w_t)$, $\mathbb{E}[\b_t] = 0$, $\mathbb{E}\| \tilde \g_t\|^2 = (1-\varepsilon^2)\mathbb{E}[\|g(\w_t, B_t)\|^2] \leq (1-\varepsilon^2)G^2$, and $\mathbb{E}\| \b_t\|^2 =p\sigma^2$; (b) is true because $\mathbb{E}[g(\w_t, B_t)] = \nabla \cL_S(\w_t)$ and $a^{\top}b \leq \|a\|\|b\|$ for vectors $a$ and $b$; (c) is true because  $\|\v_t\| =\varepsilon \|g(\w_t, B_t) \|$; (d) is true because $\varepsilon \leq \frac{\eta_t L}{2}$ and $\mathbb{E}\|g(\w_t, B_t) \| \leq G$.  

Rearrange the above inequality we have
\begin{align} \label{eq:one_step_descend}
    \eta_t \| \nabla \cL_S(\w_t)\|^2 \leq  \cL_S(\w_t) - \mathbb{E} [\cL_S(\w_{t+1})]  + \frac{L \eta_t^2}{2}(2-\varepsilon^2)G^2 ~.
\end{align}

Sum over $t =1$ to $t =T$ with $\eta_t = \frac{1}{\sqrt{T}}$, $\sigma^2 = \frac{R}{p}$ and apply expectation to each step, with $\w_J$ to be uniformly sampled from $\{\w_1, ..., \w_T\}$, we have
\begin{align*}
    \mathbb{E}\|\nabla \cL_S(\w_J)\|^2 = \frac{1}{T} \sum_{t=1}^{T} \mathbb{E}\|  \nabla \cL_S(\w_t)\|^2 \leq \frac{\cL_S \left(\mathbf{w}_{1}\right)-\cL_S^{\star}}{\sqrt{T}}+\frac{LG^2}{\sqrt{T}}.   
\end{align*}
where $\cL_S^\star$ is the minimal value of $\cL_S(\w)$. \qed




\tsgdstab*

\proof Consider any pair of datasets $S = \left(z_{1}, \ldots, z_{j}, \ldots, z_{n}\right)$ and $  S^\prime =\left(z_{1}, \ldots, z_{j}^{\prime}, \ldots, z_{n}\right)$ differing in exactly one data point $z_{j} \neq z_{j}^{\prime}$ for some fixed $j \in [n]$.  For $t \in [T]$, we write  $\w_1, ..., \w_T$ as the sequence of parameters updated by NT-SGD with input data $S$. Let $|B_t| = m$, where $1 \leq m \leq n$.
We write  $\w_1^\prime, ..., \w_T^\prime$ as the sequence of parameters updated by T-SGD with input data $S^\prime$. Let $\mathcal{I}_{1}, \ldots, \mathcal{I}_{T} \in[n]^{m}$ denote the index sets of the mini-batches selected in the $T$ iterations.

First, at every iteration $t$, fix the randomness in $\cI_t$. Let $k$ denote the number of occurrences of the index $j$ (where $S$ and $S^\prime$ differ) in $\cI_t$. Note that $k$ is a random variable depending on  $\cI_t$. Based on the update rule of T-SGD, let $\v_t =  g(\w_t, B_t) - \tilde \g_t $ and $\v_t^\prime = g(\w_t^\prime, B_t^\prime) - \tilde \g_t^\prime $. we have   
\begin{align*}
    \|\w_{t+1} -\w_{t+1}^\prime \| & = \| \w_t -\w_t^\prime + \eta_t \tilde \g_t^\prime -  \eta_t \tilde \g_t \| \\
    & = \| \w_t -\w_t^\prime + \eta_t g(\w_t^\prime, B_t^\prime) -  \eta_t  g(\w_t, B_t) + \eta_t \v_t - \eta_t \v_t^\prime\| \\
    & \leq \|\w_t - \w_t^\prime \| + \eta_t L \frac{m -k}{m} \|\w_t - \w_t^\prime \| + 2 \eta_t \frac{k}{m} G + \eta_t \|  \v_t - \v_t^\prime\|\\
    & = \|\w_t - \w_t^\prime \| + \eta_t L \frac{m -k}{m} \|\w_t - \w_t^\prime \| + 2 \eta_t \frac{k}{m} G + \eta_t \varepsilon (\|  g(\w_t, B_t)\| + \| g(\w_t^\prime, B_t^\prime)\|)\\
    & \leq \|\w_t - \w_t^\prime \| + \eta_t L \frac{m -k}{m} \|\w_t - \w_t^\prime \| + 2 \eta_t \frac{k}{m} G + \eta_t \frac{c^\prime}{n} 2 G~,
\end{align*}
Now, we invoke the randomness in $\cI_k$ and $\b_t$. Note that $k$ is a Binomial random variable with mean $\frac{m}{n}$. Hence, by taking expectation and conditioned on $t$, we end up with

\begin{equation} \label{eq:grad_expansive}
    \mathbb{E}_{\cI_t} \|\w_{t+1} -\w_{t+1}^\prime \| \leq \left(1 + (1-\frac{1}{n})\eta_t L \right)\|\w_t - \w_t^\prime \| + 2\frac{\eta_t}{n} G + \eta_t \frac{c^\prime}{n} 2 G.
\end{equation}

The above inequality shows the distance between $\|\w_{t+1} -\w_{t+1}^\prime \|$ can be bounded by $\|\w_{t} -\w_{t}^\prime \|$ with additional terms. We will use this result latter to derive the bound on the  stability of T-SGD with $T$ iterations. 

Recall the definition of stability which is the upper bound on $\sup _{z} \mathbb{E}_{\mathcal{A}}\left\| \ell(\mathcal{A}(S) ; z)- \ell\left(\mathcal{A}\left(S^{\prime}\right) ; z\right)\right\|$ for any $S, S^\prime$ differ in at most one sample. Now $\cA$ is the algorithm NT-SGD with $T$ iterations. Thus, $\cA(S) = \w_T$ and $\cA(S^\prime) = \w_T^\prime$. 
To derive the stability of T-SGD with $T$ iterations. We need to 
bound $\sup _{z} \mathbb{E}\left\| \ell(\w_T ; z)- \ell\left(\w_T^\prime ; z\right)\right\|$ for any $S,~ S^\prime \in \cZ^n$, where the expecation is over the randomness of $\cI_0,.., \cI_T$. The crux of the proof is to observe that T-SGD typically makes several steps before it picks up the one example which starts to produce the difference between parameters from running T-SGD with $S$ and $S^\prime$. 

For $\tau_0 \in \{1,.., T\}$, let  $t = \tau_0$ be the time step that NT-SGD has not picked up the $j$-th sample where $S$ and $S^\prime$ differ. Let $\mathbb{I}\{\|\w_{\tau_0} - \w_{\tau_0}^\prime \| = 0\}$ denotes the event that $\|\w_{\tau_0} - \w_{\tau_0}^\prime \| = 0$. To simplify the notation, we also use $\Delta_{\tau_0} = \|\w_{\tau_0} - \w_{\tau_0}^\prime \|$. Thus, write $\mathbb{I}\{\|\w_{\tau_0} - \w_{\tau_0}^\prime \| = 0\} = \mathbb{I}\{\Delta_{\tau_0} = 0\}$ for simplification. Based on Bayesian rule and gradient Lipschitz, we have for any $z \in \cZ$,
\begin{align} \label{eq:bayes_rule}
\nr \mathbb{E} \| \ell(\w_T; z) -  \ell(\w_T^\prime; z)\| & = \mathbb{P}\left\{\mathbb{I}\{\Delta_{\tau_0} = 0\}\right\} \mathbb{E}\left[ \| \ell(\w_T; z) -  \ell(\w_T^\prime; z)\| \mid \mathbb{I}\{\Delta_{\tau_0} = 0\} \right] \\
\nr & \quad + \mathbb{P}\left\{\mathbb{I}\{\Delta_{\tau_0} \neq 0\}\right\} \mathbb{E}\left[ \| \ell(\w_T; z) -  \ell(\w_T^\prime; z)\| \mid \mathbb{I}\{\Delta_{\tau_0} \neq 0\} \right] \\
\nr & \leq \mathbb{E}\left[ \| \ell(\w_T; z) -  \ell(\w_T^\prime; z)\| \mid \mathbb{I}\{\Delta_{\tau_0} = 0\} \right] \\
\nr & \quad + \mathbb{P}\left\{\mathbb{I}\{\Delta_{\tau_0} \neq 0\}\right\} \cdot \sup_{\w, \w^\prime, z}  \| \ell(\w; z) -  \ell(\w^\prime; z)\| \\
 & \leq G \cdot \mathbb{E}\left[ \|\w_T - \w_T^\prime\| \mid \mathbb{I}\{\Delta_{\tau_0} = 0\} \right] + \mathbb{P}\left\{\mathbb{I}\{\Delta_{\tau_0} \neq 0\}\right\} \cdot 2G 
\end{align}

Note that $\mathbb{P}\left\{\mathbb{I}\{\Delta_{\tau_0} \neq 0\}\right\}$ is the probability that in the first $\tau_0$ iterations, the $j$-th sample will be picked at least once. Let random variable 
$I_t$ denote the event that NT-SGD picks the $j$-th sample in iteration $t$ for  $t \leq T$, and $I^c_t$ denotes the event that NT-SGD does not  pick up the $j$-th sample in iteration $t$.  Since each sampling is independent, the probability of $j$-th sample being picked in each iteration is $1-(1-\frac{1}{n})^m \leq \frac{m}{n}$. Also sampling the batch $B_t$ is also independent for all $t \leq T$. We have $\mathbb{P}(I_1) = ... = \mathbb{P}(I_T)  \leq \frac{m}{n}$. 

Then we have
\begin{align} \label{eq:bad_prob}
\mathbb{P}\left\{\mathbb{I}\{\Delta_{\tau_0} \neq 0\}\right\} = \mathbb{P}\{ I_1 \cup I_2 \cup ... \cup I_{\tau_0}\}  \leq \sum_{t =1}^{\tau_0} \mathbb{P}\{I_t\}  \leq \frac{\tau_0 m}{n}~.
\end{align}

Now we bound $\mathbb{E}\left[ \|\w_T - \w_T^\prime\| \mid \mathbb{I}\{\Delta_{\tau_0} = 0\} \right]$. We apply \eqref{eq:grad_expansive} recursively from $T$ to $\tau_0 + 1$ with expectation over the randomness of $\mathcal{I}_{0}, . ., \mathcal{I}_{T}$. 

1) Considering the case $\eta_t = c/t$, we have

\begin{align*}
\mathbb{E}\left[ \|\w_T - \w_T^\prime\| \mid \mathbb{I}\{\Delta_{\tau_0} = 0\} \right]  & \overset{(a)}{\leq} \sum_{t = \tau_0 +1}^T \left\{\Pi_{k = t +1}^T \left(1+\left(1-\frac{1}{n}\right) \eta_t  L\right)\right\}\left(\frac{2G\eta_t}{n}(1 + c^\prime)\right) \\
&  \overset{(b)}{\leq} \sum_{t = \tau_0 +1}^T \left\{\Pi_{k = t +1}^T \exp\left(\left(1-\frac{1}{n}\right) \eta_t  L\right)\right\}\left(\frac{2G\eta_t}{n}(1 + c^\prime)\right) \\
&  \overset{(c)}{\leq} \sum_{t = \tau_0 +1}^T \left\{ \exp\left(\left(1-\frac{1}{n}\right) \sum_{k=t+1}^T\frac{c}{t} L\right)\right\}\left(\frac{2Gc}{tn}(1 + c^\prime)\right) \\
& \leq \sum_{t = \tau_0 +1}^T \left\{ \exp\left(\left(1-\frac{1}{n}\right) \log \frac{T}{t} cL\right)\right\}\left(\frac{2Gc}{tn}(1 + c^\prime)\right)\\
& \leq  \sum_{t = \tau_0 +1}^T \left\{\left( \frac{T}{t}\right)^{(1-\frac{1}{n})Lc}\right\} \frac{1}{t}\frac{2Gc(1+c^\prime)}{n}\\
&  = \frac{2Gc(1+c^\prime) T^{(1-\frac{1}{n})Lc}}{n} \sum_{t = \tau_0 +1}^T t^{-(1-\frac{1}{n})Lc -1} \\
& \leq \frac{2G(1+c^\prime)}{n(1-\frac{1}{n})L} \left(\frac{T}{\tau_0}\right)^{\left(1-\frac{1}{n}\right)Lc}
~,
\end{align*}
where $(a)$ is true because $\|\w_{\tau_0} - \w_{\tau_0}^\prime\| = 0$; $(b)$ is true because $(1+x) \leq \exp(x)$ for all $x$; $(c)$ is ture because $\eta_t = \frac{c}{t}$ for all $t \leq T$.  Now combine the above inequality with \eqref{eq:bad_prob} and \eqref{eq:bayes_rule}, we have 

\begin{align*}
\mathbb{E} \| \ell(\w_T; z) -  \ell(\w_T^\prime; z)\| & \leq  \frac{2G^2(1+c^\prime)}{n(1-\frac{1}{n})} \left(\frac{T}{t}\right)^{\left(1-\frac{1}{n}\right)Lc}   +  2G\frac{\tau_0 m}{n}~.
\end{align*}

The right hand side is approximately minimized when
\begin{align*}
    \tau_0 = m^{-\frac{1}{Lc +1}} T^{\frac{Lc}{Lc +1}}.
\end{align*}
Under this setting, we have
\begin{align*}
\mathbb{E} \| \ell(\w_T; z) -  \ell(\w_T^\prime; z)\| & \leq  \frac{2G^2(2+c^\prime) (m T)^{1 - \frac{1}{Lc +1}}}{n(1-\frac{1}{n})}~.
\end{align*}

2) Considering the case $\eta_t = c/\sqrt{T}$ for constant $c > 0$, we have

\begin{align*}
\mathbb{E}\left[ \|\w_T - \w_T^\prime\| \mid \mathbb{I}\{\Delta_{\tau_0} = 0\} \right]  & \overset{(a)}{\leq} \sum_{t = \tau_0 +1}^T \left\{\Pi_{k = t +1}^T \left(1+\left(1-\frac{1}{n}\right) \eta_t  L\right)\right\}\left(\frac{2G\eta_t}{n}(1 + c^\prime)\right) \\
&  \overset{(b)}{\leq} \sum_{t = \tau_0 +1}^T \left\{\Pi_{k = t +1}^T \left(1+\left(1-\frac{1}{n}\right) \frac{cL}{\sqrt{T}}\right)\right\} \left(\frac{2Gc}{\sqrt{T}}(1 + c^\prime)\right) \\
& \leq \frac{2Gc}{\sqrt{T}}(1 + c^\prime) \sum_{t = \tau_0 +1}^T \left(1+\left(1-\frac{1}{n}\right) \frac{cL}{\sqrt{T}}\right)^{T-t}
\\
& \leq \frac{2G(1 + c^\prime) }{n(1-\frac{1}{n})L} \left( 1+ (1-\frac{1}{n})\frac{cL}{\sqrt{T}}\right)^{T-\tau_0}
~,
\end{align*}
where $(a)$ is true because $\|\w_{\tau_0} - \w_{\tau_0}^\prime\| = 0$; $(b)$  is ture because $\eta_t = \frac{c}{\sqrt{T}}$ for all $t \leq T$.  Now combine the above inequality with \eqref{eq:bad_prob} and \eqref{eq:bayes_rule}, we have 
\begin{align*}
\mathbb{E} \| \ell(\w_T; z) -  \ell(\w_T^\prime; z)\| & \leq  \frac{2G^2(1 + c^\prime) }{n(1-\frac{1}{n})L} \left( 1+ (1-\frac{1}{n})\frac{cL}{\sqrt{T}}\right)^{T-\tau_0}  +  2G\frac{\tau_0 m}{n}~.
\end{align*}

The right hand side is approximately minimized when
\begin{align*}
    \tau_0 = T - \ln_{\left(1+ (1-\frac{1}{n})\frac{cL}{\sqrt{T}} \right)}\frac{m(1-\frac{1}{n}L)}{(1+c^\prime)\ln\left(1+ (1-\frac{1}{n})\frac{cL}{\sqrt{T}} \right)}
\end{align*}
Under this setting, we have
\begin{align*}
\mathbb{E} \| \ell(\w_T; z) -  \ell(\w_T^\prime; z)\| & \leq  O(\frac{G^2 mT}{n})
\end{align*}



\corrgentsgd*  

\proof Based on Theorem \ref{theo:gen_stab}, if an algorithm is $\alpha$-uniformly stable, then we have its generalization error bounded by $\alpha$. In Theorem \ref{theo:gen_stab}, we show that T-SGD with $T$ iterations, $|B_t| =1$ and $\eta_t = O(1/\sqrt{T})$ is $O(\frac{G^2T}{n})$-uniformly stable. Thus we have
$\operatorname{err}_{\mathrm{gen}}\left(\mathbf{w}_{T}\right) \leq O\left(G^{2} T / n\right)$. That completes the proof. \qed

\section{Proofs for Section \ref{sec:ntsgd_opt}}
\label{sec:app_opt}
\theoopt*

\proof At every iterate $t$, since $B_t$ is uniformly sampled from $S$, the mini-batch gradient is an unbiased estimate of the empirical gradient, i..e, $\mathbb{E}[g(\w_t, B_t)] = \nabla \cL_S(\w_t)$. Note that $\kappa_{\varepsilon,t}$ is dynamically changed over $t$ and the value of which is decided by $\varepsilon^2$ to satisfy $\|  \tilde \g_t\|^2 = (1-\varepsilon^2)\| g(\w_t, B_t)\|^2$. We have for $i = 1,..., p$,

$$  [\tilde \g_t]_i =\left\{
\begin{aligned}
\left[g(\w_t, B_t)\right]_i, \ \ \text{if} \ \  |\left[g(\w_t, B_t)\right]_i| \geq \kappa_{\varepsilon,t} \\
0, \ \ \text{if} \ \  |\left|\left[g\left(\mathbf{w}_{t}, B_{t}\right)\right]_{i}\right|| < \kappa_{\varepsilon,t} , \\
\end{aligned}
\right. 
$$

To present the analysis, we decompose $g(\w_t, B_t)$ as
\begin{align*}
 g(\w_t, B_t) =  \tilde \g_t + \v_t~,  
\end{align*}
where

$$  [\v_t]_i =\left\{
\begin{aligned}
0, \ \ \text{if} \ \  |\left[g(\w_t, B_t)\right]_i| \geq \kappa_{\varepsilon,t} \\
\left[g(\w_t, B_t)\right]_i, \ \ \text{if} \ \  |\left|\left[g\left(\mathbf{w}_{t}, B_{t}\right)\right]_{i}\right|| < \kappa_{\varepsilon,t} , \\
\end{aligned}
\right. 
$$
Then we have $\|\v_t \|^2 = \varepsilon^2 \|g(\w_t, B_t) \|^2$.
Recall the update at iterate $t$ is
\begin{align*}
    \w_{t+1}=\w_{t}-\eta_t (\tilde \g_t + \b_t).
\end{align*}

By the smoothness \footnote{Assumption \ref{asmp: zi_grad} suggests that $\cL_S$ is $L$-smoothness.} of $\cL_S(\w)$, conditioned on iterate $t$, we have

\begin{align*}
\mathbb{E} [\cL_S(\w_{t+1})] &\leq \cL_S(\w_t) + \mathbb{E} [\left\langle \nabla \cL_S(\w_t), \w_{t+1} - \w_{t} \right\rangle ]  + \frac{L}{2} \mathbb{E} [ \|  \w_{t+1} - \w_{t}\|^2] \\
& = \cL_S(\w_t) - \eta_t \mathbb{E}[\left\langle \nabla \cL_S(\w_t),~ \tilde \g_t +\b_{t}\right\rangle ]  + \frac{L \eta_t^2}{2} \mathbb{E}[ \|  \tilde \g_t +\b_{t}\|^2 ] \\
& \overset{(a)}{\leq}  \cL_S(\w_t) - \eta_t\mathbb{E}
[\left\langle \nabla \cL_S(\w_t),~ g(\w_t, B_t) - \v_t \right\rangle ]
+ \frac{L \eta_t^2}{2}\left((1-\varepsilon^2)G^2 + p\sigma^2\right) \\
& \overset{(b)}{\leq } \cL_S(\w_t) - \eta_t \|\nabla \cL_S(\w_t)\|^2 + \eta_t\| \nabla \cL_S(\w_t)\| \mathbb{E}[\| \v_t\|] + \frac{L \eta_t^2}{2}\left((1-\varepsilon^2)G^2 + p\sigma^2\right) \\
& \overset{(c)}{\leq } \cL_S(\w_t) - \eta_t \| \nabla \cL_S(\w_t)\|^2 + \eta_t G \varepsilon \mathbb{E}[\|g(\w_t, B_t) \|] + \frac{L \eta_t^2}{2}\left((1-\varepsilon^2)G^2 + p\sigma^2\right) \\
& \overset{(d )}{\leq } \cL_S(\w_t) - \eta_t \| \nabla \cL_S(\w_t)\|^2 + \frac{L \eta_t^2}{2}\left((2-\varepsilon^2)G^2 + p\sigma^2\right),
\end{align*}
where $(a)$ is true because $\b_t$ is independent of $\nabla \cL_S(\w_t)$, $\mathbb{E}[\b_t] = 0$, $\mathbb{E}\| \tilde \g_t\|^2 = (1-\varepsilon^2)\mathbb{E}[\|g(\w_t, B_t)\|^2] \leq (1-\varepsilon^2)G^2$, and $\mathbb{E}\| \b_t\|^2 =p\sigma^2$; (b) is true because $\mathbb{E}[g(\w_t, B_t)] = \nabla \cL_S(\w_t)$ and $a^{\top}b \leq \|a\|\|b\|$ for vectors $a$ and $b$; (c) is true because  $\|\v_t\| =\varepsilon \|g(\w_t, B_t) \|$; (d) is true because $\varepsilon \leq \frac{\eta_t L}{2}$ and $\mathbb{E}\|g(\w_t, B_t) \| \leq G$.  

Rearrange the above inequality we have
\begin{align} \label{eq:one_step_descend}
    \eta_t \| \nabla \cL_S(\w_t)\|^2 \leq  \cL_S(\w_t) - \mathbb{E} [\cL_S(\w_{t+1})]  + \frac{L \eta_t^2}{2}\left((2-\varepsilon^2)G^2 + p\sigma^2\right)~.
\end{align}

Sum over $t =1$ to $t =T$ with $\eta_t = \frac{1}{\sqrt{T}}$, $\sigma^2 = \frac{R}{p}$ and apply expectation to each step, with $\w_J$ to be uniformly sampled from $\{\w_1, ..., \w_T\}$, we have
\begin{align*}
    \mathbb{E}\|\nabla \cL_S(\w_J)\|^2 = \frac{1}{T} \sum_{t=1}^{T} \mathbb{E}\|  \nabla \cL_S(\w_t)\|^2 \leq \frac{\cL_S \left(\mathbf{w}_{1}\right)-\cL_S^{\star}}{\sqrt{T}}+\frac{L\left((2-\varepsilon^2)G^2 + R^2\right)}{\sqrt{T}}.   
\end{align*}
where $\cL_S^\star$ is the minimal value of $\cL_S(\w)$. \qed

\theoescape*

\proof 
The proof presented is inspired by the analysis in \cite{daneshmand2018escaping} that utilized the idea of contradiction in the lower bound and upper bound on the distance moved over a given number of iterations when the loss function stops decreasing. This idea is originated from \cite{jin2017escape,jin2018accelerated,jin2019nonconvex}. We first assume that the loss does not decrease sufficiently by $F$, i.e., $ \mathbb{E}[\cL_{S}(\w_{t_0 +\tau}) - \cL_{S}(\w_{t_0})] > - F$. The value of $F$ will be analyzed later. Under this assumption, we derive an upper bound and lower bound on the distance moved over a given number of iterations. We then show that the lower bound contradicts the upper bound for the specific choice of parameters introduced earlier.
So the proof is composed of two parts: 1) upper bound on $\|\w_{t_0 +\tau} -\w_{t_0} \|^2$ and 2) lower bound on $\|\w_{t_0 +\tau} -\w_{t_0} \|^2$. 

\textbf{Part 1: Upper bounding the distance on the iterates $\|\w_{t_0 +\tau} -\w_{t_0} \|^2$.} 

We assume that the loss does not decrease sufficiently by $F$ in $\tau$ iterates, i.e.,
\begin{equation} \label{eq:loss_bound}
 \mathbb{E}[\cL_{S}(\w_{t_0 +\tau}) - \cL_{S}(\w_{t_0})] > - F~.
\end{equation}

Then we have the iterates $\w_{t_0 +\tau}$ stay close to $\w_{t_0}$. We formalize this result in the following
lemma.

\begin{lemm} \label{lemm:ub_distance}
(Distance Upper Bound) The expected distance to the
initial parameter can be bounded as
\begin{align*}
\mathbb{E}\|\w_{t_0 +\tau} - \w_{t_0}\|^2 \leq \left(4 \eta F + 2\eta^2 p\sigma^2 \right) \cdot \tau + \left(12 \eta^2 G^2 + 2\eta^3 L p \sigma^2\right)\cdot \tau^2    
\end{align*}
as long as $\eta \leq \frac{1}{L}$.
\end{lemm}

\proof
Since the algorithm is Markovian, the update in each iteration only depends on the  current time step. Thus, it suffices to prove Lemma \ref{lemm:ub_distance} for special case $t_0 = 0$ and $\tau + t_0 = \tau$. So $\w_\tau$  is the $\tau$-th update of NT-SGD starting from $\w_0$. Let the mini-batch gradient be decomposed as $g\left(\mathbf{w}_{s}, B_{s}\right)=\tilde \g_s +\mathbf{v}_{s}$ for $s = 1,..., \tau$, with $\| \tilde \g_s \|^2 = (1 -\varepsilon^2)\|g\left(\mathbf{w}_{s}, B_{s}\right) \|^2$ and $\|\v_s \|^2 = \varepsilon^2\|g\left(\mathbf{w}_{s}, B_{s}\right) \|^2$.
Recall that the update of NT-SGD is
\begin{align*}
    \w_{s+1} = \w_{s} - \eta(\tilde \g_s + \b_s).
\end{align*}

For all $s = 1,...,\tau$, conditioned on $\w_s$,  from \eqref{eq:one_step_descend}, we have

\begin{align*} 
   \| \nabla \cL_S(\w_s)\| \leq  \frac{\cL_S(\w_s) - \mathbb{E} [\cL_S(\w_{s+1})]}{\eta}  + \frac{L \eta}{2}\left((2-\varepsilon^2)G^2 + p\sigma^2\right)~.
\end{align*}

Sum over $s = 1$ to $s = \tau$ with expectation, we have
\begin{align} \label{eq:sum_grad}
   \mathbb{E} \sum_{s=1}^\tau \| \nabla \cL_S(\w_{s-1})\|^2 \leq \frac{\mathbb{E}[\cL_{S}(\w_0) - \cL_{S} (\w_{\tau})]}{\eta} + \frac{L\eta(2G^2 + p\sigma^2) \tau}{2}
\end{align}

We can also bound the norm of the sparse gradient $\tilde \g_s$ as

\begin{align*}
\mathbb{E}\|\tilde \g_s \|^2 & = (1-\varepsilon^2)\mathbb{E}\|g(\w_s, B_s) \|^2 \leq (1-\varepsilon^2)\mathbb{E}\left[
\left(\| \nabla \cL_S(\w_s)\| + \| g(\w_s, B_s) - \nabla \cL_S(\w_s)\|\right)^2
\right] \\
& \leq (1-\varepsilon^2)\mathbb{E}\left[
2\| \nabla \cL_S(\w_s)\|^2 + 8G^2
\right]
\end{align*}

Sum over $s = 1$ to $s = \tau$, with \eqref{eq:sum_grad}, we have
\begin{align*}
\mathbb{E} \sum_{s=1}^\tau \|\g_{s-1} \|^2 &\leq (1-\varepsilon^2) \left(2 \mathbb{E} \sum_{s=1}^\tau \| \nabla \cL_S(\w_{s-1})\|_2^2 + 8 G^2 \tau\right) \\
& \leq (1-\varepsilon^2) \left(\frac{2\mathbb{E}[\cL_{S}(\w_0) - \cL_{S} (\w_{\tau})]}{\eta} + L\eta(2G^2 + p\sigma^2) \tau + 8G^2 \tau \right)
\end{align*}

Now we bound $\mathbb{E} \| \w_\tau -\w_0\|^2 $:

\begin{align*}
    \mathbb{E} \| \w_\tau -\w_0\|^2 & = \mathbb{E} \left\|  \sum_{s=1}^\tau (\w_s-\w_{s-1}) \right\|^2 = \eta^2 \mathbb{E} \left\|  \sum_{s=1}^\tau\g_{s-1} + \sum_{s=1}^\tau\b_{s-1} \right\|^2 = 2 \eta^2 \left(\mathbb{E} \left\|  \sum_{s=1}^\tau\g_{s-1} \right\|^2 + \mathbb{E} \left\|  \sum_{s=1}^\tau\b_{s-1} \right\|^2 \right) 
    \\& \overset{(a)}{\leq} 2 \eta^2 \left(\tau\cdot  \mathbb{E} \sum_{s=1}^\tau  \left\| \g_{s-1} \right\|^2 + \sum_{s=1}^\tau \mathbb{E} \left\| \b_{s-1} \right\|^2 \right) 
    = \tau \cdot 2 \eta^2\left( \mathbb{E} \left[\sum_{s=1}^\tau  \|\g_{s-1} \|^2\right] + p\sigma^2 \right)\\
    & \leq \tau \cdot 2 \eta^2 \left( \frac{2\mathbb{E}[\cL_{S}(\w_0) - \cL_{S} (\w_{\tau})]}{\eta} + L\eta(2G^2 + p\sigma^2) \tau + 8G^2 \tau + p\sigma^2 \right) \\
    & \overset{(b)}{\leq} \left(4 \eta F + 2\eta^2 p\sigma^2 \right) \cdot \tau + \left(20 \eta^2 G^2 + 2\eta^3 L p \sigma^2\right)
 \cdot \tau^2,
\end{align*}
where (a) holds because of Cauchy–Schwarz inequality and $\b_1,...,\b_\tau$ are i.i.d. Gaussian; (b) is true because $\eta L \leq 1$ and we assume $\mathbb{E}[\cL_{S} (\w_\tau) - \cL_{S}(\w_0) ] > - F$. \qed 

\textbf{Part 2: Lower bounding the distance on the iterates $\|\w_{t_0 +\tau} - \w_{t_0} \|^2$.}

To lower bound the distance, we will use the approach from \cite{daneshmand2018escaping} which used the quadratic approximation of the loss function and the Hessian Lipshitz condition. By a similar argument as in the proof of Lemma \ref{lemm:ub_distance}, it suffices to derive the lower bound in the special case $t_0 = 0$.
Since the parameter vector stays close to $\w_0$ under the condition in \eqref{eq:loss_bound}, we can
use a  Taylor expansion approximation of the function $\cL_{S}$ at $\w_0$:
$$\phi(\mathbf{w})=\cL_{S}(\w_0)+(\mathbf{w}-\mathbf{w}_0)^{\top} \nabla \cL_{S}(\mathbf{w}_0)+\frac{1}{2}(\mathbf{w}-\mathbf{w}_0)^{\top} H(\w_0)(\mathbf{w}-\mathbf{w}_0),$$
where the $H(\w_0) = \nabla^2 \cL_S(\w_0)$ is the Hessian of $\cL_{S}(\w_0)$ w.r.t. $\w_0$. 

Using Lemma 9 from \cite{daneshmand2018escaping}, we have
\begin{align*}
\|\nabla \cL_{S}(\mathbf{w})-\nabla \phi(\mathbf{w})\| \leq \frac{\rho}{2}\|\mathbf{w}-{\mathbf{w}}_0\|^{2}~.    
\end{align*}

Then we have, for $s = 1,..., \tau$,

\begin{equation} \label{eq:grad_by_param}
     \|\tilde \g_s -\nabla \phi(\mathbf{w}_s)\| \leq \| \tilde \g_s -\nabla \cL_{S}(\mathbf{w}_s)\| + \|\nabla \cL_{S}(\mathbf{w}_s)-\nabla \phi(\mathbf{w}_s)\| \leq (2-\varepsilon^2) G +  \frac{\rho}{2}\|\mathbf{w}_s-\mathbf{w}_0\|^{2}~.
\end{equation}

Furthermore, the guaranteed closeness to the initial parameter allows us to use the gradient of the quadratic objective $\phi(\w)$ in the NT-SGD steps as follows. To simplify the notation, we use $\h_s = \nabla \phi(\w_s)$ as the gradient of $\phi(\w_s)$ at $\w_s$ for $s = 0,...,\tau$.

\begin{small}
\begin{align*}
    &\w_{\tau+1 } - \w_0 \\
    & = \w_\tau -\eta \tilde \g_\tau - \eta \b_\tau -\w_0 \\
    & =  \w_\tau -  \w_0 - \eta \h_\tau + \eta(\h_\tau - \tilde \g_\tau) - \eta \b_\tau\\
    & = \w_\tau -\w_0 - \eta \nabla \cL_{S}(\w_0) -\eta H(\w_0)(\w_\tau - \w_0) + \eta(\h_\tau - \tilde \g_\tau) - \eta \b_\tau\\
    & = (\mathbb{I} - \eta H(\w_0)) (\w_\tau -\w_0) + \eta(\h_\tau - \tilde \g_\tau) - \eta \nabla \cL_{S}(\w_0) - \eta \b_\tau \\
    & = \underbrace{(\mathbb{I} - \eta H(\w_0))^\tau (\w_1 -\w_0)}_{A_\tau} + \eta \underbrace{\sum_{s =1}^\tau (\mathbb{I} - \eta H(\w_0))^{\tau-s} (\h_s - \tilde \g_s)}_{B_\tau} - \eta \underbrace{\sum_{s =1}^\tau (\mathbb{I} - \eta H(\w_0))^{\tau-s} \nabla \cL_{S}(\w_0)}_{C_\tau} - \eta \underbrace{\sum_{s =1}^\tau (\mathbb{I} - \eta H(\w_0))^{\tau-s} \b_s}_{D_\tau}
    \\
    & =  A_\tau + \eta B_\tau -\eta C_\tau - \eta D_\tau
\end{align*}
\end{small}

Now we lower bound $\mathbb{E}\|\w_\tau -\w_0 \|^2 =\mathbb{E} \|A_\tau  + \eta B_\tau -\eta C_\tau - \eta D_\tau \|^2 $ using $\|a+b\|^{2} \geq\|a\|^{2}+2 a^{\top} b$ with $a = A_\tau - \eta D_\tau$ and $b =\eta B_\tau -\eta C_\tau $. We have

\begin{align*}
\mathbb{E}\|\w_\tau -\w_0 \|^2 & = 
  \mathbb{E} \|A_\tau  + \eta B_\tau -\eta C_\tau + \eta D_\tau \|^2 \\
  & \geq   \mathbb{E} \left(\|A_\tau\|^2 + \eta^2  \| D_\tau \|^2 - 2\eta A_\tau^{\top} D_\tau \right) 
  + 2\eta \mathbb{E}\left[ \left( A_\tau +\eta D_\tau \right)^{\top} \left( B_\tau  -  C_\tau \right)\right]\\
  & \geq \mathbb{E}[\|A_\tau\|^2] + \eta^2 \mathbb{E}[\|D_\tau\|^2] + 2\eta \mathbb{E} [A_\tau^{\top} (B_\tau -C_\tau)],
\end{align*}
where the last inequality is true because $\b_s$, for $s = 1,..., \tau$, in term $D_\tau$ is independent of $B_\tau$ and $C_\tau$ and the expectation of $\b_s$ is zero. Thus the linear product of $B_\tau$ and $C_\tau$ with $D_\tau$ is zero in expectation:
$\mathbb{E}[D_\tau^{\top} (B_\tau -C_\tau)] = 0$. Also the term $- 2\eta \mathbb{E}[ A_\tau^{\top} D_\tau] \geq 0$:

\begin{align*}
   - 2\eta \mathbb{E} [ A_\tau^{\top} D_\tau]& = - 2\eta  \mathbb{E} \left\langle (\mathbb{I} - \eta H(\w_0))^\tau (\w_1 -\w_0),~ \sum_{s =1}^\tau (\mathbb{I} - \eta H(\w_0))^{\tau-s} \b_s \right\rangle  \\
    & = - 2 \eta^2 \mathbb{E} \left\langle (\mathbb{I} - \eta H(\w_0))^\tau (-\tilde \g_0 - \b_0),~ \sum_{s =1}^\tau (\mathbb{I} - \eta H(\w_0))^{\tau-s} \b_s \right\rangle \\
    & \overset{(a)}{=} - 2 \eta^2 \mathbb{E} \left\langle (\mathbb{I} - \eta H(\w_0))^\tau (-\tilde \g_0 - \b_0),~  (\mathbb{I} - \eta H(\w_0))^\tau \b_0 \right\rangle\\
    &   \overset{(b)}{=} 2 \eta^2 \mathbb{E} \left\langle (\mathbb{I} - \eta H(\w_0))^\tau \b_0,~  (\mathbb{I} - \eta H(\w_0))^\tau \b_0 \right\rangle  \\
    & = 2 \eta^2 \mathbb{E} \left[\b_0^{\top} (\mathbb{I} - \eta H(\w_0))^{2\tau} \b_0\right]\overset{(c)}{\geq} 0,
\end{align*}

where $(a)$ is true because $\b_s$ is independent of $(\mathbb{I} - \eta H(\w_0))^\tau (-\tilde \g_0 - \b_0)$ for $s =1,...,\tau$ and $\mathbb{E}[\b_s] =0$; (b) is true because $\b_0$ is independent of $\tilde \g_0$ and $\mathbb{E}[\b_0] =0$; (c) is true because  $(\mathbb{I} - \eta H(\w_0))^{2\tau}$ is a positive semi-definite matrix for $\eta < \frac{1}{L}$.

Now we lower bound
\begin{align*}
\mathbb{E}\|\w_\tau -\w_0 \|^2 &  \geq \mathbb{E}[\|A_\tau\|^2] + \eta^2 \mathbb{E}[\|D_\tau\|^2] + 2\eta \mathbb{E} [A_\tau^{\top} (B_\tau -C_\tau)] 
\end{align*}

We need to lower bound each term in the above inequality, where the computation for bounded series refers to Lemma 
\ref{lemma:bs}.

\textbf{(I) Lower-bound on $\mathbb{E} \|A_\tau\|^2$}. 

Let  $\v_{\w_0}$ be the eigenvector corresponding to $|\lambda_{\min}(H(\w_0))| =\lambda$. Let $\chi = (1+\eta \lambda)$ We have 
\begin{align*}
    \mathbb{E} [\|A_\tau\|^2] &\geq \mathbb{E}[ \|\v_{\w_0}^{\top} A_\tau\|^2] \\
    & = \mathbb{E} \left\|\v_{\w_0}^{\top} (\mathbb{I} - \eta H(\w_0))^\tau (\w_1 -\w_0)\right\|^2 \\
   & =  \eta^2 (1+\eta \lambda)^{2\tau} \mathbb{E}\left[\left(\v_{\w_0}^{\top}\left(-\tilde \g_0 + \b_0\right)\right)^{2}\right] \\
   & =  \eta^2 (1+\eta \lambda)^{2\tau} \mathbb{E}\left[\| \v_{\w_0}^{\top}\tilde \g_0\|^2 + \| \v_{\w_0}^{\top}\b_0\|^2\right] \\
    & \geq \eta^2\chi^{2\tau} \sigma^2 
\end{align*}

\textbf{(II) Lower-bound on $\mathbb{E} \|D_\tau\|^2$}. 

For a matrix $A$, let $R(A)$ denote the stable rank of $A$, i.e., $R(A) = \frac{\tr(A)}{\|A\|_2}$. Let $\Lambda_\tau = R((\mathbb{I} - \eta H(\w_0))^{2\tau})$ denote the stable rank of $(\mathbb{I} - \eta H(\w_0))^{2\tau}$. We have

\begin{align*}
\mathbb{E}\|D_\tau \|^2  & = \mathbb{E}\left\|\sum_{s =1}^\tau (\mathbb{I} - \eta H(\w_0))^{\tau-s} \b_s \right\|^2 =   \sum_{s =1}^\tau   \mathbb{E} \left\| (\mathbb{I} - \eta H(\w_0))^{\tau-s} \b_s \right\|^2 \\
& = \sigma^2  \sum_{s =1}^\tau \tr\left((\mathbb{I} - \eta H(\w_0))^{2\tau-2s}\right) \geq \sum_{s =1}^\tau (1+\eta \lambda)^{2\tau-2s} \sigma^2 R((\mathbb{I} - \eta H(\w_0))^{2\tau}) \\
& = \sum_{s =1}^\tau (1+\eta \lambda)^{2\tau-2s} \sigma^2 \Lambda_\tau > \frac{ \sigma^2 \chi^{2\tau} \Lambda_\tau}{8\eta\lambda} 
\end{align*}


\textbf{(III) Lower-bound on $\mathbb{E} [A_\tau^{\top} (B_\tau -C_\tau)]$}. 
\begin{align*}
 \mathbb{E} [A_\tau^{\top} (B_\tau -C_\tau)]  &= \mathbb{E}\left\langle (\mathbb{I}-\eta H(\w_0))^\tau\left(\mathbf{w}_{1}-\mathbf{w}_{0}\right),~ \sum_{s=1}^\tau(\mathbb{I}-\eta H(\w_0))^{\tau-s}\left(\h_s - \tilde \g_s - \nabla \cL_{S}(\w_0) \right) 
 \right\rangle \\
 & = \eta \mathbb{E}\left\langle (\mathbb{I}-\eta H(\w_0))^\tau\left(-\tilde \g_0 - \b_0\right),~ \sum_{s=1}^\tau(\mathbb{I}-\eta H(\w_0))^{\tau-s}\left(\h_s - \tilde \g_s - \nabla \cL_{S}(\w_0) \right)
 \right\rangle \\
 & \overset{(a)}{=} -\eta \mathbb{E}\left\langle (\mathbb{I}-\eta H(\w_0))^\tau\tilde \g_0,~ \sum_{s=1}^\tau(\mathbb{I}-\eta H(\w_0))^{\tau-s}\left(\h_s - \tilde \g_s - \nabla \cL_{S}(\w_0) \right)
 \right\rangle\\
 & \geq - \eta \underbrace{\left\| (\mathbb{I}-\eta H(\w_0))^\tau\tilde \g_0\right\|}_{U_{\tau,1}} \cdot 
 \underbrace{\mathbb{E} \left\| \sum_{s=1}^\tau(\mathbb{I}-\eta H(\w_0))^{\tau-s}\left(\h_s - \tilde \g_s - \nabla \cL_{S}(\w_0) \right)\right\|}_{U_{\tau,2}}~,
\end{align*}
where $(a)$ holds because $\mathbb{E}[\b_0] = 0$ and $\b_0$ is independent of  $B_\tau - C_\tau$.

We upper bound $U_{\tau,1}$:

\begin{align*}
 U_{\tau,1}  =   \left\| (\mathbb{I}-\eta H(\w_0))^\tau\tilde \g_0\right\| 
 \leq (1+\eta \lambda)^\tau G \sqrt{1-\varepsilon^2}
  = \chi^\tau G.
\end{align*}

We can also upper bound $U_{\tau,2}$:
\begin{align*}
   U_{\tau,2}  & = \mathbb{E} \left\| \sum_{s=1}^\tau(\mathbb{I}-\eta H(\w_0))^{\tau-s}\left(\h_s - \tilde \g_s - \nabla \cL_{S}(\w_0) \right)\right\|\\
   & \leq \mathbb{E}\left\| \sum_{s=1}^\tau(\mathbb{I}-\eta H(\w_0))^{\tau-s}\left(\h_s - \tilde \g_s \right)\right\| + \mathbb{E} \left\| \sum_{s=1}^\tau(\mathbb{I}-\eta H(\w_0))^{\tau-s} \nabla \cL_{S}(\w_0) \right\| \\
   & \overset{(a)} {\leq} \mathbb{E}\sum_{s=1}^\tau \left\| (\mathbb{I}-\eta H(\w_0))^{\tau-s}\right\|_2\left((2-\varepsilon^2) G + \frac{\rho}{2}\|\w_s -\w_0) \|^2\right) +  \mathbb{E} \sum_{s=1}^\tau G \left\| (\mathbb{I}-\eta H(\w_0))^{\tau-s} \right\|_2 \\
   & = \sum_{s=1}^\tau(1+\eta\lambda)^{\tau-s}\left(G(3 - \varepsilon^2) + \frac{\rho}{2}\mathbb{E}\|\w_s -\w_0 \|^2 \right) \\
   & \overset{(b)} {\leq} \sum_{s=1}^\tau(1+\eta\lambda)^{\tau-s}\left(G(3 - \varepsilon^2) + \frac{\rho}{2}\left(4 \eta F \cdot s  + 2\eta^2 p\sigma^2 \cdot s  + 20 \eta^2 G^2 \cdot s^2 +  2\eta^3 L p \sigma^2  \cdot s^2
     \right)\right)\\
   & \overset{(c)}{=} \chi^\tau \left( \frac{2G(3 -\varepsilon)}{\eta \lambda} + \frac{4\rho F}{\eta\lambda^2} + \frac{2\rho p \sigma^2}{\lambda^2} + \frac{60 \rho G^2}{\eta\lambda^3} + \frac{6\rho p \sigma^2 L}{\lambda^3}\right)~,
\end{align*}
where $(a)$ holds because $\|\h_s - \tilde \g_s\| \leq (2-\varepsilon^2)G + \frac{\rho}{2}\|\w_s -\w_0\|$ from \eqref{eq:grad_by_param}; $(b)$ holds because of Lemma \ref{lemm:ub_distance} and $(c)$ follows Lemma \ref{lemma:bs}. Now we have

\begin{align*}
\mathbb{E} [A_\tau^{\top} (B_\tau -C_\tau)] &\geq - \eta U_{\tau,1} U_{\tau,2} \\
& \geq - \eta \chi^{2\tau}  \left( \frac{2G(3 -\varepsilon^2)}{\eta \lambda} + \frac{4\rho F}{\eta\lambda^2} + \frac{2\rho p \sigma^2}{\lambda^2} + \frac{60 \rho G^2}{\eta\lambda^3} + \frac{6\rho p \sigma^2 L}{\lambda^3}\right)
\end{align*}

Combine (I) (II) and (III), we have 

\begin{align*}
    \mathbb{E}\|\w_\tau -\w_0 \|^2 & \geq \mathbb{E}\left[\left\|A_\tau\right\|^{2}\right]+\eta^{2} \mathbb{E}\left[\left\|D_\tau\right\|^{2}\right]+2 \eta \mathbb{E}\left[A_\tau^{\top}\left(B_\tau-C_{t}\right)\right] \\
    & \geq \chi^{2\tau}\left(\eta^2\sigma^2 + \frac{\eta^2\sigma^2 \Lambda_\tau}{8\eta\lambda}  - 2 \eta^2G \left( \frac{2G(3 -\varepsilon^2)}{\eta \lambda} + \frac{4\rho F}{\eta\lambda^2} + \frac{2\rho p \sigma^2}{\lambda^2} + \frac{60 \rho G^2}{\eta\lambda^3} + \frac{6\rho p \sigma^2 L}{\lambda^3}\right) \right) \\
    & \geq \chi^{2\tau} \left(\eta^2\sigma^2 + \frac{\eta^2\sigma^2\Lambda_\tau}{8\eta\lambda}  - \frac{4 \eta (3 - \varepsilon^2)G^2}{\lambda} - \frac{8 \eta \rho F G}{\lambda^2} - \frac{4 \eta^2 p \sigma^2 \rho G}{\lambda^2} - \frac{120 \eta \rho G^3}{\lambda^3} - 
    \frac{12\eta^2 \rho p\sigma^2 LG}{\lambda^3}  \right)
\end{align*}

To make the lower bound to be positive, we choose the parameters to make the following condition holds

\begin{align*}
\eta^2\sigma^2 + \frac{\eta^2\sigma^2\Lambda_\tau}{8\eta\lambda}  - \underbrace{\frac{4 \eta (3 - \varepsilon^2)G^2}{\lambda}}_{\leq \frac{\eta^2\sigma^2\Lambda_\tau}{6 \cdot 8\eta\lambda}} - \underbrace{\frac{8 \eta \rho F G}{\lambda^2}}_{{\leq \frac{\eta^2\sigma^2\Lambda_\tau}{6 \cdot 8\eta\lambda}}} - \underbrace{\frac{4 \eta^2 p \sigma^2 \rho G}{\lambda^2}}_{\leq \frac{\eta^2\sigma^2\Lambda_\tau}{6 \cdot 8\eta\lambda}} - \underbrace{\frac{120 \eta \rho G^3}{\lambda^3}}_{\leq \frac{\eta^2\sigma^2\Lambda_\tau}{6 \cdot 8\eta\lambda}} - 
\underbrace{\frac{12\eta^2 \rho p\sigma^2 LG}{\lambda^3} }_{\leq \frac{\eta^2\sigma^2\Lambda_\tau}{6 \cdot 8\eta\lambda}} \geq \eta^2\sigma^2 +  \frac{\eta^2\sigma^2\Lambda_\tau}{6 \cdot 8\eta\lambda}
\end{align*}

To satisfy the above conditions, with $|\lambda_{\min}(H(\w_0))| = \lambda \geq \sqrt{\rho \gamma}$, we have 
\begin{align*}
 \eta \leq \min\{\frac{1}{L},~ \frac{\sqrt{\gamma} \Lambda_\tau}{144\sqrt{\rho} p G},~ \frac{\gamma \Lambda_\tau}{576p GL}\}, \quad  \sigma^2 \geq 576 G^2 \cdot \max \{\frac{ 1}{\Lambda_\tau}, \frac{10 G}{\gamma \Lambda_\tau}\}, \quad \text{and} \quad F \leq \frac{3 \sqrt{\gamma} G}{2\sqrt{\rho}}~.    
\end{align*}

These choices of parameters establish an exponential lower bound on the distance as

\begin{align*}
    \mathbb{E}[\| \w_\tau -\w_0\|^2] \geq \chi^{2\tau}\left(\eta^2\sigma^2 + \frac{\eta^2\sigma^2 \Lambda_\tau}{48\eta\lambda}\right).
\end{align*}

To derive the contradiction, we have to choose the number of steps such that the established lower-bound exceeds the upper-bound in Lemma \ref{lemm:ub_distance}:

\begin{align*}
 \chi^{2\tau}\left(\eta^2\sigma^2 + \frac{\eta^2\sigma^2 \Lambda_\tau}{48\eta\lambda}\right) \geq  \left(4 \eta F + 2\eta^2 p\sigma^2 \right) \cdot \tau + \left(12 \eta^2 G^2 + 2\eta^3 L p \sigma^2\right)\cdot \tau^2 ~.
\end{align*}

Since the left hand side is exponentially growing, we can derive the contradiction by choosing a large enough number of steps as:
\begin{align*}
    \tau \geq \left(24 + 4\log\left(\frac{\sqrt{\gamma}/(2\sqrt{\rho}\eta + G)}{ G \max\{1, 10G/\gamma\}}+ 4p\right)  \right)/\left(\eta^2\rho\gamma\right). 
\end{align*}
That completes the proof. \qed

\begin{lemm} \label{lemma:bs}
\textbf{(Lemma 6 in \cite{daneshmand2018escaping})}
For all $1>\beta>0,$ the following series are bounded as
\begin{equation*}
\sum_{i=1}^{t}(1+\beta)^{t-i}  \leq 2 \beta^{-1}(1+\beta)^{t}~,
\end{equation*}

\begin{equation*}
  \sum_{i=1}^{t}(1+\beta)^{t-i} i  \leq 2 \beta^{-2}(1+\beta)^{t} ~,
\end{equation*}

\begin{equation*}
  \sum_{i=1}^{t}(1+\beta)^{t-i} i^{2}  \leq 6 \beta^{-3}(1+\beta)^{t}~. 
\end{equation*}
\end{lemm}

\section{Proofs for Section \ref{sec:ntsgd_gen}}
\label{sec:app_gen}
Proposition \ref{prop: stab_hellinger} is from \cite{mou2018generalization} that controls stability via squared Hellinger distance.

\propstabhellinger*

\proof The proof follows \cite{mou2018generalization}. Take $p(\w)$ and $p^\prime(\w)$ as the probability density functions over $\w$ from running $\cA$ with $S$ and $S^\prime$. Thus, we have
\begin{align*}
\epsilon_{n} &=\sup _{x, S, S^{\prime}}\left|\int_{\mathbb{R}^{d}} \ell(\w ; z) p(\w) d \w-\int_{\mathbb{R}^{d}} \ell(\w ; z) p^\prime(\w) d \w\right| \\
&=\sup _{x, S, S^{\prime}}\left|\int_{\mathbb{R}^{d}} \ell(\w ; z)\left(\sqrt{p(\w)}+\sqrt{p(\w)^{\prime}}\right)\left(\sqrt{p(\w)}-\sqrt{p(\w)^{\prime}}\right) d \w\right| \\
& \leq \sup \left\{\left(\int_{\mathbb{R}^{d}} \ell(\w ; z)^{2}\left(\sqrt{p(\w)}+\sqrt{p(\w)^{\prime}}\right)^{2} d \w\right)^{\frac{1}{2}}\left(\int_{\mathbb{R}^{d}}\left(\sqrt{p(\w)}-\sqrt{p(\w)^{\prime}}\right)^{2} d \w\right)^{\frac{1}{2}}\right\} \\
&<2 C \sqrt{D_{H}\left(p(\w) \| p(\w)^{\prime}\right)}~,
\end{align*}
where the first inequality follows by Cauchy-Schwarz and the last inequality follows by bounded loss function. 

One can also consider bounding the stability using Rényi divergence or Kullback–Leibler divergence, since the squared Hellinger distance can be bounded by Rényi divergence of any order $\lambda > 1/2$, i.e., $D_{H}(p \| q) \leq D_{\lambda}(p \| q), \forall \lambda \geq \frac{1}{2}$. In this paper, we follow the approach of bounding the squared Hellinger distance. 

\begin{defn}
[Rényi divergence] For two probability distribution $p$ and $q$ (density functions), the Rényi divergence of order $\lambda \in(0,1) \cup(1, \infty)$ is \vspace*{-2mm}
\begin{equation}
D_{\lambda}(p \| q) \triangleq \frac{1}{\lambda-1} \log \mathbb{E}_{x \sim q}\left(\frac{p(x)}{q(x)}\right)^{\lambda}~.   
\end{equation}
\end{defn}

The Rényi divergence for the special values $\lambda = 0, ~1, ~\infty$ is defined by taking a limit. In particular $\lambda =1$ gives the Kullback–Leibler divergence, i.e., $D_1(p\|q) = KL(p \| q) = \mathbb{E}_{x \sim p}\left[\frac{p(x)}{q(x)}\right]$ and $\lambda = \infty$ gives the log of the maximum ratio of the probabilities, i.e., $D_{\infty} = \log \sup_{x} \frac{p(x)}{q(x)}$~.

\lemmstephell*


\proof Consider two neighboring datasets $S$ and $S^\prime$
differing only in the $i_*$-th sample, i.e., $S = \{z_1, ..., z_{i_*}, .., z_n\}$ and $S^\prime = \{z_1, ..., z^\prime_{i_*}, .., z_n\}$, at iteration $t$, we write the density function of $\w_t$ and $\w^\prime_t$ as $p_t$ and $p_t^\prime$ respectively for all $t \in [T]$.
To bound the Hellinger distance between $p_T$ and $p_T^\prime$, we will focus on bounding $D_{H}\left(p_t \| p_t^\prime\right)$ for $t \in [T]$.
Recall the update at $t$-th iteration is 
\begin{align*}
\mathbf{w}_{t+1}=\mathbf{w}_{t}-\eta_{t}\tilde{\mathbf{g}}_{t}+ \eta_t^{\frac{1}{2} + \beta}\mathbf{b}_{t}, ~\text{where},~ \mathbf{b}_{t} \sim \mathcal{N}\left(0, \sigma^{2} \mathbb{I}\right).
\end{align*}
Note that $\tilde \g_t$ is the truncated gradient which depends on $\w_t$ and mini-batch $B_t$. We consider the case when $|B_t| = 1$ for all $t \in [T]$, meaning that we randomly sample one sample from $S$ to compute the stochastic gradient. Thus, we can write mini-batch gradient $g(\w_t, B_t) = \nabla \ell(\w_t, z_{i_t})$, where $i_t$ is uniformly sampled from $\{1,..., n\}$. Let $\tilde \g_t = \tilde \nabla \ell(\w_t, z_{i_t})$. Then we can write the update as
\begin{align*}
\mathbf{w}_{t+1}=\mathbf{w}_{t}-\eta_{t} \tilde \nabla \ell(\w_t, z_{i_t}) + \eta_t^{\frac{1}{2} + \beta} \sigma \mathcal{N}\left(0,\mathbb{I}\right)~.  
\end{align*}

We can also view the update as 
\begin{align*}
\mathbf{w}_{t+1}=\mathbf{w}_{t}-\eta_{t} (1-X_t)\tilde \nabla \ell(\w_t, z_{i_t}) - \eta_t X_t \tilde \nabla \ell(\w_t, z_{i_*}) + \eta_t^{\frac{1}{2} + \beta} \sigma \mathcal{N}\left(0,\mathbb{I}\right)~,
\end{align*}
where $\w_t$, $X_t$ and $i_t$ are independent random variable with $\w_t \sim p_t$, $X_t \sim \text{Bernoulli}(\frac{1}{n})$ and $i_t \sim \text{Uniform} \left(\{1, \ldots, n\}\backslash\left\{i_{*}\right\}\right)$. 

For the $t$-th step update, we consider random variable $\bm{\theta}_s, \forall s \in [0, \eta_t]$ given by
\begin{align*}
    \bm{\theta}_s = \w_t - \eta_t \tilde \nabla \ell(\w_t, z_{i_t}) - s X_t \left( \tilde \nabla \ell(\w_t, z_{i_*}) - \tilde \nabla \ell(\w_t, z_{i_t}) \right)  +   \mathcal{N}\left(0, s \cdot \eta_t^{2\beta} \sigma^2 \mathbb{I}\right)~,
\end{align*}
which can be written as $\forall s \in [0, \eta_t]$, with $\bm{\theta}_0 = \w_t - \eta_t \tilde \nabla \ell(\w_t, z_{i_t})$,
\begin{align*}
    \bm{\theta}_s = \bm{\theta}_0 - \int_0^s X_t \left( \tilde \nabla \ell(\w_t, z_{i_*}) - \tilde \nabla \ell(\w_t, z_{i_t}) \right) d \tau +   \eta_t^{\beta} \sigma \int_0^s d B_\tau~.
\end{align*}

Following \cite{raginsky2017non, mou2018generalization}, consider the conditional expectation 
\begin{align*}
    g_s(\w) = \mathbb{E}_{\mathbf{w}_{t}, i_{t}, X_{t}}\left[X_t \left( \tilde \nabla \ell(\w_t, z_{i_*}) - \tilde \nabla \ell(\w_t, z_{i_t}) \right) \mid \boldsymbol{\theta}_{s}=\mathbf{w}\right]~.
\end{align*}
The mimicking distribution results  guarantees that $\bm{\theta}_s$ has the same marginals as the Ito process
\begin{align*}
    \mathbf{v}_{s}=\mathbf{v}_{0}-\int_{0}^{s} g_{\tau}\left(\mathbf{v}_{\tau}\right) d \tau + \eta_t^{\beta} \sigma \int_{0}^{s} d \mathbf{B}_{\tau}~,
\end{align*}
where $\v_0 = \u_0 - \eta_t\tilde \nabla f(\u_0, z_{i_t}), \u_0 \sim p_k$. 
Equivalently, the solution to the following SDE has the same one-time marginal as $\bm{\theta}_s$:
\begin{align*}
d \mathbf{v}_{\tau}=- g_{\tau}\left(\mathbf{v}_{\tau}\right) d \tau + \eta_t^{\beta} \sigma  d \mathbf{B}_{\tau}~.   
\end{align*}

Let $\pi(\bm{\theta}, s)$ and $\pi(\v, s)$ denote the 
marginal pdfs of $\bm{\theta}_s$ and $\v_s$ respectively,  we have $\pi(\bm{\theta}, s) = \pi(\v, s)$. Thus, for our analysis, it suffices to focus on $\pi(\v, s)$, which satisfies the Fokker-Planck equation
\begin{align*}
\frac{\partial \pi(\mathbf{v}, s)}{\partial s}= \nabla \cdot\left(\pi(\mathbf{v}, s) g_{s}\left(\mathbf{v}_{s}\right)\right) + \frac{(\eta_t^{\beta}\sigma)^2}{2} \Delta[\pi(\mathbf{v}, s)]~,
\end{align*}
where $\Delta$ is the Laplace operator. For the counterparts for the neighboring dataset $S^\prime$, denoted by $\pi^\prime(\v, s)$, we have
\begin{align*}
\frac{\partial \pi^\prime(\mathbf{v}, s)}{\partial s}= \nabla \cdot\left(\pi^\prime(\mathbf{v}, s) g_{s}^\prime\left(\mathbf{v}_{s}\right)\right) + \frac{(\eta_t^{\beta}\sigma)^2}{2} \Delta[\pi^\prime(\mathbf{v}, s)]~,
\end{align*}

Now we can bound the variation of squared Hellinger distance. Denote $\pi_s$ and $\pi^\prime_s$ for  $\pi(\v, s)$ and $\pi^\prime(\v, s)$ for short. For step $t$ and $s \in [0, \eta_t]$,  from \cite{mou2018generalization},  we have
\begin{align*}
\frac{d}{d s} D_{H}\left(\pi_{s} \| \pi_{s}^{\prime}\right) & = -\int_{\mathbb{R}^{d}} \frac{\partial}{\partial s} \sqrt{\pi_s \pi_s^{\prime}} d \w \\
&=-\int_{\mathbb{R}^{d}} \frac{\sqrt{\pi_s^{\prime}}}{2 \sqrt{\pi_s}} \frac{\partial \pi_s}{\partial s} d \w-\int_{\mathbb{R}^{d}} \frac{\sqrt{\pi_s}}{2 \sqrt{\pi_s^{\prime}}} \frac{\partial \pi^{\prime}}{\partial s} d \w \\
& = -\int_{\mathbb{R}^{p}} \frac{\sqrt{\pi_s^{\prime}}}{2 \sqrt{\pi_s}}\left[\frac{(\eta_t^{\beta}\sigma)^2}{2} \Delta \pi_s +\nabla \cdot\left(\pi_s g_{s}(\w)\right)\right] d \mathbf{w}-\int_{\mathbb{R}^{p}} \frac{\sqrt{\pi_s}}{2 \sqrt{\pi_s^{\prime}}}\left[\frac{(\eta_t^{\beta}\sigma)^2}{2} \Delta \pi_s^{\prime}+\nabla \cdot\left(\pi_s^{\prime} g_{s}^{\prime}(\w)\right)\right] d \mathbf{w} \\
&=-\frac{1}{4} \int_{\mathbb{R}^{d}} \sqrt{\pi_s \pi_s^{\prime}}\left(\frac{(\eta_t^{\beta}\sigma)^2}{2}\left\|\nabla \log \frac{\pi_s^{\prime}}{\pi_s}\right\|^{2}  - \nabla \log \frac{\pi_s^{\prime}}{\pi_s} \cdot\left(g_{s}(\w)-g_{s}^{\prime}(\w)\right)\right) d \w \\
& \leq \frac{1}{8(\eta_t^{\beta}\sigma)^2} \int_{\mathbb{R}^{d}} \sqrt{\pi_s \pi_s^{\prime}}\left\|g_{s}(\w)-g_{s}^{\prime}(\w)\right\|^{2} d \w
\end{align*}

where the first inequality follows the chain rule, the second inequality follows the Fokker-Planck equation, and third inequality follows the integration by parts. Note that
\begin{align*}
&\left\|\sqrt{\frac{(\eta_t^{\beta}\sigma)^2}{2}}\left(\nabla \log \frac{\pi_s^{\prime}}{\pi_s}\right)-\sqrt{\frac{1}{2(\eta_t^{\beta}\sigma)^2}}\left(g_{s}(\w)-g_{s}^{\prime}(\w)\right)\right\|^{2} \geq 0 \\
\Rightarrow \quad-&\left[\frac{(\eta_t^{\beta}\sigma)^2}{2}\left\|\nabla \log \frac{\pi^{\prime}}{\pi}\right\|^{2}-\nabla \log \frac{\pi^{\prime}}{\pi} \cdot\left(g_{s}(\w)-g_{s}^{\prime}(\w)\right)\right] \leq \frac{1}{2(\eta_t^{\beta}\sigma)^2}\left\|g_{s}(\w)-g_{s}^{\prime}(\w)\right\|^{2}~. 
\end{align*}
Thus, with Lemma \ref{lemma: mou_lemma4} below, we have
\begin{align*}
\frac{d}{d s} D_{H}\left(\pi_{s} \| \pi_{s}^{\prime}\right) &  \leq \frac{1}{8(\eta_t^{\beta}\sigma)^2} \int_{\mathbb{R}^{d}} \sqrt{\pi_s \pi_s^{\prime}}\left\|g_{s}(\w)-g_{s}^{\prime}(\w)\right\|^{2} d \w \\
& \leq \frac{16(1-\epsilon^2)G^2}{\eta_t^{2\beta} \sigma^2 n^2}~. 
\end{align*}

So we have 
\begin{align*}
    D_{H}\left(p_{t+1} \| p_{t+1}^{\prime}\right)=D_{H}\left(\pi_{\eta_{t}} \| \pi_{\eta_{t}}^{\prime}\right) \leq  D_{H}\left(\pi_{0} \| \pi_{0}^{\prime}\right)+ \frac{16(1-\epsilon^2)G^2}{\eta_t^{2\beta} \sigma^2 n^2} \eta_t \leq D_{H}\left(p_{t} \| p_{t}^{\prime}\right) +  \frac{16(1-\epsilon^2)G^2\eta_t^{1-2\beta}}{\sigma^2 n^2}~
\end{align*}
where the last inequality follows the non-expansive property of $f$-divergence (including KL divergence and squared Hellinger distance). 
Note that $\pi_0$ and  $\pi_0^\prime$ is the pdf of $\bm{\theta}_0 = \mathbf{w}_{t}-\eta_{t} \tilde{\nabla} \ell\left(\mathbf{w}_{t}, z_{i_{t}}\right)$ and $\bm{\theta}^\prime_0 = \mathbf{w}_{t}^\prime-\eta_{t} \tilde{\nabla} \ell\left(\mathbf{w}_{t}^\prime, z_{i_{t}}\right)$, where $\w_t \sim p_t$ and $\w^\prime_t \sim p^\prime_t$. Let $\phi()$ denote the gradient truncation function which only depends on the random sample $i_t$ and $\w_t$, we have $\tilde \nabla \ell(\w_t, z_{i_t}) \triangleq \phi(\nabla \ell(\w_t, z_{i_t}))$. Thus, $\bm{\theta}_0$ and $\bm{\theta}^\prime_0$ follows the same mapping function. Following the non-expansive property of squared Hellinger distance, we have 
\begin{align*}
D_{H}\left(\pi_{0} \| \pi_{0}^{\prime}\right) \leq   D_{H}\left(p_{t} \| p_{t}^{\prime}\right)~.
\end{align*}

That completes the proof. \qed

\theogenstab*

\proof 
Consider two neighboring datasets $S$ and $S^\prime$
differing only in the $i_*$-th sample, i.e., $S = \{z_1, ..., z_{i_*}, .., z_n\}$ and $S^\prime = \{z_1, ..., z^\prime_{i_*}, .., z_n\}$, at iteration $t$, we write the density function of $\w_t$ and $\w^\prime_t$ as $p_t$ and $p_t^\prime$ respectively for all $t \in [T]$. Given two distributions $p, q$, the Hellinger distance is given by:
\begin{align*}
D_{H}(p \| q) \triangleq \frac{1}{2} \int_{\mathbb{R}^{d}}(\sqrt{p}-\sqrt{q})^{2} d \mathbf{w}~.
\end{align*}

Assume the loss function is uniformly bounded by $C$, following \cite{mou2018generalization} and Proposition \ref{prop: stab_hellinger}, we have
\begin{align*}
    \alpha_T \leq 2 C \sqrt{D_{H}\left(p_T \| p_T^\prime\right)}~.
\end{align*}

From Lemma \ref{lemm: step_hell}, we have 
\begin{align*}
    D_{H}\left(p_{t+1} \| p_{t+1}^{\prime}\right) \leq D_{H}\left(p_{t} \| p_{t}^{\prime}\right) +  \frac{16(1-\epsilon^2)G^2\eta_t^{1-2\beta}}{\sigma^2 n^2}~
\end{align*}

By induction, we have
\begin{align*}
 D_{H}\left(p_{T} \| p_{T}^{\prime}\right) \leq    \frac{16(1-\epsilon^2)G^2 \sum_{t=1}^T \eta_t^{1-2\beta}}{\sigma^2 n^2}~.
\end{align*}
Thus, we have
\begin{align*}
    \alpha_T \leq 2 C \sqrt{D_{H}\left(p_{T} \| p_{T}^{\prime}\right)} \leq \frac{8CG \sqrt{(1-\epsilon^2)\sum_{t=1}^T \eta_t^{1-2\beta})}}{n \sigma }~. 
\end{align*}

\begin{defn}
(non-expansive). Suppose $\boldsymbol{w}$ and $\boldsymbol{w}^{\prime}$ are two random points in $\mathbb{R}^{d},$ and their distributions are denoted as $\mathcal{P}(\boldsymbol{w}), \mathcal{P}\left(\boldsymbol{w}^{\prime}\right) .$ We say a bivariate functional $D(\cdot \| \cdot)$ defined on two density functions, is non-expansive, if for any mapping $\psi: \mathbb{R}^{d} \rightarrow \mathbb{R}^{d},$ there is
\begin{align}
D\left(\mathcal{P}(\psi(\boldsymbol{w})) \| \mathcal{P}\left(\psi\left(\boldsymbol{w}^{\prime}\right)\right) \leqslant D\left(\mathcal{P}(\boldsymbol{w}) \| \mathcal{P}\left(\boldsymbol{w}^{\prime}\right)\right)\right.    
\end{align}
\end{defn}

\begin{lemm} \label{lemma: mou_lemma4}
Suppose  $\eta_t^{1-2\beta} \leq \frac{\sigma^2 \ln 4}{12 (1-\epsilon^2) G^2}$,  then there is 
\begin{align*}
\int \sqrt{\pi_s \pi_s^{\prime}}\left\|g_{s}(\w)-g_{s}^{\prime}(\w)\right\|^{2} d \w \leq \frac{128(1-\epsilon^2)G^2}{n^2}~
\end{align*}
\end{lemm}

\proof The proof mainly follows the Lemma 4 in \cite{mou2018generalization}. 
Let $u_{s}, u_{s}^{\prime}$ denote the pdfs of $\bm{\theta}_{s}, \bm{\theta}_{s}^{\prime}$ conditioned on $X=1$ respectively, and let $v_{s}, v_{s}^{\prime}$ denote the pdfs of $\bm{\theta}_{s}, \bm{\theta}_{s}^{\prime}$ conditioned on $X=0$ respectively. We have 
\begin{align*}
    g_{s}(\w)=\frac{u_{s}(\w)}{n \pi_{s}(\w)} \mathbb{E}\left(\tilde{\nabla} \ell\left(\mathbf{w}_{t}, z_{i_{*}}\right)-\tilde{\nabla} \ell\left(\mathbf{w}_{t}, z_{i_{t}}\right) \mid \boldsymbol{\theta}_{s}=\w\right)
\end{align*}
and
\begin{align*}
    g_{s}^\prime(\w)=\frac{u_{s}^\prime(\w)}{n \pi_{s}^\prime(\w)} \mathbb{E}\left(\tilde{\nabla} \ell\left(\mathbf{w}_{t}^\prime, z_{i_{*}}\right)-\tilde{\nabla} \ell\left(\mathbf{w}_{t}^\prime, z_{i_{t}}\right) \mid \boldsymbol{\theta}_{s}=\w\right)~.
\end{align*}

Based on the Assumption \ref{asmp: bounded_gradient}, we have
\begin{align*}
  \| g_{s}(\w)\| \leq \frac{2 u_{s}(\w) \sqrt{(1-\epsilon^2)} G}{n \pi_{s}(\w)}
\end{align*}
and
\begin{align*}
  \| g_{s}^\prime(\w)\| \leq \frac{2 u_{s}^\prime(\w) \sqrt{(1-\epsilon^2)} G}{n \pi_{s}^\prime(\w)}~.
\end{align*}

Then we have
\begin{align*}
\int \sqrt{\pi_s \pi_s^{\prime}}\left\|g_{s}(\w)-g_{s}^{\prime}(\w)\right\|^{2} d \w & \leq 2 \int \sqrt{\pi_{s} \pi_{s}^{\prime}}\| g_{s}(\w)  \|^{2} d \w +2 \int \sqrt{\pi_{s} \pi_{s}^{\prime}}\left\|g_{s}^{\prime}(\w)\right\|^{2} d \w \\
& \leq 2 \sqrt{\int \pi_{s}\|g_{s}(\w)\|^{4} \int \pi_{s}^{\prime}}+2 \sqrt{\int \pi_{s}^{\prime}\left\|g_{s}^{\prime}(\w)\right\|^{4} \int \pi_{s}}\\
& = 2 \sqrt{\int \pi_{s}\|g_{s}(\w)\|^{4} }+2 \sqrt{\int \pi_{s}^{\prime}\left\|g_{s}^{\prime}(\w)\right\|^{4} }\\
& \leq  2 \sqrt{\int \pi_{s} \left( \frac{2 u_{s}(\w) \sqrt{(1-\epsilon^2)} G}{n \pi_{s}(\w)}\right)^4 }+2 \sqrt{\int \pi_{s}^{\prime} \left(\frac{2 u_{s}^\prime(\w) \sqrt{(1-\epsilon^2)} G}{n \pi_{s}^\prime(\w)}\right)^{4} }\\
& \leq \frac{8(1-\epsilon^2)G^2}{(n-1)^2} \left( \sqrt{\int \frac{u_{s}^{4}}{v_{s}^{3}}} + \sqrt{\int \frac{u_{s}^{\prime 4}}{v_{s}^{\prime 3}}} \right)~.
\end{align*}

Following the argument in \cite{mou2018generalization}, we have
\begin{align*}
\frac{d}{d s} \int_{\mathbb{R}^{d}} \frac{u_{s}^{4}}{v_{s}^{3}} & = \int \frac{12 u_s^{4}}{v_s^{3}}\left\{-\frac{(\eta_t^{\beta}\sigma)^2}{2}\left\|\nabla \log \frac{v_s}{u_s}\right\|^{2}-\left(g_{s}(\w)-g_{s}^{\prime}(\w)\right) \cdot \nabla \log \frac{u_s}{v_s}\right\}\\
& \leq \frac{6}{ \eta_t^{2\beta} \sigma^2} \int \frac{ u_s^{4}}{v_s^{3}} \left\|g_{s}(\w)-g_{s}^{\prime}(\w)\right\|^{2}\\
& \leq \frac{12 (1-\epsilon^2)G^2}{ \eta_t^{2\beta}\sigma^2} \int \frac{ u_s^{4}}{v_s^{3}}~.
\end{align*}

Thus,
\begin{align*}
    \frac{d}{d s} \ln \int \frac{u_{s}^{4}}{v_{s}^{3}} \leq \frac{12 (1-\epsilon^2) G^2}{ \eta_t^{2\beta} \sigma^2}
\end{align*}

For $s \leq \eta_t$ and $\eta_t^{1-2\beta} \leq \frac{\sigma^2 \ln 4}{12 (1-\epsilon^2) G^2}$,
we have
\begin{align*}
\ln \int \frac{u_{s}^{4}}{v_{s}^{3}} \leq \frac{12 (1-\epsilon^2) G^2}{ \eta_t^{2\beta} \sigma^2} \cdot \eta_t \leq \ln 4~.
\end{align*}
Thus, 
\begin{align*}
  \int \frac{u_{s}^{4}}{v_{s}^{3}} \leq 4  ~.
\end{align*}

Similarly we have
\begin{align*}
  \int \frac{u_{s}^{\prime4}}{v_{s}^{\prime 3}} \leq 4 ~.
\end{align*}

As a result, we have 
\begin{align*}
\int \sqrt{\pi_s \pi_s^{\prime}}\left\|g_{s}(\w)-g_{s}^{\prime}(\w)\right\|^{2} d \w & \leq     \frac{8(1-\epsilon^2)G^2}{n-1} \left( \sqrt{\int \frac{u_{s}^{4}}{v_{s}^{3}}} + \sqrt{\int \frac{u_{s}^{\prime 4}}{v_{s}^{\prime 3}}} \right) \\
& \leq \frac{32(1-\epsilon^2)G^2}{(n-1)^2} \\
& \leq \frac{128(1-\epsilon^2)G^2}{n^2} ~.
\end{align*}

That completes the proof. \qed





\end{document}